\documentclass{article}
\usepackage{acl2023}
\usepackage{times}
\usepackage[utf8]{inputenc}
\usepackage{amsfonts}
\usepackage{xcolor}
\usepackage{listings}
\usepackage{amsmath}
\usepackage{multicol}
\usepackage{tabularx}
\usepackage[export]{adjustbox}
\usepackage{hyperref}
\usepackage{amssymb}
\usepackage{xspace}
\usepackage{graphicx}
\usepackage{enumitem}
\usepackage{pdfpages}
\usepackage{longtable}
\usepackage{titlesec}
\usepackage{booktabs}
\usepackage[font={footnotesize}]{caption}
\usepackage{wrapfig}

\graphicspath{ {./images/} }
\setcitestyle{square}

\hypersetup{
    colorlinks,
    linkcolor={blue!50!black},
    citecolor={blue!50!black},
    urlcolor={blue!50!black},
}

\definecolor{codegreen}{rgb}{0,0.6,0}
\definecolor{codegray}{rgb}{0.5,0.5,0.5}
\definecolor{codepurple}{rgb}{0.58,0,0.82}
\definecolor{backcolour}{rgb}{0.95,0.95,0.92}

\renewcommand{\top}{{\textrm{T}}}
\newcommand{\githublink}{\href{https://github.com/guyd1995/embedding-space}{https://github.com/guyd1995/embedding-space}}

\newcommand{\layernorm}{LayerNorm\xspace}
\newcommand{\sublayer}{module\xspace}
\newcommand{\Sublayer}{Module\xspace}

\newcommand{\dimint}{d_{\textit{ff}}}
\newcommand{\qatt}{Q_{\text{att}}}
\newcommand{\katt}{K_{\text{att}}}
\newcommand{\vatt}{V_{\text{att}}}
\newcommand{\WQK}{W_{\text{QK}}}
\newcommand{\WVO}{W_{\text{VO}}}

\newcommand{\WV}{W_{\text{V}}}
\newcommand{\WO}{W_{\text{O}}}
\newcommand{\WQ}{W_{\text{Q}}}
\newcommand{\WK}{W_{\text{K}}}
\newcommand{\softmax}{\textrm{softmax}}
\newcommand{\topk}{\texttt{top-k}}

\providecommand{\commentout}[1]{}

\lstdefinestyle{mystyle}{
    backgroundcolor=\color{backcolour},   
    commentstyle=\color{codegreen},
    keywordstyle=\color{magenta},
    numberstyle=\tiny\color{codegray},
    stringstyle=\color{codepurple},
    basicstyle=\ttfamily\footnotesize,
    breakatwhitespace=false,         
    breaklines=true,                 
    captionpos=b,                    
    keepspaces=true,                 
    numbersep=5pt,                  
    showspaces=false,                
    showstringspaces=false,
    showtabs=false,                  
    tabsize=2
}

\author{Guy Dar$^{1}$ ~~~~~Mor Geva$^{2}$ ~~~~~
Ankit Gupta$^{1}$ ~~~~~
Jonathan Berant$^{1}$ \\
$^1$The Blavatnik School of Computer Science, Tel-Aviv University \\
$^2$Allen Institute for Artificial Intelligence \\
\small{\texttt{\{guy.dar,joberant\}@cs.tau.ac.il}}, \ 
\small{\texttt{morp@allenai.org}}, \\
\small{\texttt{ankitgupta.iitkanpur@gmail.com}}
}

\title{Analyzing Transformers in Embedding Space}
\begin{document}
\maketitle
\begin{abstract}
Understanding Transformer-based models has attracted significant attention, as they lie at the heart of recent technological advances across machine learning.
While most interpretability methods rely on running models over inputs, recent work has shown that an input-independent approach, where parameters are interpreted directly without a forward/backward pass is feasible for \emph{some} Transformer parameters, and for two-layer attention networks.
In this work, we present a conceptual framework where \emph{all} parameters of a trained Transformer are interpreted by projecting them into the \emph{embedding space}, that is, the space of vocabulary items they operate on.
Focusing mostly on GPT-2 for this paper, we provide diverse evidence to support our argument. First, an empirical analysis showing that parameters of both pretrained and fine-tuned models can be interpreted in embedding space. Second, we present two applications of our framework: (a) aligning the parameters of different models that share a vocabulary, and (b) constructing a classifier \emph{without training} by
``translating'' the parameters of a fine-tuned classifier to parameters of a different model that was only pretrained. Overall, our findings show that at least in part, we can abstract away model specifics and understand Transformers in the embedding space.
\end{abstract}

\section{Introduction}
\begin{figure*}
    \centering
    \vspace*{-0.5cm}
    \hspace*{-0.19cm}
    \includegraphics[scale=0.48, valign=t]{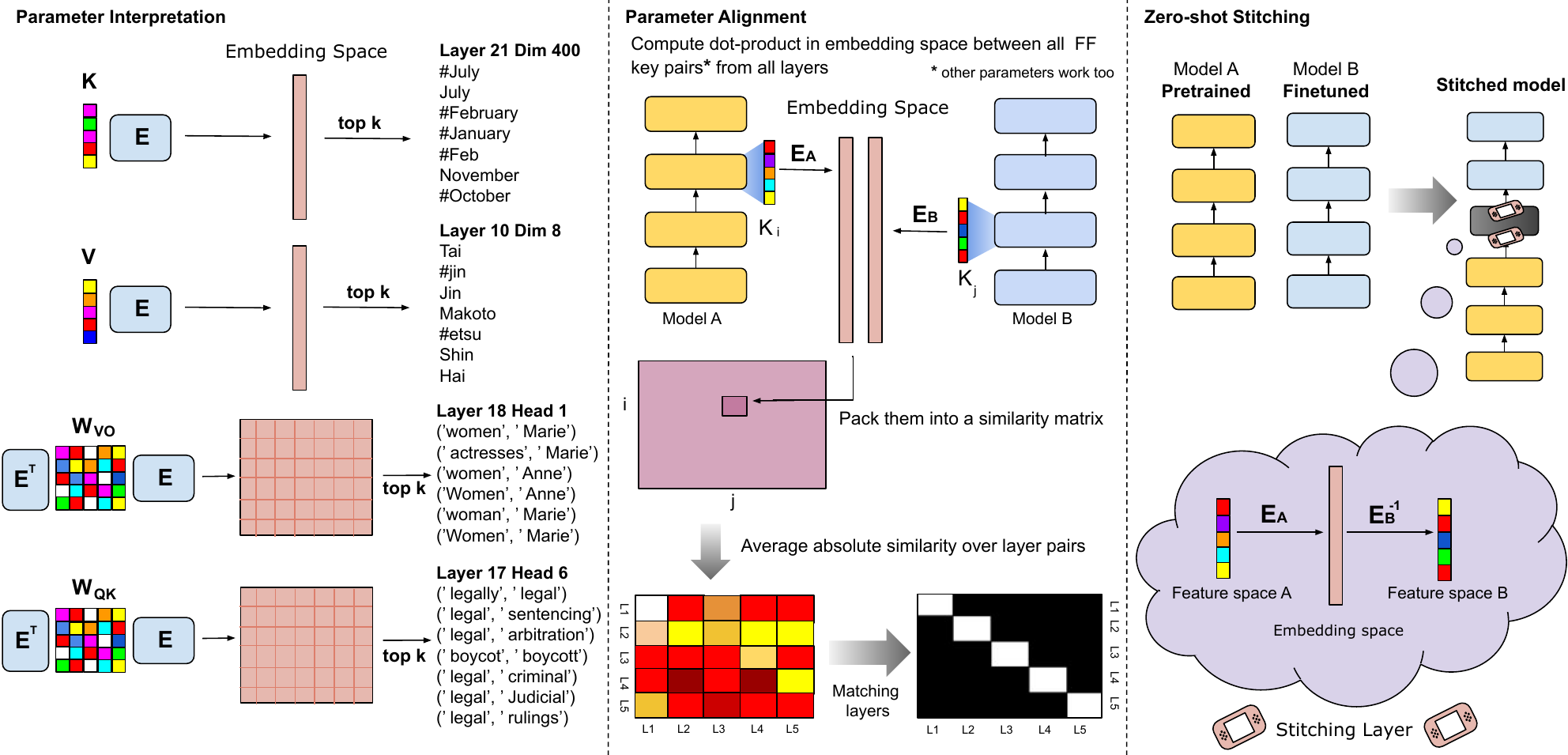}
\caption{Applications of the embedding space view. \emph{Left}: interpreting parameters in embedding space. The most active vocabulary items in a  feed-forward key ($k$) and a feed-forward value ($v$). The most active pairs of vocabulary items in an attention query-key matrix $\WQK$ and an attention value-output matrix $\WVO$ (see \S\ref{sec:background}).
\emph{Center}: Aligning the parameters of different BERT instances that share a vocabulary. \emph{Right}: Zero-shot ``stitching'', where representations of a fine-tuned classifier are translated through the embedding space (multiplying by $E_A E_B^{-1}$) to a pretrained-only model.}
    \label{fig:figure_1}
\end{figure*}
Transformer-based models \citep{vaswani2017} currently dominate Natural Language Processing \citep{bert, radford2019language, opt} as well as many other fields of machine learning \citep{vit, image_gpt, wav2vec2}. Consequently, understanding their inner workings has been a topic of great interest. Typically, work on interpreting Transformers relies on feeding inputs to the model and analyzing the resulting activations \citep{adi_probing, shi_probes, what_does_bert}. Thus, interpretation involves an expensive forward, and sometimes also a backward pass, over multiple inputs.
Moreover, such interpretation methods are conditioned on the input and are not guaranteed to generalize to all inputs. In the evolving literature on static interpretation, i.e., without forward or backward passes, \cite{geva2022} showed that the value vectors of the Transformer feed-forward \sublayer{} (the second layer of the feed-forward network) can be interpreted by projecting them into the embedding space, i.e., multiplying them by the embedding matrix to obtain a representation over vocabulary items.\footnote{We refer to the unique items of the vocabulary as \emph{vocabulary items}, and to the (possibly duplicate) elements of a tokenized input as \emph{tokens}. When clear, we might use the term \textit{token} for \textit{vocabulary item}.} \cite{anthropic} have shown that in a 2-layer attention network, weight matrices can be interpreted in the embedding space as well. Unfortunately, their innovative technique could not be extended any further.

In this work, 
we extend and unify the theory and findings of \cite{anthropic} and \cite{geva2022}. We present a zero-pass, input-independent framework to understand the behavior of Transformers. Concretely, we interpret \emph{all} weights of a pretrained language model (LM) in embedding space, including both keys and values of the feed-forward module (\cite{kv_memories, geva2022} considered just FF values) as well as all attention parameters (\cite{anthropic} analyzed simplified architectures up to two layers of attention with no MLPs). 

Our framework relies on a simple observation. Since \cite{geva2022} have shown that one can project hidden states to the embedding space via the embedding matrix, we intuit this can be extended to other parts of the model by projecting to the embedding space and then \textit{projecting back} by multiplying with a right-inverse of the embedding matrix. Thus, we can recast inner products in the model as inner products \emph{in embedding space}. Viewing inner products this way, we can interpret such products as
interactions between pairs of vocabulary items.
This applies to (a) interactions between attention queries and keys as well as to (b) interactions between attention value vectors and the parameters that project them at the output of the attention \sublayer{}.
Taking this perspective to the extreme, one can view Transformers as operating implicitly in the embedding space. This entails \textit{the existence of a single linear space} that depends only on the tokenizer, in which parameters of different Transformers can be compared.
Thus, one can use the embedding space to compare and transfer information across different models that share a tokenizer.

We provide extensive empirical evidence for the validity of our framework, focusing mainly on GPT-2 medium \citep{radford2019language}. We use GPT-2 for two reasons. First, we do this for concreteness, as this paper is mainly focused on introducing the new framework and not on analyzing its predictions. Second, and more crucially, unlike many other architectures (such as BERT \citep{bert}, RoBERTa \citep{roberta}, and T5 \citep{t5}), the GPT family has a \textit{linear} language modeling head (LM head) -- which is simply the output embedding matrix. All the other architectures' LM heads are two layer networks that contain \textit{non-linearities} before the output embedding matrix. Our framework requires a linear language modeling head to work. That being said, we believe in practice this will not be a major obstacle, and we indeed see in the experiments that model alignment works well for BERT in spite of the theoretical difficulties. We leave the non-linearities in the LM head for future work.

On the interpretation front (Fig.~\ref{fig:figure_1}, Left), we provide qualitative and quantitative evidence that Transformer parameters can be interpreted in embedding space. We also show that when fine-tuning GPT-2 on a sentiment analysis task (over movie reviews), projecting \emph{changes} in parameters into embedding space yields words that characterize sentiment towards movies. 
Second (Fig.~\ref{fig:figure_1}, Center), we show that given two distinct instances of BERT pretrained from different random seeds \citep{sellam2022multiberts}, we can align layers of the two instances by casting their weights into the embedding space. We find that indeed layer \emph{i} of the first instance aligns well to layer \emph{i} of the second instance, showing the different BERT instances converge to a semantically similar solution. 
Last (Fig.~\ref{fig:figure_1}, Right), we take a model fine-tuned on a sentiment analysis task and ``transfer'' the learned weights to a different model that was only pretrained by going through the embedding spaces of the two models. 
We show that in 30\% of the cases, this procedure, termed \emph{stitching}, results in a classifier that reaches an impressive accuracy of 70\% on the IMDB benchmark \citep{imdb} without any training. 

Overall, our findings suggest that analyzing Transformers in embedding space is valuable both as an interpretability tool and as a way to relate different models that share a vocabulary and that it opens the door to interpretation methods that operate in embedding space only. Our code is available at \githublink. 
\section{Background}
\label{sec:background}
We now present the main components of the Transformer \citep{vaswani2017} relevant to our analysis.
We discuss the residual stream view of Transformers, and recapitulate a view of the attention layer parameters as \emph{interaction matrices} $\WVO$ and $\WQK$ \citep{anthropic}. 
Similar to them, we exclude biases and layer normalization from our analysis. 

\subsection{Transformer Architecture}\label{sec:transformer-arch}

The Transformer consists of a stack of layers, each including an attention \sublayer{} followed by a Feed-Forward (FF) \sublayer{}. All inputs and outputs are sequences of $N$ vectors of dimensionality $d$.

\paragraph{Attention \Sublayer}
takes as input a sequence of representations $X \in \mathbb{R}^{N \times d}$, and each layer $L$ is parameterized by four matrices
$W^{(L)}_{Q}, W^{(L)}_{K}, W^{(L)}_{V}, W^{(L)}_{O} \in \mathbb{R}^{d \times d}$ (we henceforth omit the layer superscript for brevity). The input $X$ is projected to produce queries, keys, and values:
$\qatt = X W_Q, \katt = X W_K, \vatt = X W_V$. Each one of $\qatt, \katt, \vatt$ is split along the columns to $H$ different \textit{heads} of dimensionality $\mathbb{R}^{N \times \frac{d}{H}}$,  denoted by $\qatt^i, \katt^i, \vatt^i$ respectively. We then compute $H$ \textit{attention maps}:
    \[
    A^i = \softmax\left(\frac{\qatt^i \katt^{i \top}}{\sqrt{d / H}} + M\right) \in \mathbb{R}^{N \times N},
    \]
where $M \in \mathbb{R}^{N \times N}$ is the attention mask. Each attention map is applied to the corresponding value head as $A^i \vatt^i$, results are concatenated along columns and projected via $W_O$. The input to the \sublayer is added via a residual connection, and thus the attention \sublayer's output is:
\begin{align}
\label{eq:attn_eq}
X + \textbf{Concat}\Bigg[A^1 \vatt^1, \dots, A^i \vatt^i, \dots, A^{H} \vatt^{H}\Bigg] W_O.
\end{align}

\paragraph{FF \Sublayer} is a two-layer neural network, applied to each position independently. Following past terminology \citep{sukhbaatar2019augmenting,kv_memories}, weights of the first layer are called \textit{FF keys} and weights of the second layer \textit{FF values}. 
This is an analogy to attention, as the FF \sublayer too can be expressed as: $f(Q K^\top) V$, where $f$ is the activation function, 
$Q \in \mathbb{R}^{N \times d}$ is the output of the attention \sublayer and the input to the FF \sublayer, and $K, V \in \mathbb{R}^{\dimint \times d}$ are the weights of the first and second layers of the FF \sublayer. Unlike attention, keys and values are learnable parameters. The output of the FF \sublayer is added to the output of the attention \sublayer to form the output of the layer via a residual connection. The output of the $i$-th layer is called  the $i$-th \textit{hidden state}.

\paragraph{Embedding Matrix}
To process sequences of discrete tokens, Transformers use an embedding matrix $E \in \mathbb{R}^{d \times e}$ that provides a $d$-dimensional representation to vocabulary items before entering the \emph{first} Transformer layer. In different architectures, including GPT-2, the same embedding matrix $E$ is often used \citep{press_tying} to take the output of the \emph{last} Transformer layer and project it back to the vocabulary dimension, i.e., into the \emph{embedding space}. In this work, we show how to interpret all the components of the Transformer model in the embedding space.

\subsection{The Residual Stream}
\label{subsec:residual}


We rely on a useful view of the Transformer through its residual connections popularized by \cite{anthropic}.\footnote{Originally introduced in \cite{lesswrong_logit_lens}.} Specifically, each layer takes a hidden state as input and adds information to the hidden state through its residual connection. Under this view, the hidden state is a \textit{residual stream} passed along the layers, from which information is read, and to which information is written at each layer.
\cite{anthropic} and \cite{geva2022} observed that the residual stream is often barely updated in the last layers, and thus the final prediction is determined in early layers and the hidden state is mostly passed through the later layers.

An exciting consequence of the residual stream view is that we can project hidden states in \emph{every} layer into embedding space by multiplying the hidden state with the embedding matrix $E$, treating the hidden state as if it were the output of the last layer.
\cite{lm_debugger} used this approach to interpret the prediction of Transformer-based language models, and we follow a similar approach.

\subsection[WQK and WVO]{$\WQK$ and $\WVO$}
Following \cite{anthropic}, we describe the attention \sublayer{} in terms of \textit{interaction matrices} $\WQK$ and $\WVO$ which will be later used in our mathematical derivation.
The computation of the attention \sublayer{} (\S\ref{sec:transformer-arch}) can be re-interpreted as follows. The attention projection matrices $\WQ, \WK, \WV$ can be split along the \textit{column} axis to $H$ equal parts  denoted by $\WQ^i, \WK^i, \WV^i \in \mathbb{R}^{d \times \frac{d}{H}}$ for $1\leq i \leq H$. Similarly, the attention output matrix $\WO$ can be split along the \textit{row} axis into $H$ heads, $\WO^i \in \mathbb{R}^{\frac{d}{H} \times d}$.
We define the \textit{interaction matrices} as
\[
\begin{aligned}
\WQK^i := \WQ^i \WK^{i \top} \in \mathbb{R}^{d \times d}, \\
\WVO^i := \WV^i \WO^i \in \mathbb{R}^{d \times d}.
 \end{aligned}
\]
Importantly, $\WQK^i, \WVO^i$ are \textit{input-independent}. Intuitively, $\WQK$ encodes the amount of attention between pairs of tokens. Similarly, in $\WVO^i$, the matrices $\WV$ and $\WO$ can be viewed as a transition matrix that determines how attending to certain tokens affects the subsequent hidden state.

We can restate the attention equations in terms of the interaction matrices. Recall (Eq. \ref{eq:attn_eq}) that the output of the $i$'th head of the attention \sublayer is $A^i \vatt^i$ and the final output of the attention \sublayer is (without the residual connection):
\vspace*{-0.1cm}
\begin{align}
\textbf{Concat}\Bigg[A^1 \vatt^1, ..., A^i \vatt^i, ..., A^H \vatt^H \Bigg] \WO =  
\\
\sum_{i=1}^H A^i (X \WV^i) \WO^i =  \sum_{i=1}^H A^i X \WVO^i. \notag
\label{eq:wvo_eq}
\end{align}
Similarly, the attention map $A^i$ at the $i$'th head in terms of $\WQK$ is ($\softmax$ is done row-wise):
\begin{gather}
    A^i = \softmax\left(\frac{(X \WQ^i) (X \WK^i)^\top}{\sqrt{d / H}} + M\right)
    \\
     = \softmax\left( \frac{X (\WQK^i) X^\top}{\sqrt{d / H}}  +  M\right). \notag
\label{eq:wqk}
\end{gather}

\section{Parameter Projection}
\label{sec:theory}

In this section, we propose that Transformer parameters can be projected into embedding space for interpretation purposes. We empirically support our framework's predictions in \S\ref{sec:experiments}-\S\ref{sec:applications}. 

Given a matrix $A \in \mathbb{R}^{N \times d}$, we can project it into embedding space by multiplying by the embedding matrix $E$ as $\hat{A} = AE \in \mathbb{R}^{N \times e}$. 
Let $E'$ be a right-inverse of $E$, that is, $E E' = I \in \mathbb{R}^{d \times d}$.\footnote{$E'$ exists if $d \leq e$ and $E$ is full-rank.}
We can reconstruct the original matrix with $E'$ as $A =  A(E E') = \hat{A} E'$. We will use this simple identity to reinterpret the model's operation in embedding space. To simplify our analysis we ignore \layernorm and biases. This has been justified in prior work \citep{anthropic}. Briefly, \layernorm can be ignored because normalization changes only magnitudes and not the direction of the update. At the end of this section, we discuss why in practice we choose to use $E' = E^\top$ instead of a seemingly more appropriate right inverse, such as the pseudo-inverse \citep{moore1920, bjerhammar1951, penrose1955}. In this section, we derive our framework and summarize its predictions in Table~\ref{tab:proj_summary}.


\paragraph{Attention \Sublayer} \label{sec:w-vo-derivation}
Recall that $\WVO^i := \WV^i \WO^i \in \mathbb{R}^{d \times d}$ is the interaction matrix between attention values and the output projection matrix for attention head $i$. By definition, the output of each head is: $A^iX\WVO^i = A^i \hat{X} E' \WVO^i$.
Since the output of the attention \sublayer is  added to the residual stream, we can assume according to the residual stream view that it is meaningful to project it to the embedding space, similar to FF values.
Thus, we expect the sequence of $N$ $e$-dimensional vectors $(A^iX\WVO^i)E = A^i \hat{X} (E' \WVO^i E)$ to be interpretable.
Importantly, the role of $A^i$ is just to mix the representations of the updated $N$ input vectors. This is similar to the FF \sublayer, where FF values (the parameters of the second layer) are projected into embedding space, and FF keys (parameters of the first layer) determine the \textit{coefficients} for mixing them.
Hence, we can assume that the interpretable components are in the term $\hat{X} (E' \WVO^i E)$. 

Zooming in on this operation, we see that it takes the previous hidden state in the embedding space ($\hat{X}$) and produces an output in the embedding space which will be incorporated into the next hidden state through the residual stream.
Thus, $E' \WVO^i E$ is a \emph{transition matrix} that takes a representation of the embedding space and outputs a new representation in the same space.

Similarly, the matrix $\WQK^i$ can be viewed as a bilinear map (Eq. \ref{eq:wqk}). To interpret it in embedding space, we perform the following operation with $E'$:
    \begin{gather*}
    X \WQK^i X^\top = (X E E') \WQK^i (X E E')^\top  = \\
    (X E) E' \WQK^i E'^\top (X E)^\top = \hat{X} (E' \WQK^i E'^\top) \hat{X}^\top.
    \end{gather*}
Therefore, the interaction between tokens at different positions is determined by an $e \times e$ matrix that expresses the interaction between pairs of vocabulary items. 

\paragraph{FF \Sublayer} \cite{geva2022} showed that FF value vectors $V  \in \mathbb{R}^{\dimint \times d}$ are meaningful when projected into embedding space, i.e., for a FF value vector $v \in \mathbb{R}^{d}$, $v E \in \mathbb{R}^{e}$ is interpretable (see \S\ref{sec:transformer-arch}). In vectorized form, the rows of $VE \in \mathbb{R}^{\dimint \times e}$ are interpretable. On the other hand, the keys $K$ of the FF layer are multiplied on the left by the output of the attention \sublayer, which are the queries of the FF layer. Denoting the output of the attention \sublayer by $Q$, we can write this product as $QK^\top = \hat{Q} E' K^\top = \hat{Q} (K E'^
\top)^\top$. Because $Q$ is a hidden state, we assume according to the residual stream view that $\hat{Q}$ is interpretable in embedding space. When multiplying $\hat{Q}$ by $K E'^\top$, we are capturing the interaction in embedding space between each query and key, and thus expect $K E'^\top$ to be interpretable in embedding space as well.

Overall, FF keys and values are intimately connected -- the $i$-th key controls the coefficient of the $i$-th value, so we expect their interpretation to be related. While not central to this work, we empirically show that key-value pairs in the FF \sublayer{} are similar in embedding space in Appendix~\ref{appendix:related_pairs}. 

\begin{table*}[t]
\vspace*{-0.5cm}
\centering
\footnotesize
\begin{tabular}{ l c c c }
\toprule
   & \textbf{Symbol} & \textbf{Projection} & \textbf{Approximate Projection} \\
 FF values & $v$ & $vE$ & $v E$ \\ 
 FF keys & $k$ & $kE'^\top$ & $kE$ \\
 Attention query-key  & $\WQK^i$ & $E' \WQK^i E'^\top$ & $E^\top \WQK^i E$ \\  
 Attention value-output  & $\WVO^i$ & $E' \WVO^i E$ & $E^\top \WVO^i E$ \\
\midrule
Attention value subheads  & $\WV^{i, j}$ & $\WV^{i,j} E'^\top$ & $\WV^{i,j} E$ \\
Attention output subheads  & $\WO^{i, j}$ & $\WO^{i,j} E$ & $\WO^{i,j} E$ \\
Attention query subheads  & $\WQ^{i, j}$ & $\WQ^{i,j} E'^\top$ & $\WQ^{i,j} E$ \\
Attention key subheads  & $\WK^{i, j}$ & $\WK^{i,j} E'^\top$ & $\WK^{i,j} E$ \\
\bottomrule

\end{tabular}
\caption{A summary of our approach for projecting Transformer components into embedding space. The `Approximate Projection' shows the projection we use in practice where $E' = E^\top$.}
\label{tab:proj_summary}
\end{table*}

\paragraph{Subheads} Another way to interpret the matrices $\WVO^i$ and $\WQK^i$ is through the \textit{subhead view}. 
We use the following identity: $AB = \sum_{j=1}^{b} A_{:, j} B_{j, :}$, which holds for arbitrary matrices $A \in \mathbb{R}^{a \times b}, B \in \mathbb{R}^{b \times c}$, where $A_{:, j} \in \mathbb{R}^{a \times 1}$ are the \textit{columns} of the matrix $A$ and $B_{j, :} \in \mathbb{R}^{1 \times c}$ are the \textit{rows} of the matrix $B$.
Thus, we can decompose $\WVO^i$ and $\WQK^i$ into a sum of $\frac{d}{H}$ rank-1 matrices:  
\vspace*{-7pt}
\begin{gather*}
\WVO^i = \sum_{j=1}^{\frac{d}{H}} \WV^{i,j} \WO^{i,j}, \quad
\WQK^i = \sum_{j=1}^{\frac{d}{H}} \WQ^{i, j} {\WK^{i, j}}^\top.
\end{gather*}
\vspace*{-2pt}
where $\WQ^{i,j}, \WK^{i,j}, \WV^{i,j} \in \mathbb{R}^{d \times 1}$ are columns of $\WQ^i, \WK^i, \WV^i$ respectively, and $\WO^{i,j} \in \mathbb{R}^{1 \times d}$ are the rows of $\WO^i$. We call these vectors \textit{subheads}.
This view is useful since it allows us to interpret subheads directly by multiplying them with the embedding matrix $E$. Moreover, it shows a parallel between interaction matrices in the attention \sublayer{} and the FF \sublayer{}. Just like the FF \sublayer{} includes key-value pairs as described above, for a given head, its interaction matrices are a sum of interactions between pairs of subheads (indexed by $j$), which are likely to be related in embedding space. We show this is indeed empirically the case for pairs of subheads in Appendix \ref{appendix:related_pairs}. 

\textbf{Choosing $E'=E^\top$} \ \ In practice, we do not use an exact right inverse (e.g. the pseudo-inverse). We use the transpose of the embedding matrix $E' = E^\top$ instead. The reason pseudo-inverse doesn't work is that for interpretation we apply a top-$k$ operation after projecting to embedding space (since it is impractical for humans to read through a sorted list of $50K$ tokens). So, we only keep the list of the vocabulary items that have the $k$ largest logits, for manageable values of $k$. In Appendix~\ref{appendix:echoice}, we explore the exact requirements for $E'$ to interact well with top-$k$. We show that the top $k$ entries of a vector projected with the pseudo-inverse do not represent the entire vector well in embedding space. We define \textit{keep-$k$ robust invertibility} to quantify this. It turns out that empirically $E^\top$ is a decent \textit{keep-k robust inverse} for $E$ in the case of GPT-2 medium (and similar models) for plausible values of $k$. We refer the reader to Appendix~\ref{appendix:echoice} for details. 

To give intuition as to why $E^\top$ works in practice, we switch to a different perspective, useful in its own right. Consider the FF keys for example -- they are multiplied on the left by the hidden states. In this section, we suggested to re-cast this as $h^T K = (h^T E) (E' K)$. Our justification was that the hidden state is interpretable in the embedding space. A related perspective (dominant in previous works too; e.g. \cite{mickus2022dissect}) is thinking of the hidden state as an aggregation of interpretable updates to the residual stream. That is, schematically, $h = \sum_{i=1}^k \alpha_i r_i$, where $\alpha_i$ are scalars and $r_i$ are vectors corresponding to specific concepts in the embedding space (we roughly think of a concept as a list of tokens related to a single topic). Inner product is often used as a similarity metric between two vectors. If the similarity between a column $K_i$ and $h$ is large, the corresponding $i$-th output coordinate will be large. Then we can think of $K$ as a \textit{detector} of concepts where each neuron (column in $K$) lights up if a certain concept is ``present'' (or a superposition of concepts) in the inner state.  To understand which concepts each detector column encodes we see which tokens it responds to. Doing this for all (input) token embeddings and packaging the inner products into a vector of scores is equivalent to simply multiplying by $E^\top$ on the left (where $E$ is the input embedding in this case, but for GPT-2 they are the same). A similar argument can be made for the interaction matrices as well. For example for $\WVO$, to understand if a token embedding $e_i$ maps to a $e_j$ under a certain head, we apply the matrix to $e_i$, getting $e_i^T \WVO$ and use the inner product as a similarity metric and get the score $e_i^T \WVO e_j$.



\section{Interpretability Experiments}
\label{sec:experiments}
\begin{figure*}[t]
    \centering
    \vspace*{-1.5cm}
    \hspace*{-1.4cm}
    \includegraphics[width=19cm]{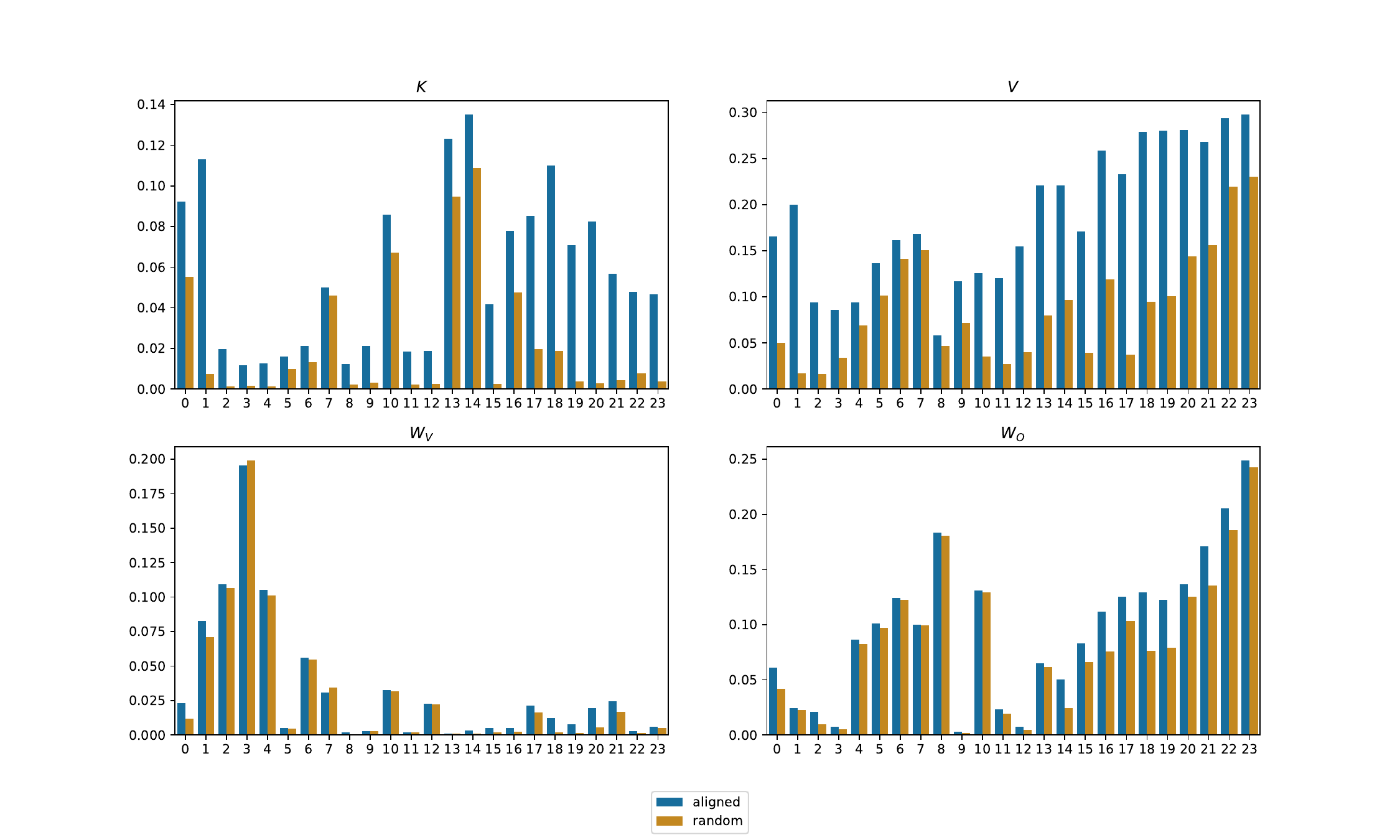}
    \caption{Left: Average $R_k$ score ($k=100$) across tokens per layer for activated parameter vectors against both the aligned hidden state $\hat{h}$ at the output of the layer and a randomly sampled hidden state $\hat{h}_\text{rand}$. Parameters are FF keys (top-left), FF values (top-right), attention values (bottom-left), and attention outputs (bottom-right).\vspace*{-10pt}
    }
    \label{fig:kv_vs_states}
\end{figure*}

In this section, we provide empirical evidence for the viability of our approach as a tool for interpreting Transformer parameters. For our experiments, we use Huggingface Transformers (\cite{huggingface}; License: Apache-2.0).


\subsection{Parameter Interpretation Examples}
\textit{Attention Module} \ \ \ We take GPT-2 medium (345M parameters; \cite{radford2019language}) and manually analyze its parameters. GPT-2 medium has a total of 384 attention heads (24 layers and 16 heads per layer). We take the embedded transition matrices $E' \WVO^i E$ for all heads and examine the top-$k$ pairs of vocabulary items. As there are only 384 heads, we manually choose a few heads and present the top-$k$ pairs in Appendix~\ref{appendix:samples_wvo} ($k=50$). We observe that different heads capture different types of relations between pairs of vocabulary items including word parts, heads that focus on gender, geography, orthography, particular part-of-speech tags, and various semantic topics. In Appendix~\ref{appendix:samples_wqk} we perform a similar analysis for $\WQK$. We supplement this analysis with a few examples from GPT-2 base and large (117M, 762M parameters -- respectively) as proof of concept, similarly presenting interpretable patterns.

A technical note: $\WVO$ operates on row vectors, which means it operates in a ``transposed'' way to standard intuition -- which places inputs on the left side and outputs on the right side. It does not affect the theory, but when visualizing the top-$k$ tuples, we take the transpose of the projection $(E' \WVO^i E)^\top$ to get the ``natural'' format \verb|(input token, output token)|. Without the transpose, we would get the \textit{same} tuples, but in the format \verb|(output token, input token)|. Equivalently, in the terminology of linear algebra, it can be seen as a linear transformation that we represent in the basis of row vectors and we transform to the basis of column vectors, which is the standard one.



\textit{FF Module} \ \ \ Appendix \ref{appendix:samples_ffn} provides examples of key-value pairs from the FF \sublayer{}s of GPT-2 medium. We show random pairs $(k, v)$ from the set of those pairs such that when looking at the top-100 vocabulary items for $k$ and $v$, at least 15\% overlap. Such pairs account for approximately 5\% of all key-value pairs. The examples show how key-value pairs often revolve around similar topics such as media, months, organs, etc. We again include additional examples from GPT-2 base and large.

\textit{Knowledge Lookup} \ \ \ Last, we show we can use embeddings to locate FF values (or keys) related to a particular topic. We take a few vocabulary items related to a certain topic, e.g., \texttt{[`cm', `kg', `inches']}, average their embeddings,\footnote{We subtract the average embedding $\mu$ from $E$ before averaging, which improves interpretability.} and rank all FF values (or keys) based on their dot-product with the average. Appendix \ref{appendix:examples_lookup} shows a few examples of FF values found with this method that are related to programming, measurements, and animals.



\vspace*{-0.1cm}
\subsection{Hidden State and Parameters}
\vspace*{-0.1cm}
\label{subsec:weights_vs_states}
One merit of zero-pass interpretation is that it does not require running inputs through the model. Feeding inputs might be expensive and non-exhaustive. 
In this section and \textit{in this section only}, we run a forward pass over inputs and examine if the  embedding space representations of dynamically computed hidden states are ``similar'' to the representations of the activated static parameter vectors. Due to the small number of examples we run over, the overall GPU usage is still negligible.

A technical side note: we use GPT-2, which applies \layernorm to the Transformer output before projecting it to the embedding space with $E$. Thus, conservatively, \layernorm should be considered as part of the projection operation. 
Empirically, however, we observe that projecting parameters directly without \layernorm works well, which simplifies our analysis in \S\ref{sec:theory}. Unlike parameters, we apply \layernorm to hidden states before projection to embedding space to improve interpretability. This nuance was also present in the code of \cite{lm_debugger}.

\paragraph{Experimental Design} 
We use GPT-2 medium 
and run it over 60 examples from IMDB (25,000 train, 25,000 test examples; \cite{imdb}).\footnote{Note that IMDB was designed for sentiment analysis and we use it here as a general-purpose corpus.} This provides us with a dynamically-computed hidden state $h$ for every token and at the output of every layer. For the projection $\hat{h} \in \mathbb{R}^e$ of each such hidden state, we take the projections of the $m$ most active parameter vectors $\{\hat{x}_i\}_{i=1}^{m}$ in the layer that computed $h$ and check if they cover the dominant vocabulary items of $\hat{h}$ in embedding space. Specifically, let $\texttt{top-k}(w E)$ be the $k$ vocabulary items with the largest logits in embedding space for a vector $w \in \mathbb{R}^d$. We compute: 
\begin{equation*}
R_k(\hat{x}_1, ..., \hat{x}_m, \hat{h}) = \frac{| \texttt{top-k}(\hat{h})\cap \bigcup_{i=1}^{m} \texttt{top-k}(\hat{x}_i)|}{k},
\end{equation*}
to capture if activated parameter vectors cover the main vocabulary items corresponding to the hidden state.

We find the $m$ most active parameter vectors separately for FF keys ($K$), FF values ($V$), attention value \textit{subheads} ($\WV$) (see \S\ref{sec:theory}), and attention output subheads ($\WO$), where the activation of each parameter vector is determined by the vector's ``coefficient'' as follows. For a FF key-value pair $(k, v)$ the coefficient is $\sigma(q^\top k)$, where $q \in \mathbb{R}^d$ is an input to the FF module, and $\sigma$ is the FF non-linearity. For attention, value-output subhead pairs $(v,o)$ the coefficient is $x^\top v$, where $x$ is the input to this component (for attention head $i$, the input is one of the rows of $A^i X$, see  Eq.~\ref{eq:wvo_eq}).



\paragraph{Results and Discussion} Figure \ref{fig:kv_vs_states} presents
the $R_k$ score averaged across tokens per layer.
As a baseline, we compare $R_k$ of the activated vectors $\{\hat{x}_i\}_{i=1}^{m}$ of the \textit{correctly-aligned} hidden state $\hat{h}$ at the output of the relevant layer (blue bars) against the $R_k$ when \textit{randomly sampling} $\hat{h}_\text{rand}$ from all the hidden states (orange bars). We conclude that representations in embedding space induced by activated parameter vector mirror, at least to some extent, the representations of the hidden states themselves. Appendix~\S\ref{appendix:weights_vs_logits} shows a variant of this experiment, where we compare activated parameters throughout GPT-2 medium's layers to the \textit{last} hidden state, which produces the logits used for prediction.

\subsection{Interpretation of Fine-tuned Models}
\label{subsec:interpret_finetune}

We now show that we can interpret the \emph{changes} a model goes through during fine-tuning through the lens of embedding space.
We fine-tune the top-3 layers of the  12-layer GPT-2 base (117M parameters) with a sequence classification head on IMDB sentiment analysis (binary classification)
and compute the difference between the original parameters and the fine-tuned model. 
We then project the difference of parameter vectors into embedding space and test if the change is interpretable w.r.t. sentiment analysis. 

Appendix~\ref{appendix:finetune_examples} shows examples of projected differences randomly sampled from the fine-tuned layers. Frequently, the difference or its negation is projected to nouns, adjectives, and adverbs that express sentiment for a movie, such as \emph{`amazing'}, \emph{`masterpiece'}, \emph{`incompetence'}, etc. This shows that the differences are indeed projected into vocabulary items that characterize movie reviews' sentiments. This behavior is present across $\WQ, \WK, \WV, K$, but not $V$ and $\WO$, which curiously are the parameters added to the residual stream and not the ones that react to the input directly. 
\section{Aligning Models in Embedding Space}
\label{sec:applications}
The assumption Transformers operate in embedding space leads to an exciting possibility -- we can relate \emph{different} models to one another so long as they share the vocabulary and tokenizer. In \S\ref{subsec:layer_alignment}, we show that we can align the layers of BERT models trained with different random seeds. In \S\ref{subsec:stiching}, we show the embedding space can be leveraged to ``stitch'' the parameters of a fine-tuned model to a model that was not fine-tuned.

\subsection{Layer Alignment}
\label{subsec:layer_alignment}

\begin{figure*}
    \centering
    \vspace*{-0.5cm}
    \setlength{\belowcaptionskip}{-10pt}
    \hspace*{-0.5cm}
    \includegraphics[width=8.4cm]{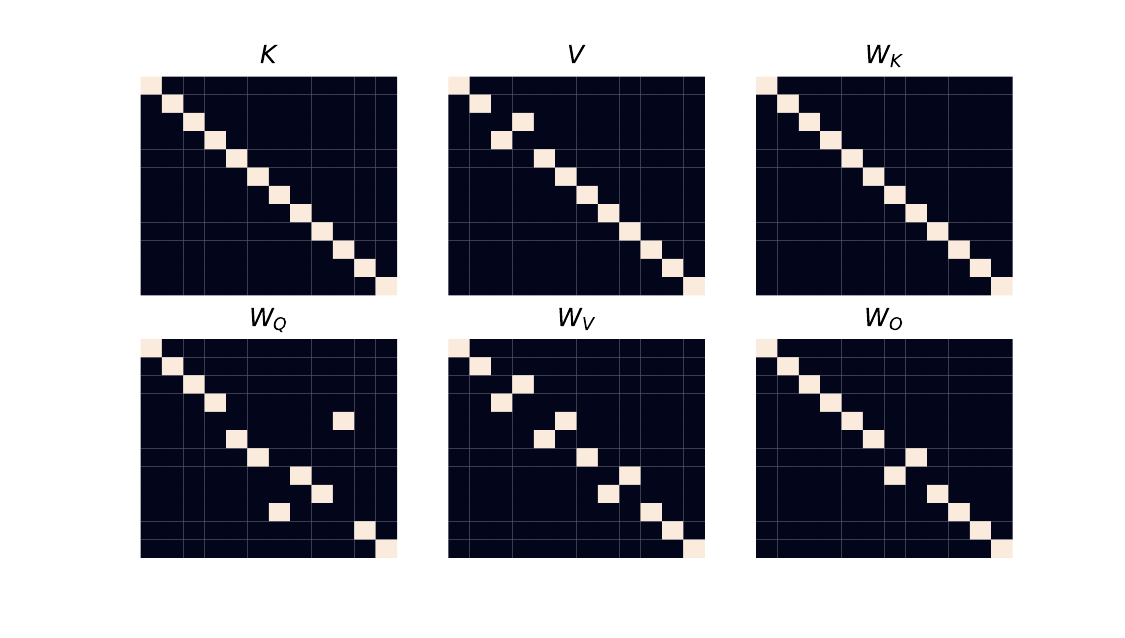} 
    \hspace*{-0.9cm}
    \includegraphics[width=8.4cm]{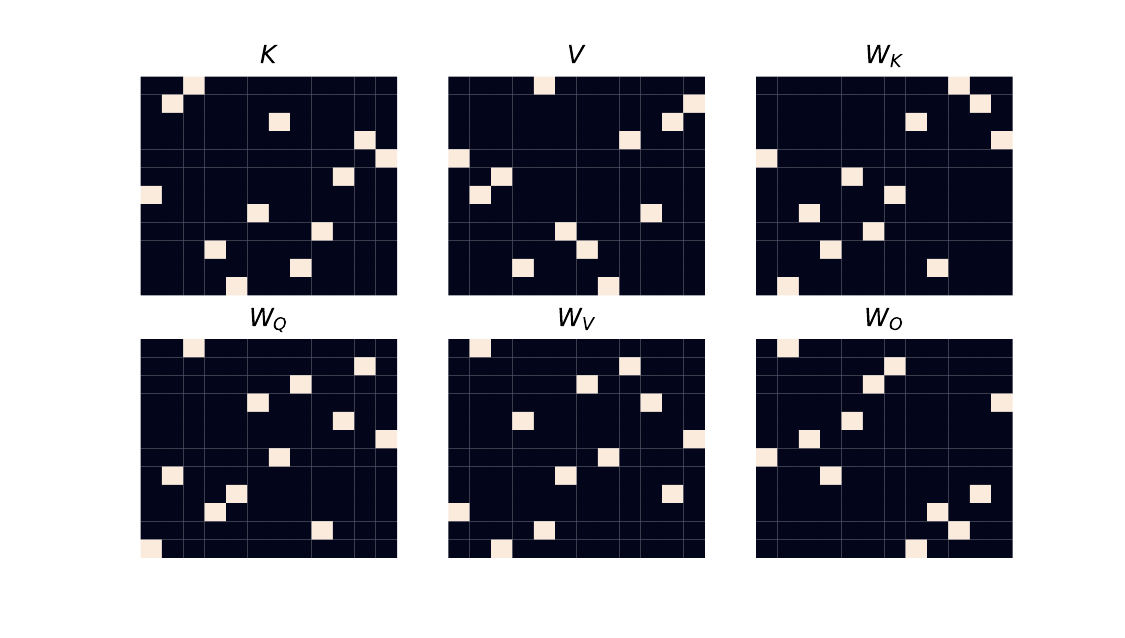}
    \vspace*{-0.45cm}
    \caption{Left: Aligning \textit{in embedding space} the layers of two different BERT models initialized from different random seeds for all parameter groups. Layers that have the same index tend to align with one another. Right: Alignment in feature space leads to unintelligible patterns.}
    \label{fig:param_alignment}
\end{figure*}

\paragraph{Experimental Design} 
Taking our approach to the extreme, the embedding space is a universal space, which depends only on the tokenizer, in which Transformer parameters and hidden states reside. Thus, we can align parameter vectors from different models in this space and compare them even if they come from different models, as long as they share a vocabulary. 

To demonstrate this, we use MultiBERTs (\citep{sellam2022multiberts}; License: Apache-2.0), which contains 25 different instantiations of BERT-base (110M parameters) initialized from different random seeds.\footnote{Estimated compute costs: around 1728 TPU-hours for each pre-training run, and around 208 GPU-hours plus 8 TPU-hours for associated fine-tuning experiments.} We take parameters from two MultiBERT seeds and compute the correlation between their projections to embedding space. For example, let $V_A, V_B$ be the FF values of models $A$ and $B$. We can project the values into embedding space: $V_A E_A, V_B E_B$, where $E_A, E_B$ are the respective embedding matrices, and compute Pearson correlation between projected values. This produces a similarity matrix $\Tilde{\mathcal{S}} \in \mathbb{R}^{|V_A| \times |V_B|}$, where each entry is the correlation coefficient between  projected values from the two models. 
We bin $\Tilde{\mathcal{S}}$ by layer pairs and average the absolute value of the scores in each bin (different models might encode the same information in different directions, so we use absolute value) to produce a matrix $\mathcal{S} \in \mathbb{R}^{L \times L}$, where $L$ is the number of layers -- that is, the average (absolute) correlation between vectors that come from layer $\ell_A$ in model A and layer $\ell_B$ in Model B is registered in entry $(\ell_A, \ell_B)$ of  $\mathcal{S}$. 

Last, to obtain a one-to-one layer alignment, we use the Hungarian algorithm \citep{hungarian}, which assigns exactly one layer from the first model to a layer from the second model. The algorithm's objective is to maximize, given a similarity matrix $\mathcal{S}$, the sum of scores of the chosen pairs, such that each index in one model is matched with exactly one index in the other. We repeat this for all parameter groups ($\WQ, \WK, \WV, \WO, K$).

\paragraph{Results and Discussion} 
Figure~\ref{fig:param_alignment} (left) shows the resulting alignment. Clearly, parameters from a certain layer in model $A$ tend to align to the same layer in model $B$ across all parameter groups.
This suggests that different layers from different models that were trained separately (but with the same training objective and data) serve a similar function. As further evidence, we show that if not projected, the matching appears absolutely random in Figure~\S\ref{fig:param_alignment} (right). We show the same results for other seed pairs as well in Appendix~\ref{appendix:alignment_extra}.


\subsection{Zero-shot Stitching}
\label{subsec:stiching}

Model stitching \citep{lenc2015, csiszrik2021, bansal2021} is a relatively under-explored feature of neural networks, particularly in NLP. The idea is that different models, even with different architectures, can learn representations that can be aligned through a \textit{linear} transformation, termed \emph{stitching}. Representations correspond to hidden states, and thus one can learn a transformation matrix from one model's hidden states to an equivalent hidden state in the other model. Here, we show that going through embedding space one can align the hidden states of two models, i.e., stitch, \textit{without training}. 

Given two models, we want to find a linear stitching transformation to align their representation spaces. According to our theory, given a hidden state $v\in \mathbb{R}^{d_1}$ from model $A$, we can project it to the embedding space as $v E_A$, where $E_A$ is its embedding matrix. Then, we can re-project to the feature space of model B, with $E_B^{+} \in \mathbb{R}^{e \times d_2}$, where $E_B^{+}$ is the Penrose-Moore pseudo-inverse of the embedding matrix $E_B$.\footnote{Since we are not interested in interpretation we use an exact right-inverse and not the transpose.}
This transformation can be expressed as multiplication with the kernel $K_{AB} := E_A E_B^{+} \in \mathbb{R}^{d_1 \times d_2}$. 
We employ the above approach to take 
representations of a fine-tuned classifier, $A$, and stitch them on top of a model $B$ that was only pretrained, to obtain a new classifier based on $B$. 

\begin{figure}
    \centering
    \includegraphics[width=8.5cm]{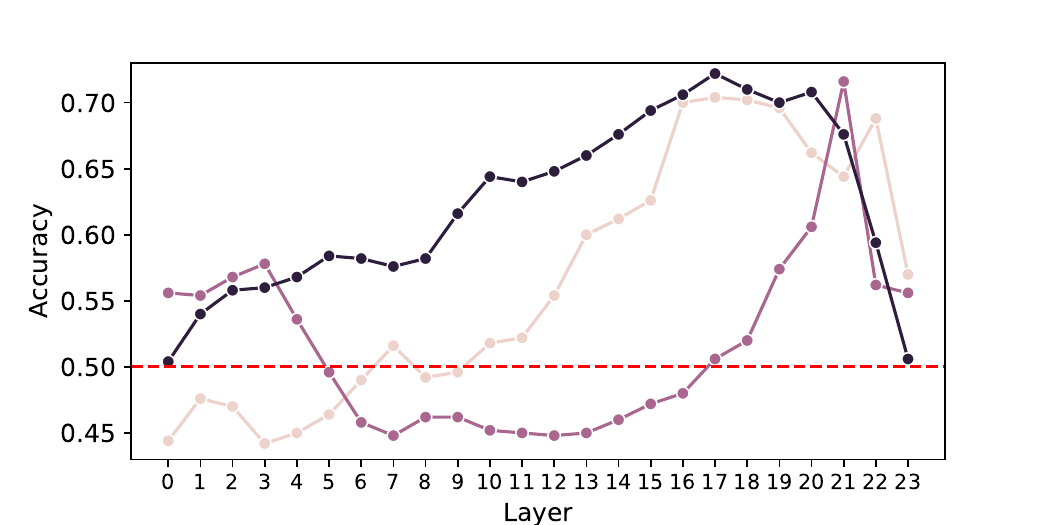}
    \caption{Accuracy on the IMDB evaluation set. We ran stitching randomly 11 times and obtained 3 models with higher than random accuracy when stitching over top layers. Dashed red line indicates random performance.}
    \label{fig:stitching_acc}
    \vspace*{-0.5cm}
\end{figure}
\paragraph{Experimental Design}
We use the 24-layer GPT-2 medium as model $A$ and 12-layer GPT-2 base model trained in \S\ref{subsec:interpret_finetune} as model $B$. We fine-tune the last three layers of model $B$ on IMDB, as explained in \S\ref{subsec:interpret_finetune}.
Stitching is simple and is performed as follows. Given the sequence of $N$ hidden states $H_A^\ell \in \mathbb{R}^{N \times d_1}$ at the output of layer $\ell$ of model $A$ ($\ell$ is a hyperparameter), we apply the \emph{stitching layer}, which multiplies the hidden states with the kernel, computing $H_A^\ell K_{AB}$. This results in hidden states $H_B \in \mathbb{R}^{N \times d_2}$, used as input to the three fine-tuned layers from $B$.


\paragraph{Results and Discussion}
Stitching produces models with accuracies that are higher than random on IMDB evaluation set, but not consistently. Figure~\ref{fig:stitching_acc} shows the accuracy of stitched models against the
layer index from model $A$ over which stitching is performed. Out of 11 random seeds, three models obtained accuracy that is significantly higher than the baseline 50\% accuracy, reaching an accuracy of roughly 70\%, when stitching is done over the top layers.

\section{Related Work}
Interpreting Transformers is a broad area of research that has attracted much attention in recent years. 
A large body of work has focused on analyzing hidden representations, mostly through probing  \citep{adi_probing, shi_probes,  rediscovers, primer_bertology}.
\cite{voita_evolution} used statistical tools to analyze the evolution of hidden representations throughout layers. Recently, \cite{mickus2022dissect} proposed to decompose the hidden representations into the contributions of different Transformer components. Unlike these works, we interpret parameters rather than the hidden representations.

Another substantial effort has been to interpret specific network components. Previous work analyzed single neurons \citep{grain_sand, individual_neurons}, attention heads \citep{what_does_bert, voita_analyzing}, and feedforward values \citep{kv_memories, knowledge_neurons, elhage2022solu}.  
While these works mostly rely on input-dependent neuron activations, we inspect ``static'' model parameters, and provide a comprehensive view of all Transformer components.

Our work is most related to efforts to interpret specific groups of Transformer parameters. \cite{cammarata2020thread} made observations about the interpretability of weights of neural networks. \cite{anthropic} analyzed 2-layer attention networks. We extend their analysis to multi-layer pre-trained Transformer models. \cite{kv_memories, lm_debugger, geva2022} interpreted feedforward values in embedding space. We coalesce these lines of work and offer a unified interpretation framework for Transformers in embedding space.
\section{Discussion}
While our work has limitations (see \S\ref{sec:limitations}), we think the benefits of our work overshadow its limitations. We provide a simple approach and a new set of tools to interpret Transformer models and compare them. The realm of input-independent interpretation methods is still nascent and it might provide a fresh perspective on the internals of the Transformer, one that allows to glance intrinsic properties of specific parameters, disentangling their dependence on the input. Moreover, many models are prohibitively large for practitioners to run. Our method requires only a fraction of the compute and memory requirements, and allows interpreting a single parameter in isolation.

Importantly, our framework allows us to view parameters from different models as residents of a canonical embedding space, where they can be compared in model-agnostic fashion. This has interesting implications. We demonstrate two consequences of this observation (model alignment and stitching) and argue future work can yield many more use cases. 

\section{Limitations}
\label{sec:limitations}
Our work has a few limitations that we care to highlight.
First, it focuses on interpreting models through the vocabulary lens. While we have shown evidence for this, it does not preclude other factors from being involved. Second, we used $E'=E^\top$, but future research may find variants of $E$ that improve performance. Additionally, most of the work focused on GPT-2. This is due to shortcomings in the current state of our framework, as well as for clear presentation. We believe non-linearities in language modeling are resolvable, as is indicated in the experiment with BERT. 

In terms of potential bias in the framework, some parameters might consider terms related to each due to stereotypes learned from the corpus.

\bibliographystyle{abbrvnat}
\bibliography{references}

\appendix
\clearpage
\onecolumn
\section{Rethinking Interpretation}
\label{appendix:echoice}

\begin{figure*}[ht]
\centering
\includegraphics[width=8cm]{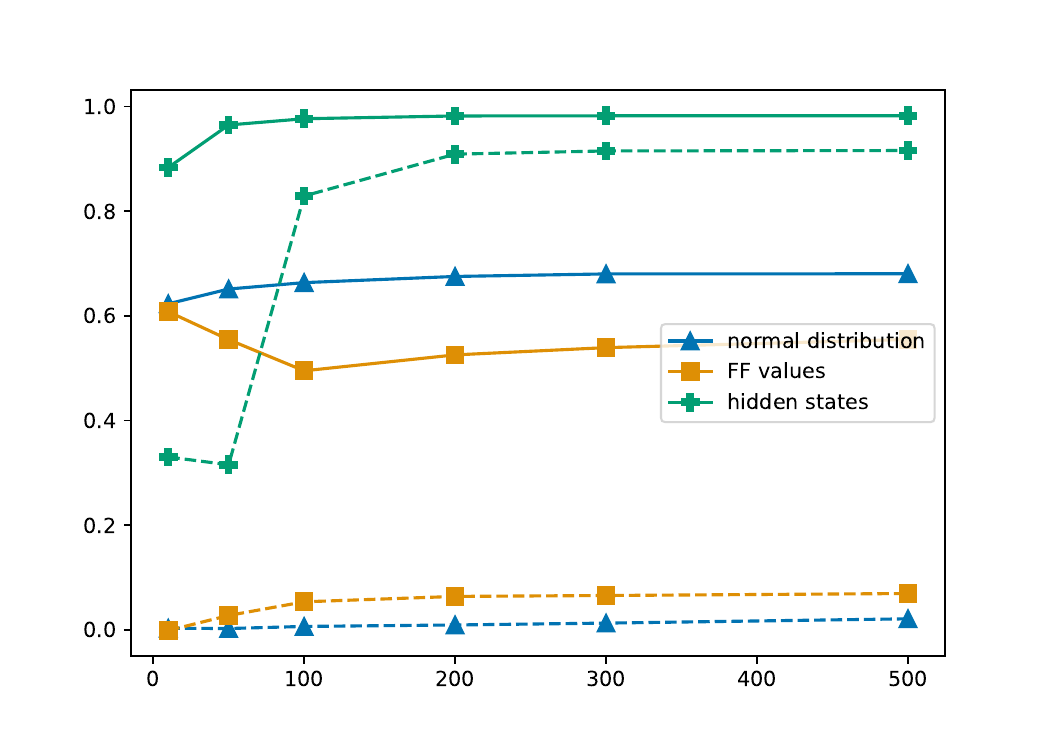}
\hspace*{-1cm}
\includegraphics[width=8cm]{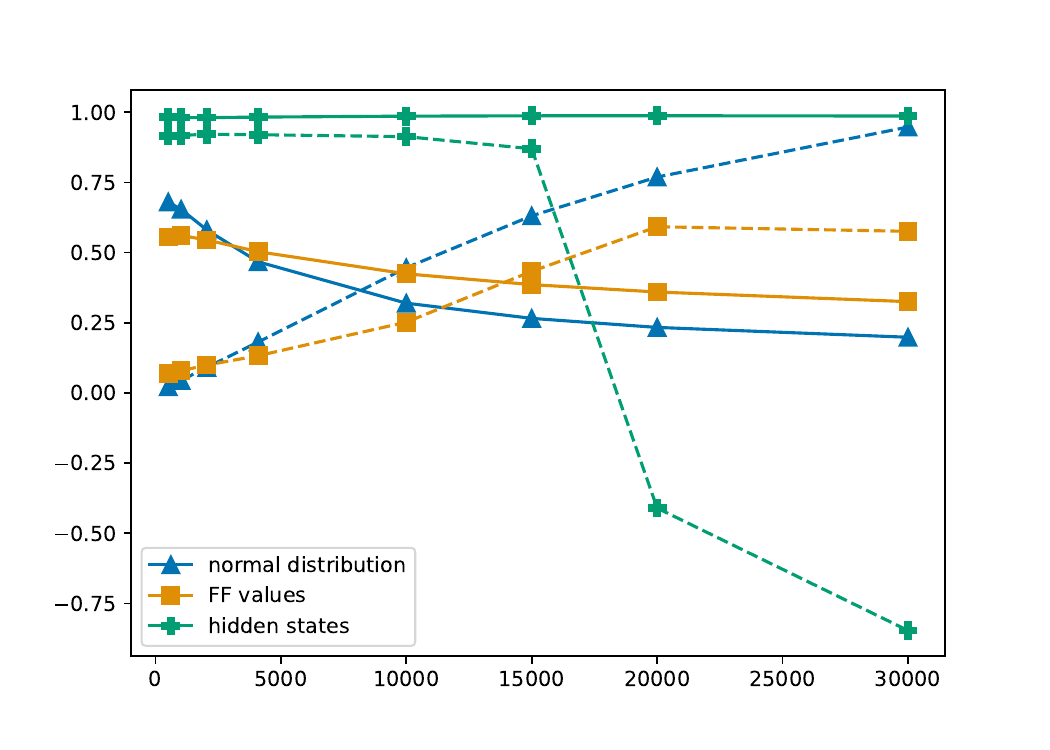}

\includegraphics[width=8cm]{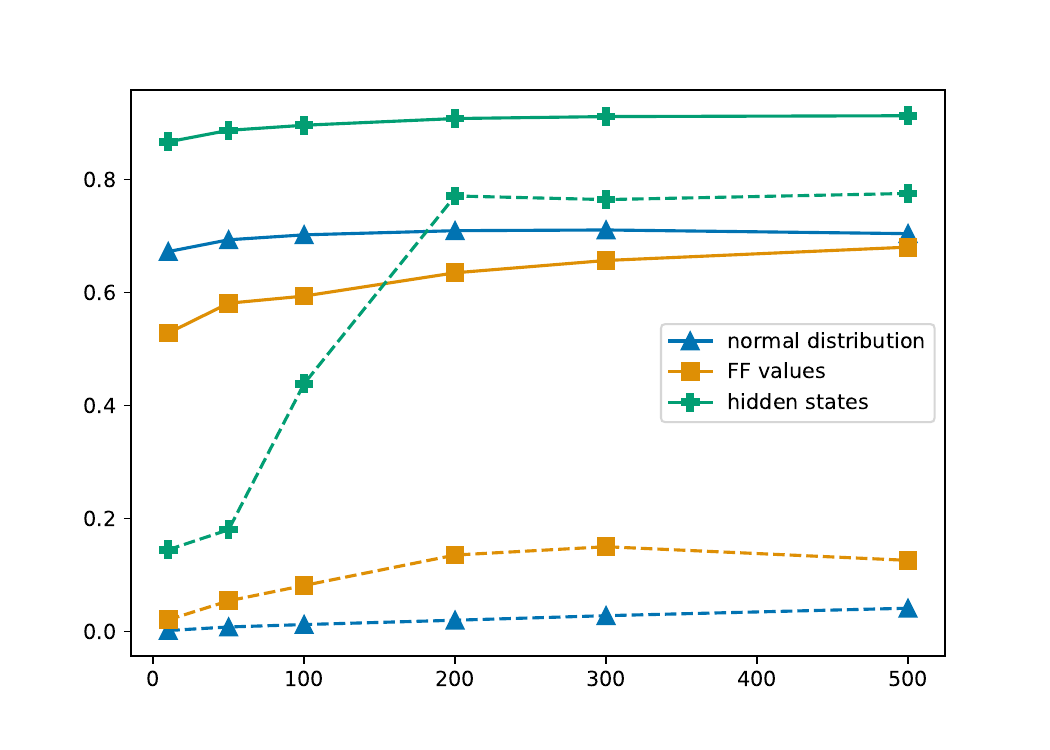}
\hspace*{-1cm}
\includegraphics[width=8cm]{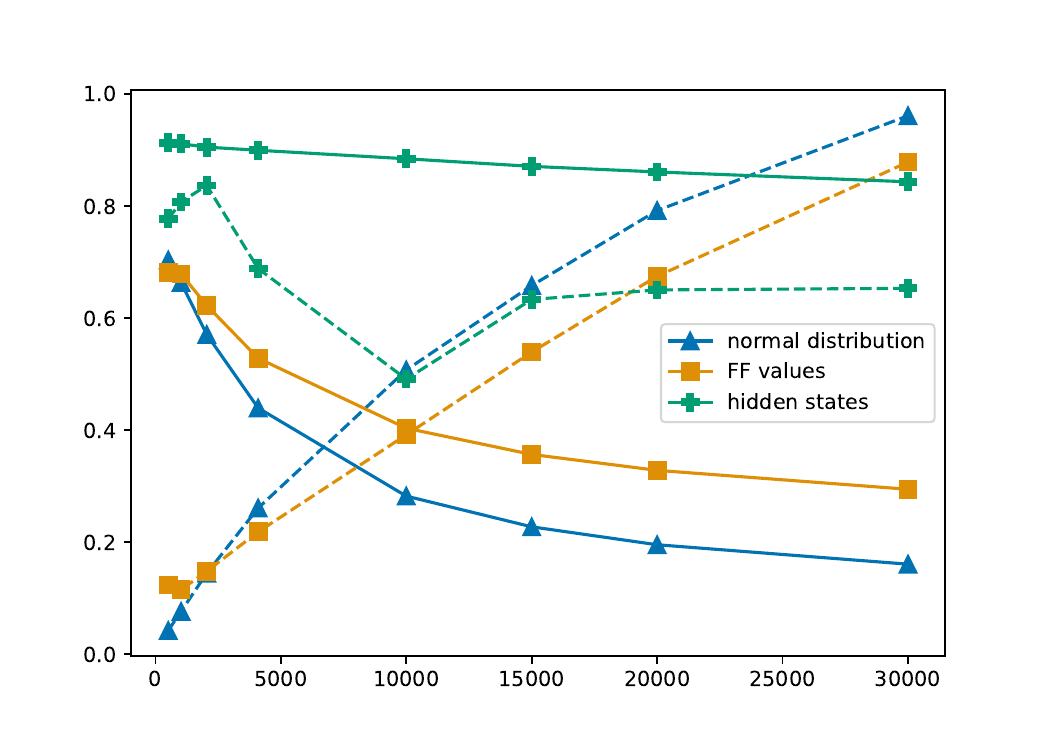}

\includegraphics[width=8cm]{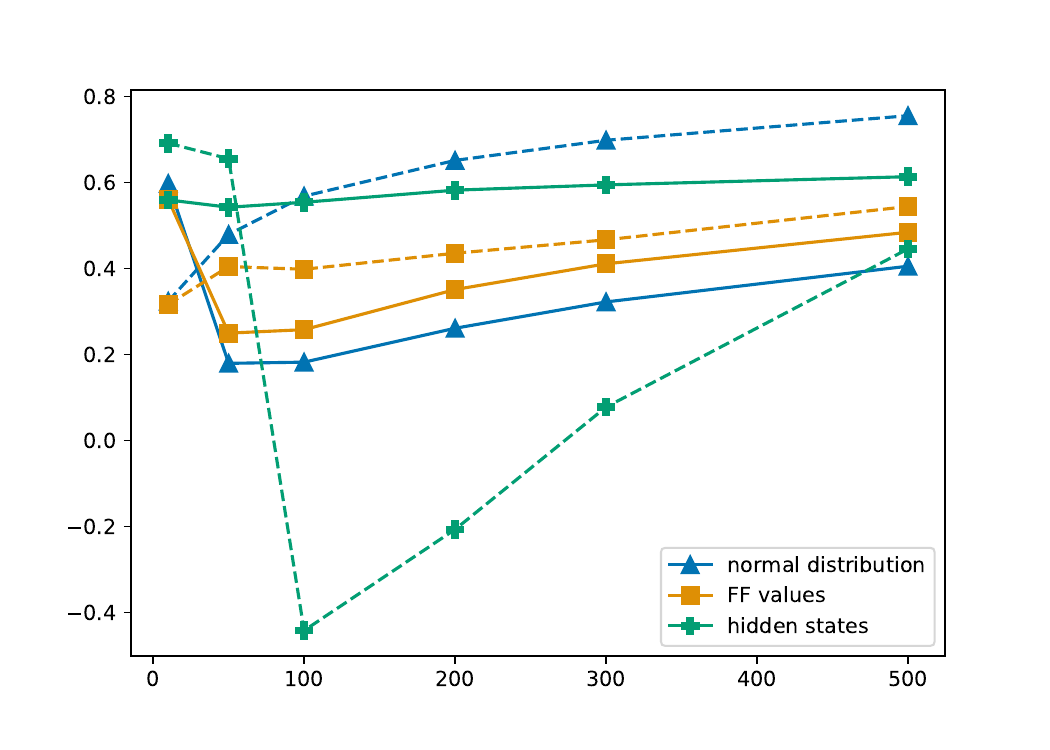}
\hspace*{-1cm}
\includegraphics[width=8cm]{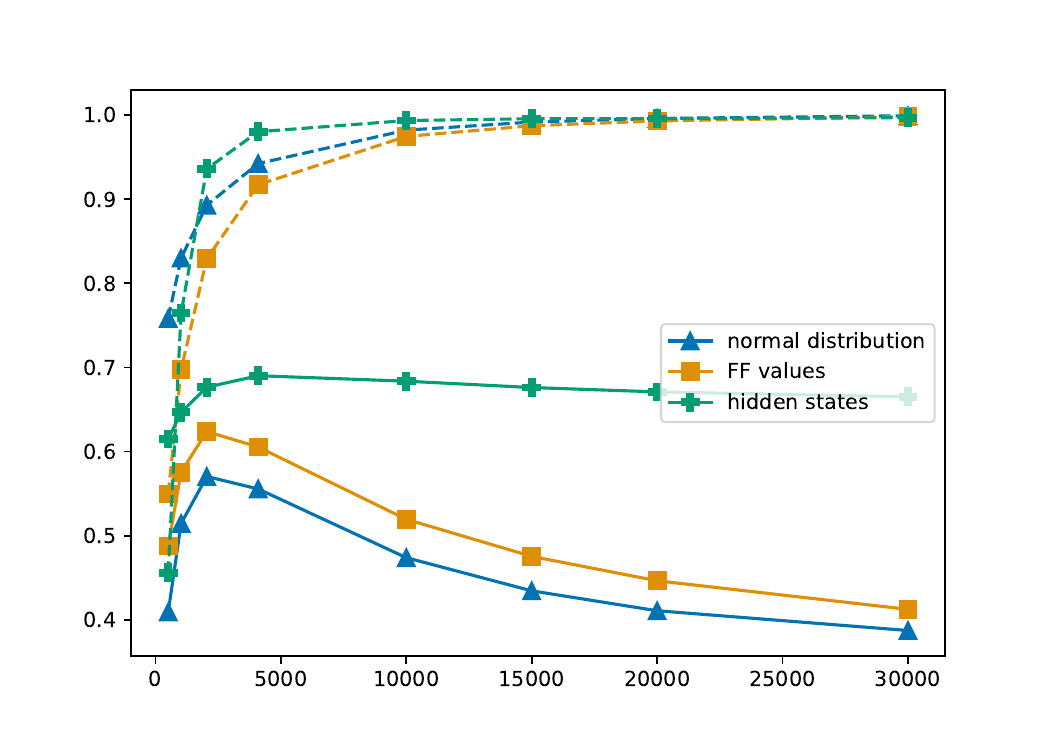}
\caption{Each row represents a model in the following order from top to bottom: GPT-2 base, GPT-2 medium, GPT-2 large. \emph{Left}: The \texttt{keep-k} inverse scores for three distributions: normal distribution, hidden states, and FF values, for $k \in \{10, 50, 100, 200, 300, 500\}$. \emph{Right}: for $k \in \{10, 50, 100, 200, 300, 500\}$.}
\label{fig:keep_k}
\end{figure*}

The process of interpreting a vector $v$ in \cite{geva2022} proceeds in two steps: first the \textit{projection} of the vector to the embedding space ($v E$); then, we use the list of the tokens that were assigned the largest values in the projected vector, i.e.: $\topk(v E)$, as the \textit{interpretation} of the projected vector. This is reasonable since (a) the most activated coordinates contribute the most when added to the residual stream, and (b) this matches how we eventually decode: we project to the embedding space and consider the top-1 token (or one of the few top tokens, when using beam search).

In this work, we interpret inner products and matrix multiplications in the embedding space: given two vectors $x, y \in \mathbb{R}^d$, their inner product $x^\top y$ can be considered in the embedding space by multiplying with $E$ and then by one of its right inverses (e.g., its pseudo-inverse $E^{+}$ \citep{moore1920, bjerhammar1951, penrose1955}): $x^\top y = x^\top E E^{+} y = (x^\top E) (E^{+} y)$. Assume $x E$ is interpretable in the embedding space, crudely meaning that it represents logits over vocabulary items. We expect $y$, which interacts with $x$, to also be interpretable in the embedding space. Consequently, we would like to take $E^{+}y$ to be the projection of $y$. However, this projection does not take into account the subsequent interpretation using top-$k$. The projected vector $E^{+} y$ might be harder to interpret in terms of its most activated tokens. To alleviate this problem, we need a different ``inverse'' matrix $E'$ that works well when considering the top-$k$ operation. Formally, we want an $E'$ with the following ``robustness'' guarantee:  $\texttt{keep-k}(x^\top E) \texttt{keep-k}(E' y) \approx x^\top y$, where $\texttt{keep-k}(v)$ is equal to $v$ for coordinates whose absolute value is in the top-$k$, and zero elsewhere.

This is a stronger notion of inverse -- not only is $EE' \approx I$, but even when truncating the vector in the embedding space we can still reconstruct it with $E'$. 

We claim that $E^\top$ is a decent instantiation of $E'$ and provide some empirical evidence. 
While a substantive line of work \citep{ethayarajh, gao2018representation, wang2020, isoscore} has shown that embedding matrices are not isotropic (an isotropic matrix $E$ has to satisfy $E E^\top = \alpha I$ for some scalar $\alpha$), we show that it is isotropic enough to make $E^\top$ a legitimate compromise. We randomly sample 300 vectors drawn from the normal distribution $\mathcal{N}(0, 1)$, and compute for every pair $x, y$ the cosine similarity between $x^\top y$ and $\texttt{keep-k}(x^\top E) \texttt{keep-k}(E'y)$ for $k=1000$, and then average over all pairs. We repeat this for $E' \in \{E^{+}, E^\top\}$ and obtain a score of $0.10$ for $E^{+}$, and $0.83$ for $E^\top$, showing the $E^\top$ is better under when using top-$k$. More globally, we compare $E' \in \{E^{+}, E^\top\}$ for $k \in \{10, 50, 100, 200, 300, 500\}$ with three distributions:
\begin{itemize}[noitemsep] 
    \item[-] $x, y$ drawn from the normal $\mathcal{N}(0, 1)$ distribution
    \item[-] $x, y$ chosen randomly from the FF values
    \item[-] $x, y$ drawn from hidden states along Transformer computations.
\end{itemize}
In Figure~\ref{fig:keep_k} we show the results, where dashed lines represent $E^{+}$ and solid lines represent $E^\top$. The middle row shows the plots for GPT-2 medium, which is the main concern of this paper. For small values of $k$ (which are more appropriate for interpretation), $E^\top$ is superior to $E^{+}$ across all distributions. Interestingly, the hidden state distribution is the only distribution where $E^{+}$ has similar performance to $E^\top$. Curiously, when looking at higher values of $k$ the trend is reversed ($k = \{512, 1024, 2048, 4096, 10000, 15000, 20000, 30000\}$) - see Figure~\ref{fig:keep_k} (Right).

This settles the deviation from findings showing embedding matrices are not isotropic, as we see that indeed as $k$ grows, $E^\top$ becomes an increasingly bad approximate right-inverse of the embedding matrix.
The only distribution that keeps high performance with $E^\top$ is the hidden state distribution, which is an interesting direction for future investigation. 

For completeness, we provide the same analysis for GPT-2 base and large in Figure~\ref{fig:keep_k}. We can see that GPT-2 base gives similar conclusions. GPT-2 large, however, seems to show a violent zigzag movement for $E^{+}$ but for most values it seems to be superior to $E^\top$. It is however probably best to use $E^\top$ since it is more predictable. This zigzag behavior is very counter-intuitive and we leave it for future work to decipher.


\section{Additional Material}
\label{appendix:additional}
\subsection{Corresponding Parameter Pairs are Related}
\label{appendix:related_pairs}
\begin{figure*}[t]
    \centering
    \hspace*{-1.5cm}
    \includegraphics[width=18cm]{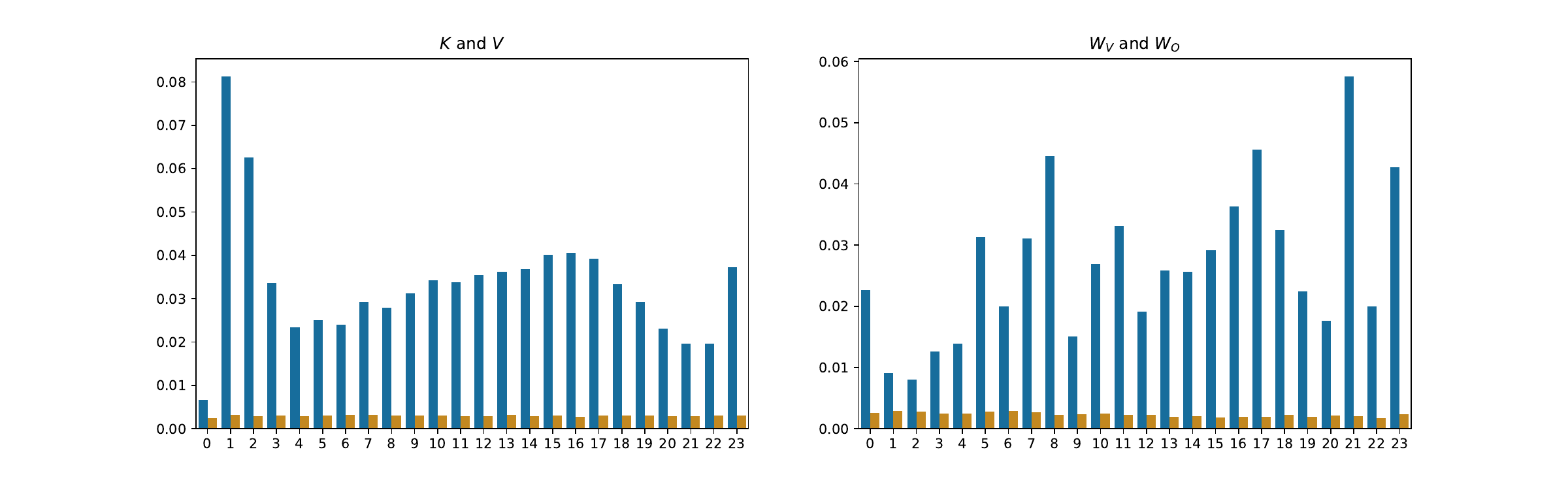}
        \includegraphics[width=8cm]{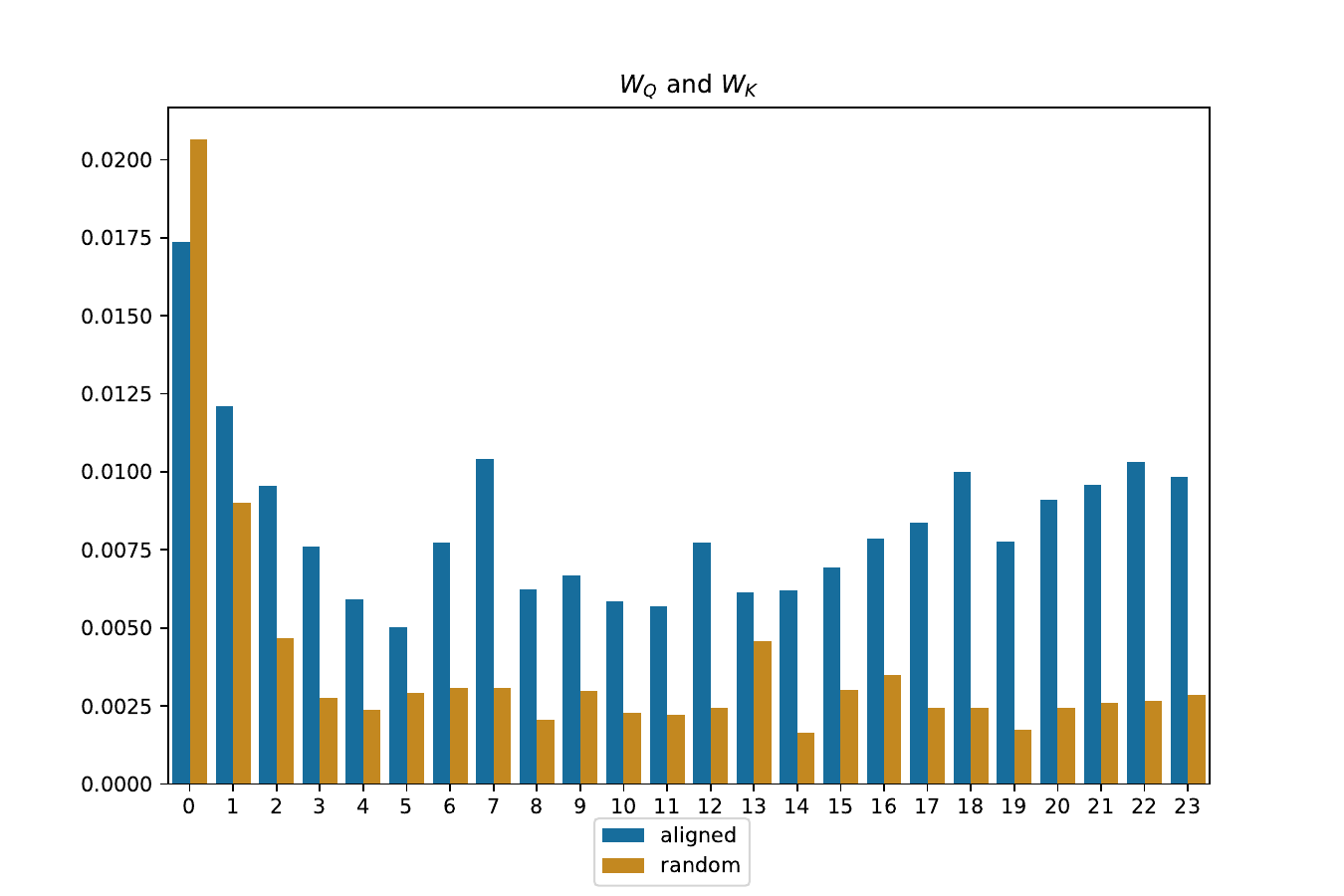}
    \caption{Average $\textrm{Sim}_{k}(\hat{x}, \hat{y})$ for $k = 100$ by layer, where blue is when matching pairs are aligned, and orange is when pairs are shuffled within the layer. Top Left: FF keys and FF values. Top Right: The subheads of $W_O$ and $W_V$. Bottom: The subheads of $W_Q$ and $W_K$.}
    \label{fig:related_pairs}
\end{figure*}
We define the following metric applying on vectors \textit{after projecting} them into the embedding space: 
\[
    \textrm{Sim}_{k}(\hat{x}, \hat{y}) = \frac{| \texttt{top-k}(\hat{x}) \cap \texttt{top-k}(\hat{y})|}{| \texttt{top-k}(\hat{x}) \cup \texttt{top-k}(\hat{y})|}
\]
where $\texttt{top-k}(v)$ is the set of $k$ top activated indices in the vector $v$ (which correspond to tokens in the embedding space).
This metric is the Jaccard index \citep{jaccard} applied to the top-$k$ tokens from each vector. In Figure \ref{fig:related_pairs}, Left, we demonstrate that FF key vectors and their corresponding value vectors are more similar (in embedding space) than two random key and value vectors. In Figure \ref{fig:related_pairs}, Right, we show a similar result for attention value and output vectors.  In Figure \ref{fig:related_pairs}, Bottom, the same analysis is done for attention query and key vectors. 
This shows that there is a much higher-than-chance relation between corresponding FF keys and values (and the same for attention values and outputs). 


\subsection{Final Prediction and Parameters}
\label{appendix:weights_vs_logits}
\begin{figure*}
    \hspace*{-1.5cm}
    \includegraphics[width=18cm]{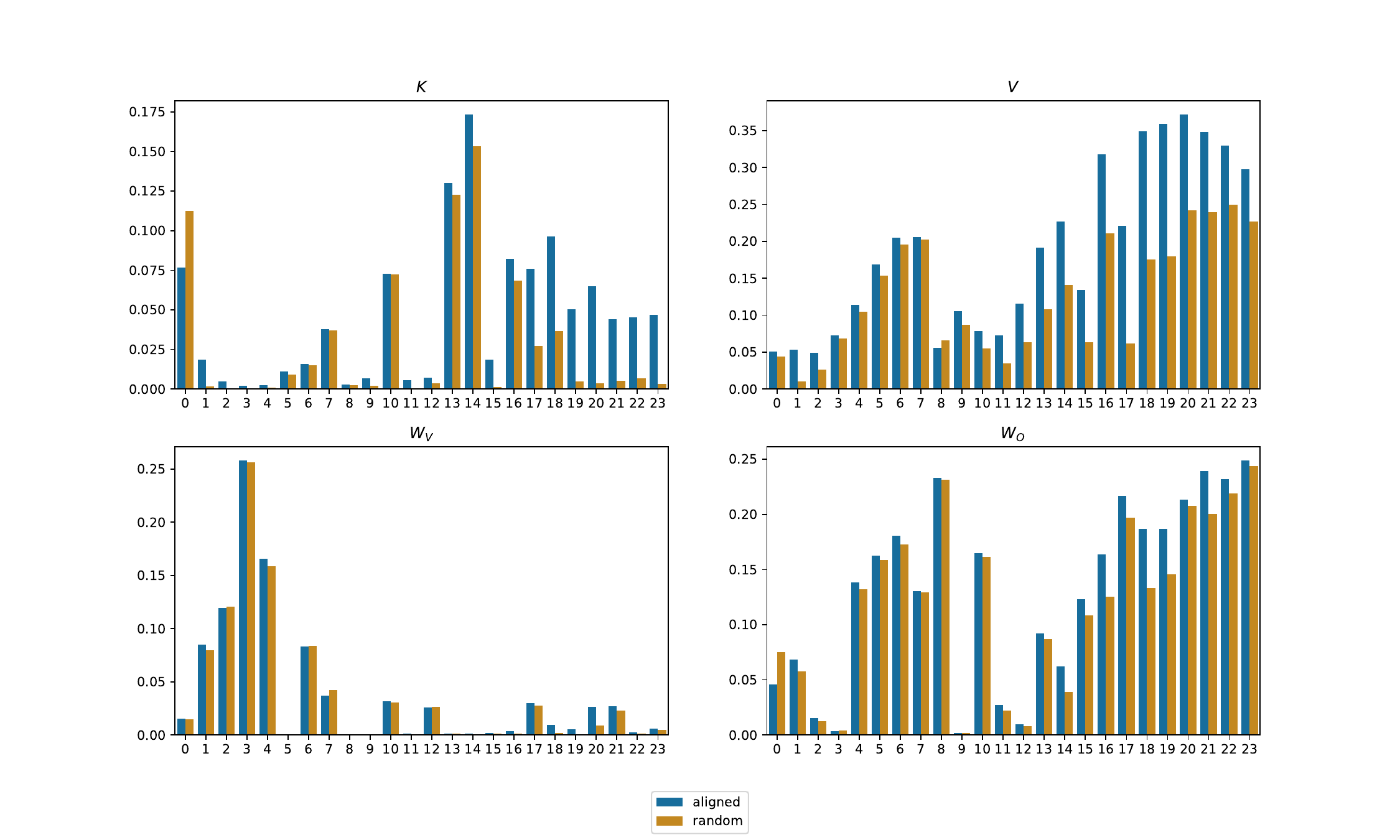}
    \caption{Left: Average $R_k$ score ($k=100$) across tokens per layer for activated parameter vectors against both the aligned hidden state $\hat{h}$ at the output of the \emph{final} layer and a randomly sampled hidden state $\hat{h}_\text{rand}$. Parameters are FF keys (top-left), FF values (top-right), attention values (bottom-left), and attention outputs (bottom-right).\vspace*{-10pt}
    }
    \label{fig:weights_vs_last}
\end{figure*}

We show that the final prediction of the model is correlated in embedding space with the most activated parameters from each layer. This implies that these objects are germane to the analysis of the final prediction in the embedding space, which in turn suggests that the embedding space is a viable choice for interpreting these vectors. 
Figure \ref{fig:weights_vs_last} shows that just like \S\ref{subsec:weights_vs_states}, correspondence is better when hidden states are not randomized, suggesting their parameter interpretations have an impact on the final prediction.

\clearpage
{


\subsection{Parameter Alignment Plots for Additional Model Pairs}
\label{appendix:alignment_extra}

Alignment in embedding space of layers of pairs of BERT models trained with different random seeds for additional model pairs.

    \subsubsection*{Seed 1 VS Seed 2}

    \includegraphics[width=7.5cm]{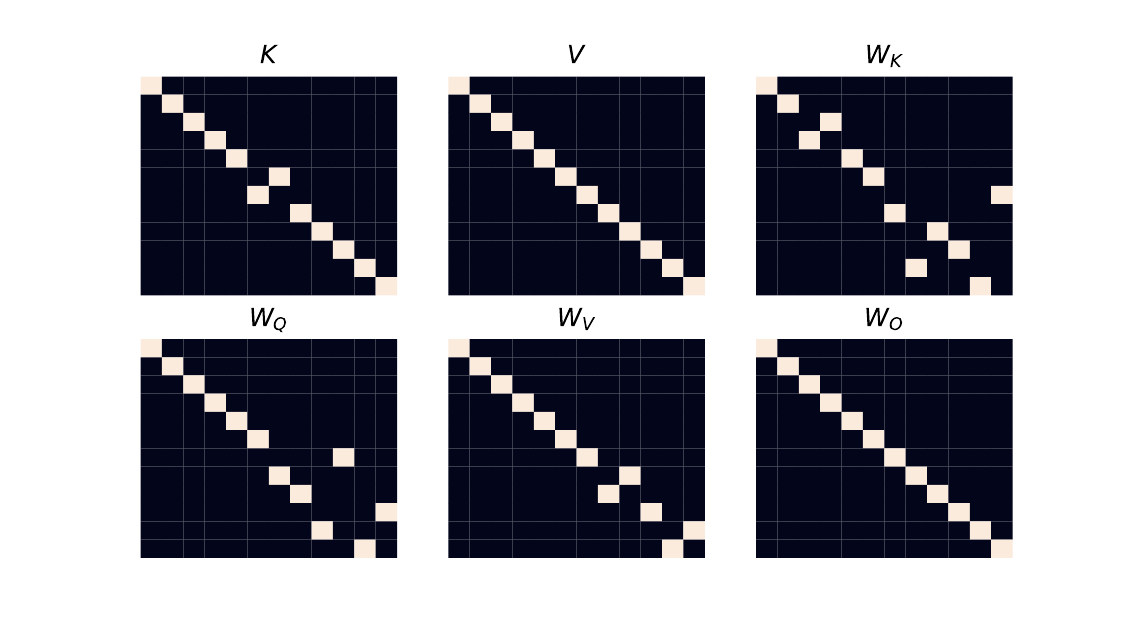} 
    \includegraphics[width=7.5cm]{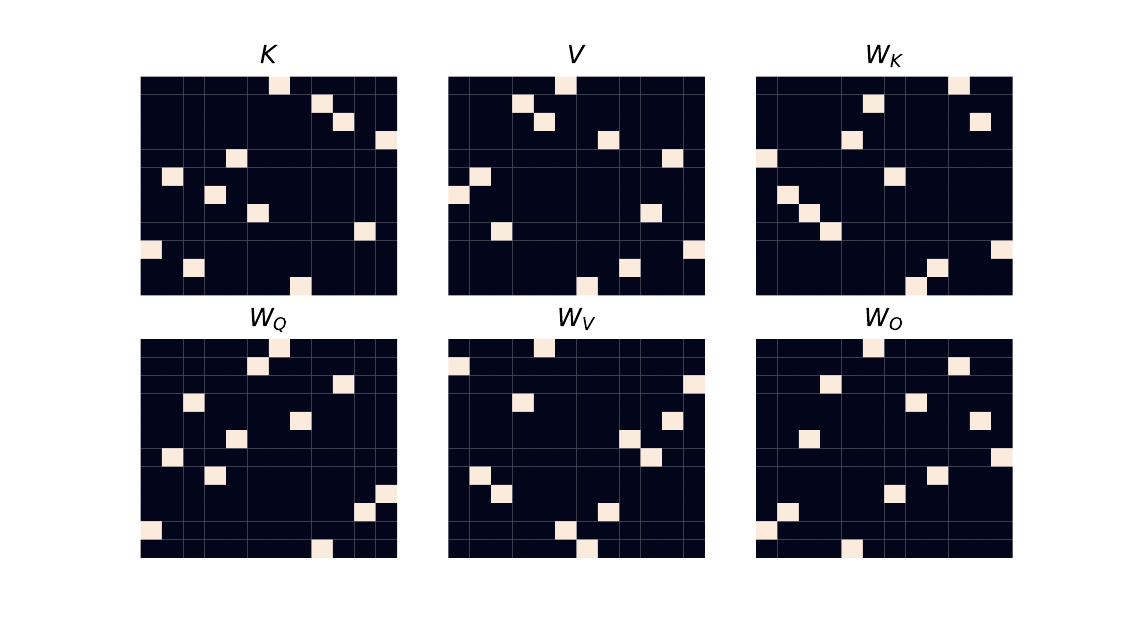}
    
    \subsubsection*{Seed 2 VS Seed 3}
    \includegraphics[width=7.5cm]{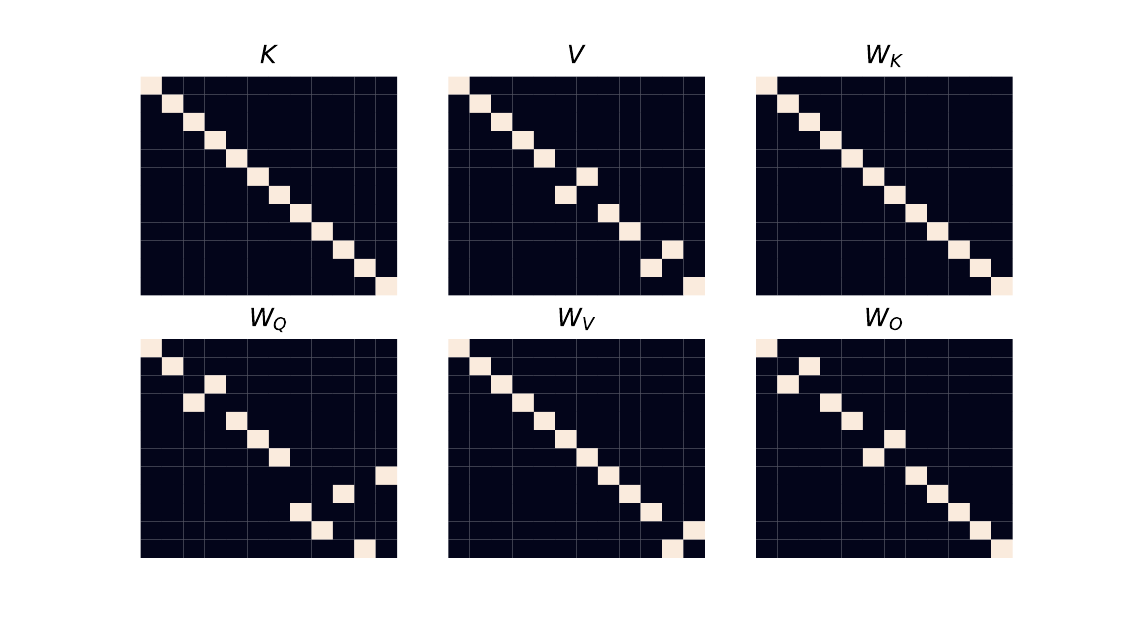} 
    \includegraphics[width=7.5cm]{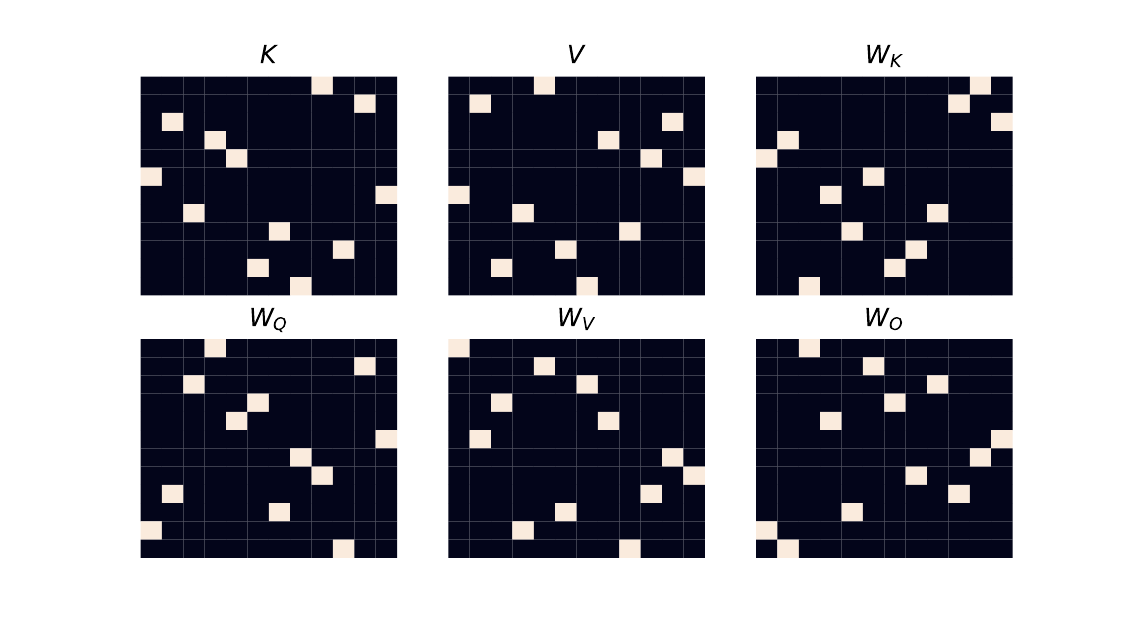}
        
    \subsubsection*{Seed 3 VS Seed 4}
    \includegraphics[width=7.5cm]{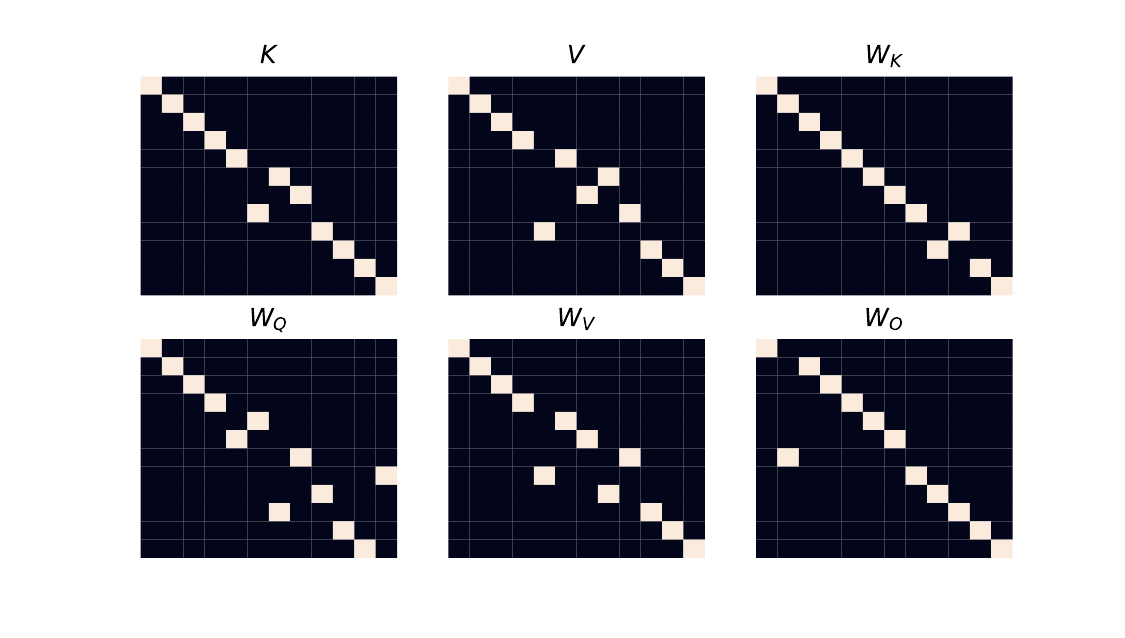} 
    \includegraphics[width=7.5cm]{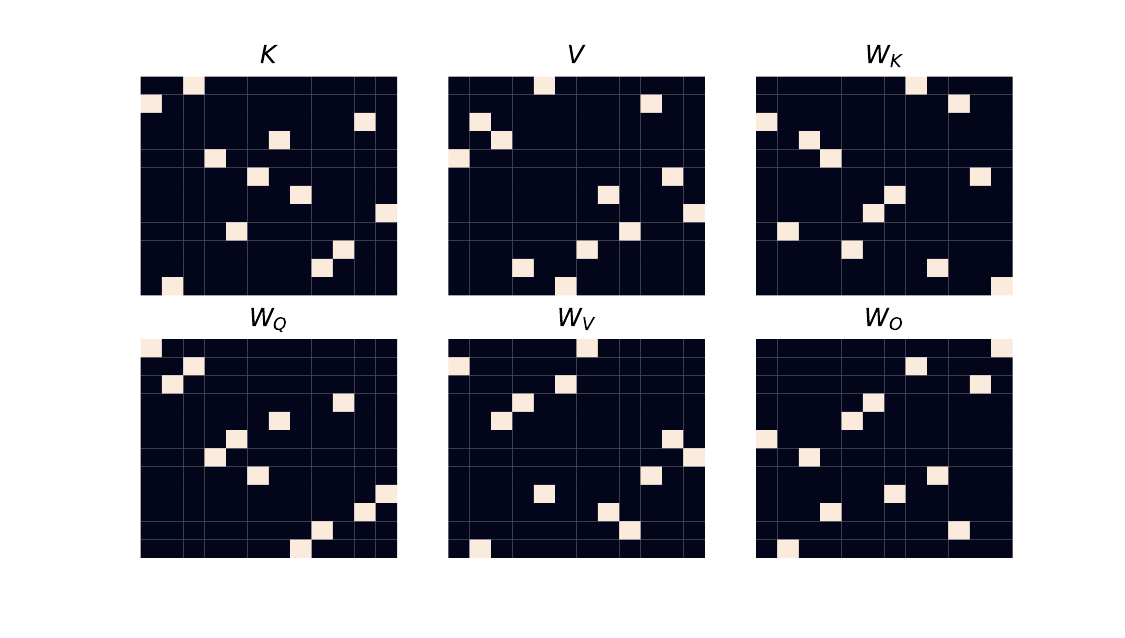}
    
    \subsubsection*{Seed 4 VS Seed 5}
    \vspace*{-0.45cm}
    \includegraphics[width=7.5cm]{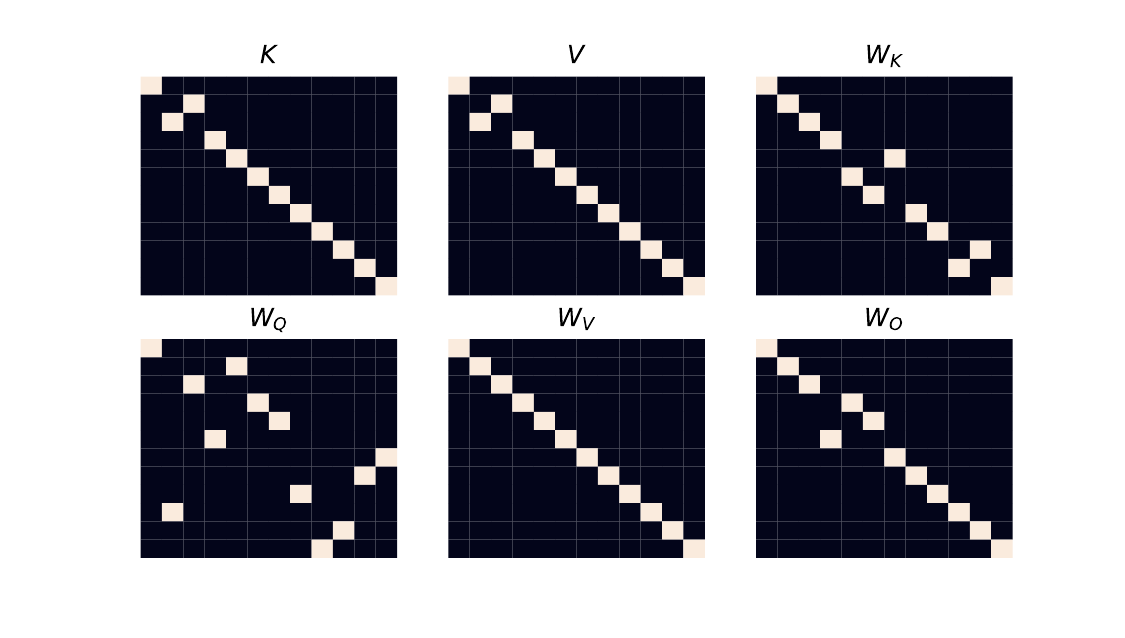} 
    \includegraphics[width=7.5cm]{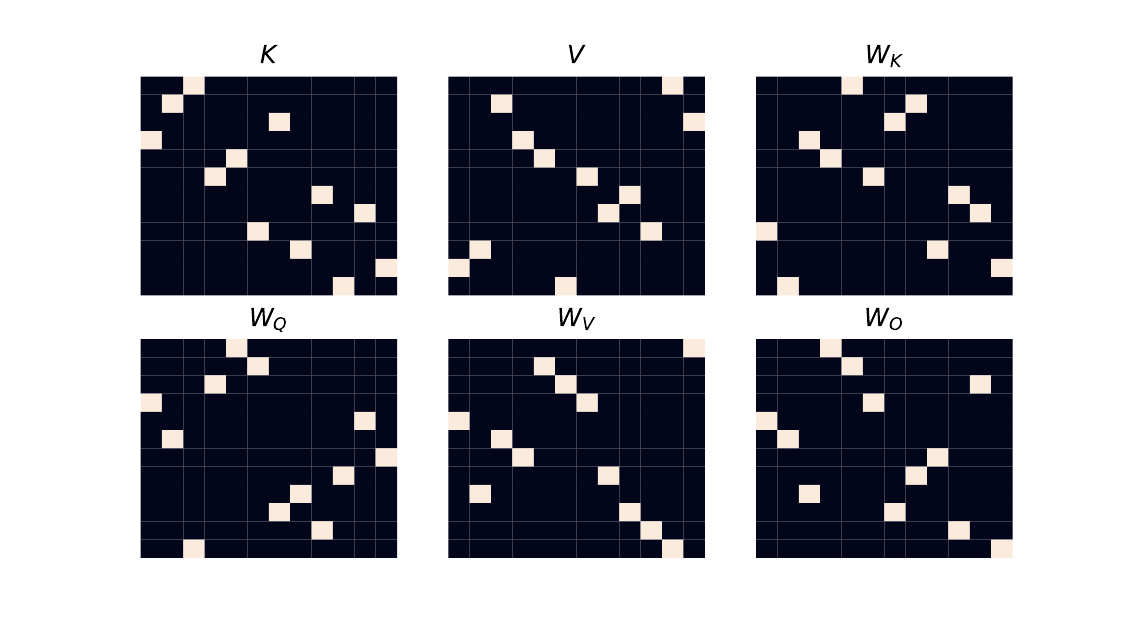}
}
\clearpage
\twocolumn
\section{Example Cases}
\label{appendix:samples}
\subsection{$\WVO$ Matrices}
\label{appendix:samples_wvo}

Below we show output-value pairs from different heads of GPT-2 medium. For each head, we show the 50 pairs with the largest values in the $e \times e$ transition matrix. There are 384 attention heads in GPT-2 medium from which we manually choose a subset. Throughout the section some lists are marked with asterisks indicating the way this particular list was created:
\begin{itemize}
\item[*] -  pairs of the form $(x, x)$ were excluded from the list
\item[**] - pairs where both items are present in the corpus (we use IMDB training set).
\end{itemize}

Along with GPT-2 medium, we also provide a few examples from GPT-2 base and GPT-2 large.

\subsubsection{Low-Level Language Modeling}
\textit{GPT-2 Medium} - Layer 21 Head 7\textsuperscript{*}
\begin{lstlisting}[backgroundcolor=\color{white}]
('NF', 'FN'),
('Ram', ' Ramos'),
('Hug', ' Hughes'),
('gran', 'GR'),
('FN', 'NF'),
('CLA', 'CL'),
('McC', ' McCain'),
('Marsh', ' Marshall'),
(' Hughes', 'Hug'),
('Tan', ' Tanner'),
('nih', 'NH'),
('NRS', 'NR'),
(' Bowman', 'Bow'),
(' Marshall', 'Marsh'),
('Jac', ' Jacobs'),
('Hay', ' Hayes'),
(' Hayes', 'Hay'),
('McC', ' McCorm'),
('NI', 'NR'),
(' sidx', ' Dawson'),
(' Tanner', 'Tan'),
('gra', 'GR'),
('JA', 'jac'),
('zos', 'zo'),
('NI', 'NF'),
('McC', ' McCull'),
(' Jacobs', 'Jac'),
(' Beetle', ' Beet'),
('GF', 'FG'),
('jas', 'ja'),
('Wil', ' Wilkinson'),
(' Ramos', 'Ram'),
('GRE', 'GR'),
(' NF', 'FN'),
(' McCorm', 'McC'),
('Scar', ' Scarborough'),
(' Baal', 'Ba'),
('FP', 'FG'),
('FH', 'FN'),
(' Garfield', 'Gar'),
('jas', 'jac'),
('nuts', 'nut'),
('WI', ' Wis'),
(' Vaughn', ' Vaughan'),
('FP', 'PF'),
('RNA', 'RN'),
(' Jacobs', 'jac'),
('FM', 'FN'),
(' Knox', 'Kn'),
('NI', 'nic')\end{lstlisting}
\textit{GPT-2 Medium} -  Layer 19 Head 13
\footnotesize{(first letter/consonant of the word and last token of the word)}
\begin{lstlisting}[backgroundcolor=\color{white}]
(' R', 'senal'),     #  arsenal
('senal', 'R'),
(' G', 'vernment'),  #  government
(' Madness', ' M'),
(' M', ' Mayhem'),
(' W', 'nesday'),    #  wednesday
('vernment', 'G'),
('M', ' Madness'),
(' N', 'lace'),      #  necklace
('nesday', 'W'),
('Rs', 'senal'),
(' g', 'vernment'),
(' N', 'farious'),   #  nefarious
('eneg', ' C'),
(' r', 'senal'),
(' F', 'ruary'),     #  february
('senal', 'RIC'),
(' R', 'ondo'),
(' N', ' Mandela'),  #  nelson
(' Mayhem', 'M'),
(' RD', 'senal'),
(' C', 'estine'),
('Gs', 'vernment'),
('RF', 'senal'),
(' N', 'esis'),
(' N', 'Reviewed'),
(' C', 'arette'),    #  cigarette
('rome', ' N'),
(' N', 'theless'),   #  nonetheless
('lace', 'N'),
(' H', 'DEN'),
(' V', ' versa'),
(' P', 'bably'),     #  probably
('vernment', 'GF'),
('g', 'vernment'),
('GP', 'vernment'),
(' C', 'ornia'),     #  california
('ilipp', ' F'),
(' N', 'umbered'),
(' C', 'arettes'),
('RS', 'senal'),
(' N', 'onsense'),
('RD', 'senal'),
('RAL', 'senal'),
(' F', 'uci'),
('R', 'ondo'),
(' RI', 'senal'),
(' H', 'iday'),      #  holiday
('senal', ' Rx'),
(' F', 'odor')
 \end{lstlisting}
\textit{GPT-2 Medium} - Layer 20 Head 9
\begin{lstlisting}[backgroundcolor=\color{white}]
('On', ' behalf'),
(' On', ' behalf'),
(' on', ' behalf'),
('during', ' periods'),
('within', ' bounds'),
(' inside', ' envelope'),
('outside', 'door'),
('inside', ' envelope'),
(' Under', ' regime'),
(' during', ' periods'),
(' LIKE', 'lihood'),
(' on', ' occasions'),
('Under', ' regime'),
('inside', 'door'),
('during', 'period'),
('Like', 'lihood'),
(' During', ' periods'),
('Inside', ' envelope'),
('for', ' sake'),
(' inside', ' doors'),
(' under', ' regime'),
(' ON', ' behalf'),
('for', ' purposes'),
('On', ' occasions'),
('inside', ' doors'),
(' on', ' basis'),
(' Under', ' regimes'),
('outside', 'doors'),
('inside', ' Osc'),
('During', ' periods'),
(' inside', 'door'),
(' UNDER', ' regime'),
(' under', ' regimes'),
('Under', ' regimes'),
('inside', 'doors'),
('inside', 'zx'),
('during', ' period'),
('inside', 'ascript'),
('Inside', 'door'),
(' On', ' occasions'),
('BuyableInstoreAndOnline', 'ysc'),
(' Inside', ' envelope'),
('during', ' pauses'),
('under', ' regime'),
(' on', ' occasion'),
('outside', ' doors'),
(' UNDER', ' banner'),
('within', ' envelope'),
(' here', 'abouts'),
('during', ' duration')
\end{lstlisting}
\textit{GPT-2 Base} - Layer 10 Head 11\textsuperscript{**} \begin{lstlisting}[escapechar=\%, backgroundcolor=\color{white}]
(' sources', 'ources')
(' repertoire', ' reperto')
(' tales', ' stories')
(' stories', ' tales')
(' journals', ' magazines')
('stories', ' tales')
(' journal', ' journals')
(' magazines', 'Magazine')
(' magazines', ' newspapers')
(' reperto', ' repertoire')
(' cameras', ' Camer')
(' source', ' sources')
(' newspapers', ' magazines')
(' position', ' positions')
(' tale', ' tales')
(' positions', ' position')
(' obstacles', ' hurdles')
(' chores', ' tasks')
(' journals', ' papers')
(' role', ' roles')
(' hurdles', ' obstacles')
(' journals', ' journal')
(' windows', ' doors')
(' ceiling', ' ceilings')
(' loophole', ' loopholes')
(' Sources', 'ources')
('source', ' sources')
(' documentaries', ' films')
(' microphone', ' microphones')
(' cameras', ' camera')
('Journal', ' journals')
(' restrooms', ' bathrooms')
(' tasks', ' chores')
(' perspectives', ' viewpoints')
(' shelf', ' shelves')
(' rooms', ' bedrooms')
(' hurdle', ' hurdles')
(' barriers', ' fences')
(' magazines', ' journals')
(' journals', 'Magazine')
(' sources', ' source')
(' manuals', ' textbooks')
(' story', ' stories')
(' labs', ' laboratories')
(' tales', ' Stories')
(' chores', ' duties')
(' roles', ' role')
(' ceilings', ' walls')
(' microphones', ' microphone')
(' pathway', ' pathways')
\end{lstlisting}
\textit{GPT-2 Large} - Layer 27 Head 6
\begin{lstlisting}[escapechar=\%, backgroundcolor=\color{white}]
(' where', 'upon'),
('where', 'upon'),
('with', ' regard'),
('with', ' regards'),
(' with', ' regards'),
(' Where', 'upon'),
(' Like', 'lihood'),
('of', ' course'),
(' with', ' regard'),
(' LIKE', 'lihood'),
('Where', 'upon'),
('from', ' afar'),
('with', 'stood'),
(' FROM', ' afar'),
(' like', 'lihood'),
(' WHERE', 'upon'),
('Like', 'lihood'),
(' with', 'stood'),
(' of', ' course'),
('of', 'course'),
('Of', ' course'),
(' from', ' afar'),
(' WITH', ' regard'),
(' where', 'abouts'),
('with', ' impunity'),
(' WITH', ' regards'),
('With', 'stood'),
('for', ' purposes'),
('with', ' respect'),
(' With', 'stood'),
('like', 'lihood'),
(' Of', ' course'),
('With', ' regard'),
(' With', ' regard'),
('where', 'abouts'),
(' WITH', 'stood'),
('With', ' regards'),
(' OF', ' course'),
(' From', ' afar'),
(' with', ' impunity'),
(' With', ' regards'),
(' with', ' respect'),
('From', ' afar'),
('with', 'standing'),
(' on', ' behalf'),
(' by', 'products'),
(' for', ' purposes'),
(' or', 'acle'),
('for', ' sake'),
(' with', 'standing')
  \end{lstlisting}

\subsubsection{Gender}
 \textit{GPT-2 Medium} - Layer 18 Head 1 \begin{lstlisting}[backgroundcolor=\color{white}]
('women', ' Marie'),
(' actresses', ' Marie'),
('women', ' Anne'),
('Women', ' Anne'),
('woman', ' Marie'),
('Women', ' Marie'),
('woman', ' Anne'),
('Woman', ' Marie'),
(' actresses', ' Anne'),
(' heroine', ' Marie'),
('Women', 'Jane'),
(' heroine', ' Anne'),
('women', 'Jane'),
('Women', ' actresses'),
('Woman', ' Anne'),
('Women', ' Esther'),
('women', ' Esther'),
('girls', ' Marie'),
('Mrs', ' Anne'),
(' actress', ' Marie'),
('women', ' actresses'),
('Woman', 'Jane'),
(' girls', ' Marie'),
(' actresses', 'Jane'),
('Woman', 'Anne'),
('Girls', ' Marie'),
('women', 'Anne'),
('Girls', ' Anne'),
('Woman', ' actresses'),
(' Women', ' Marie'),
(' Women', ' Anne'),
(' girls', ' Anne'),
('girl', ' Anne'),
('Women', 'Anne'),
('Woman', 'Women'),
('girls', ' Anne'),
(' actresses', 'Anne'),
('women', ' Michelle'),
(' Actress', ' Marie'),
('girl', ' Marie'),
(' Feminist', ' Anne'),
(' women', ' Marie'),
('Women', ' Devi'),
('Women', ' Elizabeth'),
(' actress', ' Anne'),
('Mrs', 'Anne'),
('answered', 'Answer'),
('woman', 'Anne'),
('Woman', 'maid'),
('women', 'Marie') 
\end{lstlisting}
\textit{GPT-2 Large} -  Layer 27 Head 12
\begin{lstlisting}[escapechar=\%, backgroundcolor=\color{white}]
(' herself', ' Marie'),
(' hers', ' Marie'),
('she', ' Marie'),
(' she', ' Marie'),
(' her', ' Marie'),
('She', ' Marie'),
(' hers', 'Maria'),
(' actresses', ' actresses'),
(' herself', 'Maria'),
(' her', 'Maria'),
(' herself', ' Anne'),
('She', 'Maria'),
(' hers', ' Louise'),
(' herself', ' Louise'),
(' hers', ' Anne'),
(' hers', 'pher'),
('she', 'Maria'),
(' actress', ' actresses'),
(' herself', ' Isabel'),
(' herself', 'pher'),
(' she', 'Maria'),
(' SHE', ' Marie'),
(' herself', ' Gloria'),
(' herself', ' Amanda'),
(' Ivanka', ' Ivanka'),
(' her', ' Louise'),
(' herself', ' Kate'),
(' her', 'pher'),
(' her', ' Anne'),
(' she', 'pher'),
('she', ' Louise'),
(' herself', 'Kate'),
(' she', ' Louise'),
(' she', ' Anne'),
(' She', ' Marie'),
('she', ' Gloria'),
('She', ' Louise'),
(' hers', ' Gloria'),
(' herself', ' Diana'),
('She', ' Gloria'),
('she', ' Anne'),
('she', 'pher'),
('Her', ' Marie'),
(' she', ' Gloria'),
(' Paleo', ' Paleo'),
(' hers', ' Diana')
\end{lstlisting}
\textit{GPT-2 Base} - Layer 9 Head 7\textsuperscript{**} \begin{lstlisting}[escapechar=\%, backgroundcolor=\color{white}]
(' her', ' herself')
('She', ' herself')
(' she', ' herself')
('she', ' herself')
('Her', ' herself')
(' She', ' herself')
(' SHE', ' herself')
('their', ' themselves')
(' hers', ' herself')
('Their', ' themselves')
(' Her', ' herself')
(' Their', ' themselves')
(' THEIR', ' themselves')
(' HER', ' herself')
(' their', ' themselves')
('They', ' themselves')
('His', ' himself')
(' herself', 'erest')
('they', ' themselves')
('his', ' himself')
('Their', 'selves')
(' They', ' themselves')
(' herself', ' Louise')
('their', 'selves')
('her', ' herself')
(' his', ' himself')
(' herself', ' Marie')
('He', ' himself')
('She', ' Louise')
(' they', ' themselves')
('their', 'chairs')
(' herself', ' dow')
(' herself', 'eva')
(' THEY', ' themselves')
(' herself', ' Mae')
(' His', ' himself')
('clinton', 'enegger')
('She', 'erest')
(' her', ' Louise')
(' herself', ' Devi')
(' Their', 'selves')
('Their', 'chairs')
(' Himself', 'enegger')
(' she', ' Louise')
(' herself', ' Anne')
('Its', ' itself')
(' her', 'erest')
(' herself', ' Christina')
('she', 'erest')
('their', ' selves')
\end{lstlisting} 
 \subsubsection{Geography}
\textit{GPT-2 Base} - Layer 11 Head 2\textsuperscript{**} \begin{lstlisting}[escapechar=\%, backgroundcolor=\color{white}]
(' Halifax', ' Scotia')
('Saudi', ' Arabia')
(' Nova', ' Scotia')
(' Tamil', ' Nadu')
(' Finnish', 'onen')
(' Saudi', ' Arabia')
('Pitt', 'sburgh')
('Dutch', 'ijk')
(' Schwartz', 'enegger')
(' Afghans', ' Kabul')
(' Icelandic', 'sson')
(' Finland', 'onen')
('Pitt', 'enegger')
(' Czech', 'oslov')
(' Manitoba', ' Winnipeg')
(' Malaysian', ' Lumpur')
(' Swedish', 'borg')
(' Saskatchewan', ' Sask')
(' Chennai', ' Nadu')
(' Argentine', ' Aires')
(' Iceland', ' Icelandic')
(' Swedish', 'sson')
(' Tasman', ' Nadu')
('Houston', ' Astros')
('Colorado', ' Springs')
(' Kuala', ' Lumpur')
('Tai', 'pport')
('Houston', ' Dynamo')
(' Manitoba', 'Marginal')
(' Afghan', ' Kabul')
(' Buenos', ' Aires')
(' Alberta', ' Calgary')
(' Stockholm', 'sson')
(' Sweden', 'borg')
('Brazil', ' Paulo')
(' Iceland', 'sson')
(' Winnipeg', ' Manitoba')
(' Sweden', 'sson')
(' Carolina', ' Hurricanes')
(' Dutch', 'ijk')
(' Swed', 'borg')
(' Aki', 'pport')
(' Winnipeg', 'Marginal')
(' Argentine', ' pes')
(' Halifax', 'imore')
(' Brisbane', 'enegger')
(' Melbourne', ' Nadu')
(' Adelaide', ' Nadu')
(' Cambod', ' Nguyen')
(' Vietnamese', ' Nguyen') 
 \end{lstlisting}
\textit{GPT-2 Medium} - Layer 16 Head 6\textsuperscript{*} \begin{lstlisting}[escapechar=\%, backgroundcolor=\color{white}]
(' Chennai', ' Mumbai'),
('India', ' Mumbai'),
(' Mumbai', ' Chennai'),
(' Queensland', ' Tasmania'),
('India', ' Rahul'),
('India', ' Gujar'),
(' Chennai', ' Bangalore'),
('England', 'Scotland'),
(' Chennai', ' Kerala'),
(' Delhi', ' Mumbai'),
('Britain', 'Scotland'),
(' Bangalore', ' Mumbai'),
('Pakistan', 'India'),
('Scotland', 'Ireland'),
(' Mumbai', ' Bangalore'),
(' Bangalore', ' Chennai'),
(' Aadhaar', ' Gujar'),
(' Mumbai', ' Maharashtra'),
(' Maharashtra', ' Gujarat'),
(' Gujarat', ' Gujar'),
('Australian', 'Australia'),
('India', ' Gujarat'),
(' Rahul', ' Gujar'),
(' Maharashtra', ' Mumbai'),
('Britain', 'England'),
('India', ' Chennai'),
(' Mumbai', ' Bombay'),
(' Tamil', ' Kerala'),
(' Hindi', ' Mumbai'),
(' Tasmania', ' Tasman'),
(' Mumbai', 'India'),
(' Hindi', ' Gujar'),
(' Maharashtra', ' Gujar'),
(' Australians', 'Austral'),
(' Maharashtra', ' Kerala'),
('India', ' Bangalore'),
('India', ' Kerala'),
('India', ' Bombay'),
('Australia', 'Austral'),
(' Aadhaar', 'India'),
(' Sharma', ' Mumbai'),
('Australian', 'Austral'),
(' Mumbai', ' Kerala'),
('Scotland', 'England'),
(' Mumbai', ' Gujar'),
(' Rahul', ' Mumbai'),
(' Queensland', ' Tasman'),
(' Tamil', ' Chennai'),
(' Gujarat', ' Maharashtra'),
('India', ' Modi')
\end{lstlisting}
\textit{GPT-2 Medium} - Layer 16 Head 2\textsuperscript{*}
\begin{lstlisting}[escapechar=\%, backgroundcolor=\color{white}]
('Austral', ' Australians'),
('Australia', 'Austral'),
(' Canberra', 'Austral'),
('Austral', ' Canberra'),
(' Winnipeg', ' Edmonton'),
('Australian', 'Austral'),
(' Alberta', ' Edmonton'),
('Australia', ' Australians'),
(' Australians', 'Austral'),
('Ukraine', 'ovych'),
(' Quebec', ' Canad'),
('Australian', ' Australians'),
(' Winnipeg', ' Manitoba'),
(' Manitoba', ' Winnipeg'),
('Canadian', 'Canada'),
('Moscow', ' Bulgar'),
(' Manitoba', ' Edmonton'),
('berra', 'Austral'),
('Austral', 'Australian'),
(' Ukrainians', 'ovych'),
('Canada', ' Canadians'),
(' Canberra', ' Australians'),
('Canada', 'Canadian'),
(' Yanukovych', 'ovych'),
('Canada', ' Trudeau'),
(' Dmitry', ' Bulgar'),
(' Australia', 'Austral'),
(' Mulcair', ' Canad'),
('berra', ' Canberra'),
('Turkish', 'oglu'),
('udeau', 'Canada'),
(' Edmonton', ' Oilers'),
('Australia', ' Canberra'),
('Canada', ' Edmonton'),
(' Edmonton', ' Calgary'),
(' Alberta', ' Calgary'),
('udeau', ' Trudeau'),
(' Calgary', ' Edmonton'),
('Canadian', ' Trudeau'),
('Australian', ' Canberra'),
(' Vancouver', ' Canucks'),
('Australia', 'Australian'),
(' Vancouver', ' Fraser'),
('Canadian', ' Edmonton'),
('Austral', 'elaide'),
('Tex', ' Braz'),
('Canada', ' RCMP'),
('Moscow', 'sov'),
('Russia', ' Bulgar'),
(' Canadians', 'Canada')
\end{lstlisting}
\textit{GPT-2 Medium} -  Layer 21 Head 12\textsuperscript{*}
\begin{lstlisting}[escapechar=\%, backgroundcolor=\color{white}]
(' Indonesian', ' Indones'),
(' Vietnamese', ' Nguyen'),
(' Indonesian', ' Jakarta'),
(' Indonesian', ' Indonesia'),
('Turkish', 'oglu'),
(' Indonesia', ' Indones'),
(' Jakarta', ' Indones'),
(' Korean', ' Koreans'),
(' Turkish', 'oglu'),
(' Taiwan', ' Taiwanese'),
(' Thai', ' Nguyen'),
(' Brazilian', 'Brazil'),
(' Indones', ' Indonesia'),
('Tai', ' Taiwanese'),
(' Istanbul', 'oglu'),
(' Indones', ' Indonesian'),
(' Indones', ' Jakarta'),
(' Laos', ' Nguyen'),
(' Slovenia', ' Sloven'),
(' Koreans', ' Korean'),
(' Cambod', ' Nguyen'),
('Italy', 'zzi'),
(' Taiwanese', 'Tai'),
(' Indonesia', ' Jakarta'),
(' Indonesia', ' Indonesian'),
(' Bulgarian', ' Bulgaria'),
(' Iceland', ' Icelandic'),
(' Korea', ' Koreans'),
('Brazil', ' Brazilian'),
(' Bulgarian', ' Bulgar'),
(' Malaysian', ' Malays'),
(' Ankara', 'oglu'),
(' Bulgaria', ' Bulgarian'),
(' Malays', ' Indones'),
(' Taiwanese', ' Tai'),
('Turkey', 'oglu'),
('Brazil', ' Janeiro'),
('Italian', 'zzi'),
(' Kuala', ' Malays'),
('Japanese', ' Fuk'),
(' Jakarta', ' Indonesian'),
(' Taiwanese', ' Taiwan'),
(' Erdogan', 'oglu'),
(' Viet', ' Nguyen'),
(' Philippine', ' Filipino'),
(' Jakarta', ' Indonesia'),
(' Koreans', ' Jong'),
(' Filipino', ' Duterte'),
(' Azerbaijan', ' Azerbai'),
(' Bulgar', ' Bulgarian')
\end{lstlisting}
\textit{GPT-2 Large} -  Layer 23 Head 5
\begin{lstlisting}[escapechar=\%, backgroundcolor=\color{white}]
('Canada', ' Trudeau'),
(' Canadians', ' Trudeau'),
('Canadian', ' Trudeau'),
(' Queensland', ' Tasman'),
(' Tasman', ' Tasman'),
(' Canada', ' Trudeau'),
(' Canberra', ' Canberra'),
(' Winnipeg', ' Winnipeg'),
(' Canberra', ' Tasman'),
('Canadian', 'Canada'),
(' Canadian', ' Trudeau'),
(' Brisbane', ' Brisbane'),
(' Quebec', ' Trudeau'),
('Canadian', ' Canadian'),
(' Brisbane', ' Tasman'),
(' Tasmania', ' Tasman'),
('Canadian', ' Canadians'),
(' RCMP', ' Trudeau'),
(' Manitoba', ' Trudeau'),
(' Queensland', ' Brisbane'),
(' Queensland', ' Canberra'),
('Canada', ' Saskatchewan'),
('Canadian', ' Saskatchewan'),
('Canada', ' Canadian'),
(' RCMP', ' Saskatchewan'),
(' Canberra', ' Brisbane'),
(' Canadians', 'Canada'),
(' Winnipeg', ' Trudeau'),
('Canadian', ' Canada'),
('Canada', ' Canadians'),
('Australian', ' Canberra'),
(' Melbourne', ' Canberra'),
(' RCMP', ' Canad'),
(' Canadians', ' Canadians'),
('CBC', ' Trudeau'),
(' Canadian', ' Canadian'),
('Canadian', ' Winnipeg'),
(' Australians', ' Canberra'),
(' Quebec', 'Canada'),
(' Canadian', 'Canada'),
(' NSW', ' Canberra'),
('Toronto', ' Canad'),
('Canada', 'Canada'),
(' NSW', ' Tasman'),
(' RCMP', ' RCMP'),
(' Canadian', ' Canadians'),
(' Saskatchewan', ' Saskatchewan'),
(' Canadians', ' Saskatchewan'),
('Canadian', ' Canad'),
(' Ottawa', ' Winnipeg')
\end{lstlisting}

\subsubsection{British Spelling}
\textit{GPT-2 Medium} - Layer 19 Head 4 \begin{lstlisting}[escapechar=\%, backgroundcolor=\color{white}]
(' realise', ' Whilst'),
(' Whilst', ' Whilst'),
(' realised', ' Whilst'),
(' organise', ' Whilst'),
(' recognise', ' Whilst'),
(' civilisation', ' Whilst'),
(' organisation', ' Whilst'),
(' whilst', ' Whilst'),
(' organising', ' Whilst'),
(' organised', ' Whilst'),
(' organis', ' Whilst'),
(' util', ' Whilst'),
(' apologise', ' Whilst'),
(' emphas', ' Whilst'),
(' analyse', ' Whilst'),
(' organisations', ' Whilst'),
(' recognised', ' Whilst'),
(' flavours', ' Whilst'),
(' colour', ' Whilst'),
('colour', ' Whilst'),
(' Nasa', ' Whilst'),
(' Nato', ' Whilst'),
(' analys', ' Whilst'),
(' flavour', ' Whilst'),
(' colourful', ' Whilst'),
(' colours', ' Whilst'),
(' realise', ' organising'),
(' behavioural', ' Whilst'),
(' coloured', ' Whilst'),
(' learnt', ' Whilst'),
(' favourable', ' Whilst'),
('isation', ' Whilst'),
(' programmes', ' Whilst'),
(' realise', ' organis'),
(' authorised', ' Whilst'),
(' practise', ' Whilst'),
(' criticised', ' Whilst'),
(' organisers', ' Whilst'),
(' organise', ' organising'),
(' analysed', ' Whilst'),
(' programme', ' Whilst'),
(' behaviours', ' Whilst'),
(' humour', ' Whilst'),
('isations', ' Whilst'),
(' tyres', ' Whilst'),
(' aluminium', ' Whilst'),
(' realise', ' organised'),
(' favour', ' Whilst'),
(' ageing', ' Whilst'),
(' organise', ' organis')
\end{lstlisting}

\subsubsection{Related Words}
\textit{GPT-2 Medium} - Layer 13 Head 8\textsuperscript{*}
\begin{lstlisting}[backgroundcolor=\color{white}]
(' miraculous', ' mirac'),
(' miracle', ' mirac'),
(' nuance', ' nuanced'),
(' smarter', 'Better'),
(' healthier', ' equitable'),
(' liberated', ' liberating'),
(' untouched', ' unaffected'),
(' unbiased', ' equitable'),
('failed', ' inconsistent'),
(' liberated', ' emanc'),
(' humane', ' equitable'),
(' liberating', ' liberated'),
('failed', ' incompatible'),
(' miracles', ' mirac'),
(' peacefully', ' consensual'),
(' unconditional', ' uncond'),
(' unexpectedly', ' unexpected'),
(' untouched', ' unconditional'),
(' healthier', 'Better'),
(' unexpected', ' unexpectedly'),
(' peacefully', ' graceful'),
(' emancipation', ' emanc'),
(' seamlessly', ' effortlessly'),
(' peacefully', ' honorable'),
(' uncond', ' unconditional'),
(' excuses', ' rubbish'),
(' liberating', ' emanc'),
(' peacefully', ' equitable'),
(' gracious', ' Feather'),
(' liberated', ' emancipation'),
(' nuances', ' nuanced'),
(' avoids', 'icable'),
(' freeing', ' liberated'),
(' freeing', ' liberating'),
(' lousy', ' inconsistent'),
('failed', ' lousy'),
(' unaffected', ' unconditional'),
('ivable', ' equitable'),
('Honest', ' equitable'),
(' principled', 'erning'),
('surv', ' survival'),
(' lackluster', 'ocre'),
(' liberating', ' equitable'),
('Instead', 'Bah'),
(' inappropriate', ' incompatible'),
(' emanc', ' emancipation'),
(' unaffected', ' unchanged'),
(' peaceful', ' peacefully'),
(' safer', ' equitable'),
(' uninterrupted', ' unconditional')
\end{lstlisting}
\textit{GPT-2 Medium} - Layer 12 Head 14\textsuperscript{*}
\begin{lstlisting}[backgroundcolor=\color{white}]
(' died', ' perished'),
(' dies', ' perished'),
(' testifying', ' testify'),
(' interven', ' intervened'),
(' advising', ' advises'),
(' disband', ' disbanded'),
(' perished', 'lost'),
(' perished', ' died'),
(' applaud', ' applauded'),
(' dictate', ' dictates'),
(' prevailed', ' prev'),
(' advising', ' advise'),
('thood', 'shed'),
('orsi', 'Reviewed'),
(' perished', ' dies'),
(' publishes', 'published'),
(' prevail', ' prevailed'),
(' dies', ' died'),
(' testifying', ' testified'),
(' testify', ' testifying'),
(' governs', ' dictates'),
(' complicity', ' complicit'),
(' dictate', ' dictated'),
('CHO', 'enough'),
('independence', ' skelet'),
(' prescribe', ' Recomm'),
(' perished', 'essential'),
('CHO', 'noticed'),
(' approving', 'avorable'),
(' perished', ' perish'),
(' oversee', ' overseeing'),
('shed', ' skelet'),
('chart', 'EY'),
(' overseeing', ' presiding'),
('pees', ' fundament'),
('appro', ' sanction'),
(' prevailed', ' prevail'),
(' regulates', ' governs'),
('shed', 'tails'),
('chart', ' Period'),
('hower', 'lihood'),
(' prevail', ' prev'),
('helps', ' aids'),
(' dict', ' dictated'),
(' dictates', ' dictated'),
('itta', ' Dise'),
('CHO', 'REC'),
('ORTS', 'exclusive'),
('helps', ' Helpful'),
('ciples', 'bart')
\end{lstlisting}
\textit{GPT-2 Medium} - Layer 14 Head 1\textsuperscript{*}
\begin{lstlisting}[backgroundcolor=\color{white}]
(' incorrectly', ' misunderstand'),
(' properly', ' Proper'),
(' incorrectly', ' inaccur'),
(' wrongly', ' misunderstand'),
(' incorrectly', ' misinterpret'),
(' incorrectly', ' incorrect'),
(' incorrectly', ' mistakes'),
(' incorrectly', ' misunderstanding'),
(' properly', ' proper'),
(' incorrectly', 'fail'),
(' incorrectly', ' faulty'),
(' incorrectly', ' misrepresent'),
(' fails', ' failing'),
(' incorrectly', ' inaccurate'),
(' incorrectly', ' errors'),
(' Worse', ' harmful'),
(' wrong', ' misunderstand'),
(' improperly', ' misunderstand'),
(' incorrectly', 'wrong'),
(' incorrectly', ' harmful'),
(' incorrectly', ' mistake'),
(' incorrectly', ' mis'),
(' fails', 'fail'),
(' Worse', ' detrimental'),
(' properly', ' rightful'),
(' inappropriately', ' misunderstand'),
(' unnecessarily', ' harmful'),
(' unnecessarily', ' neglect'),
(' properly', ' correctly'),
(' Worse', ' Worst'),
(' fails', ' failure'),
(' adequately', ' satisfactory'),
(' incorrectly', ' defective'),
(' mistakenly', ' misunderstand'),
(' Worse', ' harming'),
(' incorrectly', ' mishand'),
(' adequately', 'adequ'),
(' incorrectly', ' misuse'),
(' fails', 'Failure'),
(' Worse', ' hurts'),
('wrong', ' misunderstand'),
(' incorrectly', ' mistakenly'),
(' fails', ' failures'),
(' adequately', ' adequate'),
(' correctly', ' properly'),
(' Worse', ' hurting'),
(' correctly', ' Proper'),
(' fails', ' fail'),
(' incorrectly', ' mistaken'),
(' adversely', ' harming')
\end{lstlisting}
\textit{GPT-2 Large} -  Layer 24 Head 9
 \begin{lstlisting}[escapechar=\%, backgroundcolor=\color{white}]
(' interviewer', ' interviewer'),
(' lectures', ' lectures'),
(' lecture', ' lecture'),
(' interview', 'Interview'),
(' interview', ' interview'),
(' interview', ' interviewer'),
(' interviewing', ' interviewing'),
(' magazine', ' magazine'),
(' Reviews', ' Reviews'),
(' reviewer', ' reviewer'),
(' reviewers', ' reviewers'),
(' lectures', ' lecture'),
(' testers', ' testers'),
(' editors', ' editors'),
(' interviewer', ' interview'),
(' Interview', 'Interview'),
(' interviewer', 'Interview'),
('Interview', 'Interview'),
(' lecture', ' lectures'),
(' interviewing', ' interviewer'),
(' journal', ' journal'),
(' interviewer', ' interviewing'),
(' blogs', ' blogs'),
(' editorial', ' editorial'),
(' tests', ' tests'),
(' presentations', ' presentations'),
(' Editorial', ' Editorial'),
(' interview', ' Interview'),
(' reviewer', ' reviewers'),
(' interviews', 'Interview'),
(' interview', ' interviewing'),
(' interviewer', ' Interview'),
(' interviews', ' interview'),
(' Interview', ' Interview'),
(' interviewing', 'Interview'),
('Interview', ' interviewer'),
(' testifying', ' testifying'),
(' reviewers', ' reviewer'),
(' blogging', ' blogging'),
(' broadcast', ' broadcast'),
(' Interview', ' interviewer'),
(' magazine', ' magazines'),
(' editorial', ' Editorial'),
(' interview', ' interviews'),
(' interviewing', ' interview'),
(' Interview', ' interview'),
(' interviews', ' interviews'),
(' tests', 'tests'),
(' interviews', ' interviewing'),
('Interview', ' interview')\end{lstlisting}
\textit{GPT-2 Medium} - Layer 14 Head 13\textsuperscript{*} \begin{lstlisting}[backgroundcolor=\color{white}]
(' editorial', ' editors'),
(' broadcasting', ' broadcasters'),
(' broadcasts', ' broadcasting'),
(' broadcasts', ' broadcast'),
(' broadcasters', ' Broadcasting'),
(' Editorial', ' editors'),
(' broadcast', ' broadcasters'),
(' broadcast', ' Broadcasting'),
(' lecture', ' lectures'),
(' broadcasting', ' Broadcast'),
(' broadcaster', ' broadcasters'),
(' broadcasts', ' broadcasters'),
(' publishing', ' Publishers'),
(' broadcast', ' broadcasting'),
(' Broadcasting', ' broadcasters'),
(' Publishing', ' Publishers'),
(' lectures', ' lecture'),
(' editorial', ' Editors'),
(' broadcasting', ' broadcast'),
(' broadcasts', ' Broadcasting'),
(' broadcasters', ' broadcasting'),
(' journalistic', ' journalism'),
('Journal', 'reports'),
(' Broadcasting', ' Broadcast'),
('Publisher', ' Publishers'),
(' Broadcasting', 'azeera'),
('Journal', 'Reporting'),
(' journalism', ' journalistic'),
(' broadcaster', ' Broadcasting'),
(' broadcaster', ' broadcasting'),
(' broadcasting', ' broadcaster'),
(' publication', ' editors'),
('journal', ' journalism'),
('Journal', ' Journalists'),
(' documentaries', ' documentary'),
(' filmed', ' filming'),
(' publishing', ' publishers'),
('Journal', ' journalism'),
(' broadcasts', ' Broadcast'),
(' broadcasters', ' broadcast'),
('Journal', ' articles'),
('reports', ' reporting'),
(' manuscript', ' manuscripts'),
(' publishing', ' publish'),
(' broadcasters', 'azeera'),
(' publication', ' Publishers'),
(' publications', ' Publishers'),
(' Newsp', ' newspapers'),
(' broadcasters', ' Broadcast'),
('Journal', ' Readers')
\end{lstlisting}
 

\subsection{Query-Key Matrices} 
\label{appendix:samples_wqk}
\textit{GPT-2 Large} - Layer 19 Head 7\textsuperscript{**}
 \begin{lstlisting}[escapechar=\%, backgroundcolor=\color{white}]
(' tonight', 'Friday'),
(' Copyright', 'Returns'),
('TM', 'review'),
(' Weekend', 'Preview'),
(' tonight', 'Thursday'),
(' recently', 'Closure'),
(' Copyright', 'Contents'),
(' Copyright', 'Wisconsin'),
(' Copyright', 'Methods'),
(' tonight', 'Sunday'),
(' tomorrow', ' postpone'),
(' tomorrow', ' tonight'),
(' recently', 'acerb'),
(' Copyright', 'Rated'),
(' myself', ' my'),
(' Copyright', 'Cop'),
(' Wednesday', 'Closure'),
(' Billion', ' 1935'),
(' tonight', 'Saturday'),
(' tonight', ' celebr'),
(' tomorrow', ' postponed'),
(' Copyright', 'Show'),
(' Wednesday', 'Friday'),
(' Copyright', 'Earn'),
(' Billion', ' 1934'),
(' Eric', 'Larry'),
(' 2015', 'Released'),
(' Copyright', 'Rat'),
(' tomorrow', ' postp'),
(' 2017', 'Latest'),
(' previous', 'obin'),
(' controversial', 'Priv'),
(' recently', ' nightly'),
('Base', ' LV'),
(' recently', 'Project'),
(' historically', ' globalization'),
(' recently', ' vulner'),
(' tonight', 'Wednesday'),
(' Copyright', 'Abstract'),
(' Tuesday', 'Friday'),
(' Anthony', 'Born'),
(' Budget', 'Premium'),
(' tonight', 'Welcome'),
('yle', 'lite'),
(' Wednesday', 'Latest'),
(' Latest', 'show'),
(' B', ' pione'),
(' Copyright', 'cop'),
(' Pablo', ' Dia'),
(' recent', 'Latest')
  \end{lstlisting}
\textit{GPT-2 Medium} - Layer 22 Head 1
\begin{lstlisting}[backgroundcolor=\color{white}]
(' usual', ' usual'),
(' occasional', ' occasional'),
(' aforementioned', ' aforementioned'),
(' general', ' usual'),
(' usual', ' slightest'),
('agn', 'ealous'),
(' traditional', ' usual'),
(' free', 'amina'),
(' major', ' major'),
(' frequent', ' occasional'),
(' generous', ' generous'),
(' free', 'lam'),
(' regular', ' usual'),
(' standard', ' usual'),
(' main', ' usual'),
(' complete', ' Finished'),
(' main', 'liest'),
(' traditional', ' traditional'),
(' latest', ' aforementioned'),
(' current', ' aforementioned'),
(' normal', ' usual'),
(' dominant', ' dominant'),
(' free', 'ministic'),
(' brief', ' brief'),
(' biggest', 'liest'),
('usual', ' usual'),
(' rash', ' rash'),
(' regular', ' occasional'),
(' specialized', ' specialized'),
(' free', 'iosis'),
(' free', 'hero'),
(' specialty', ' specialty'),
(' general', 'iosis'),
(' nearby', ' nearby'),
(' best', 'liest'),
(' officially', ' formal'),
(' immediate', 'mediate'),
(' special', ' ultimate'),
(' free', 'otropic'),
(' rigorous', ' comparative'),
(' actual', ' slightest'),
(' complete', ' comparative'),
(' typical', ' usual'),
(' modern', ' modern'),
(' best', ' smartest'),
(' free', ' free'),
(' highest', ' widest'),
(' specialist', ' specialist'),
(' appropriate', ' slightest'),
(' usual', 'liest')
\end{lstlisting}
\textit{GPT-2 Large} -  Layer 20 Head 13 \textsuperscript{**}
 \begin{lstlisting}[escapechar=\%, backgroundcolor=\color{white}]
(' outdoors', ' outdoors'),
(' outdoor', ' outdoors'),
(' Gre', 'burg'),
(' healing', ' healing'),
(' indoor', ' outdoors'),
(' Hemp', 'burg'),
(' Ticket', ' Ticket'),
(' accommodations', ' accommodations'),
('eco', 'aco'),
('prem', 'otti'),
(' Candy', 'cott'),
(' decorative', ' ornament'),
('yan', 'ava'),
(' deadlines', ' schedule'),
(' Lor', 'ian'),
(' architectural', ' ornament'),
(' Ratings', ' Ratings'),
(' Bod', 'za'),
(' exotic', ' exotic'),
(' food', ' baths'),
(' Marketplace', ' Marketplace'),
(' heal', ' healing'),
(' Ex', 'ilus'),
(' indoors', ' outdoors'),
(' therm', ' therm'),
(' bleach', ' coated'),
(' Sod', 'opol'),
(' District', ' Metropolitan'),
(' Anonymous', ' Rebell'),
(' Corn', 'burg'),
(' indoor', ' indoors'),
(' R', 'vale'),
('rom', 'otti'),
(' ratings', ' Ratings'),
(' attendance', ' attendance'),
(' destinations', ' destinations'),
(' VIDEOS', ' VIDEOS'),
('yan', 'opol'),
(' Suffolk', 'ville'),
(' retali', ' against'),
('mos', 'oli'),
(' pacing', ' pacing'),
(' Spectrum', ' QC'),
(' Il', 'ian'),
(' archived', ' archived'),
(' Pledge', ' Pledge'),
('alg', 'otti'),
(' Freedom', 'USA'),
('anto', 'ero'),
(' decorative', ' decoration')
 \end{lstlisting}
\textit{GPT-2 Medium} - Layer 0 Head 9
\begin{lstlisting}[backgroundcolor=\color{white}]
('59', '27'),
('212', '39'),
('212', '38'),
('217', '39'),
('37', '27'),
('59', '26'),
('54', '88'),
('156', '39'),
('212', '79'),
('59', '28'),
('57', '27'),
('212', '57'),
('156', '29'),
('36', '27'),
('217', '79'),
('59', '38'),
('63', '27'),
('72', '39'),
('57', '26'),
('57', '34'),
('59', '34'),
('156', '27'),
('91', '27'),
('156', '38'),
('63', '26'),
('59', '25'),
('138', '27'),
('217', '38'),
('72', '27'),
('54', '27'),
('36', '29'),
('72', '26'),
('307', '39'),
('37', '26'),
('217', '57'),
('37', '29'),
('54', '38'),
('59', '29'),
('37', '28'),
('307', '38'),
('57', '29'),
('63', '29'),
('71', '27'),
('138', '78'),
('59', '88'),
('89', '27'),
('561', '79'),
('212', '29'),
('183', '27'),
('54', '29')
\end{lstlisting}
\textit{GPT-2 Medium} - Layer 17 Head 6\textsuperscript{*}
\begin{lstlisting}[backgroundcolor=\color{white}]
(' legally', ' legal'),
(' legal', ' sentencing'),
(' legal', ' arbitration'),
(' boycot', ' boycott'),
(' legal', ' criminal'),
(' legal', ' Judicial'),
(' legal', ' rulings'),
(' judicial', ' sentencing'),
(' marketing', ' advertising'),
(' legal', ' confidential'),
(' protesting', ' protest'),
(' recruited', ' recruit'),
(' recruited', ' recruits'),
(' judicial', ' criminal'),
(' legal', ' exemptions'),
(' demographics', ' demographic'),
(' boycott', ' boycot'),
(' sentencing', ' criminal'),
(' recruitment', ' recruits'),
(' recruitment', ' recruit'),
(' Constitutional', ' sentencing'),
(' Legal', ' sentencing'),
(' constitutional', ' sentencing'),
(' legal', ' subpoena'),
(' injury', ' injuries'),
(' FOIA', ' confidential'),
(' legal', ' licenses'),
(' donation', ' donations'),
(' disclosure', ' confidential'),
(' negotiation', ' negotiating'),
(' Judicial', ' legal'),
(' legally', ' criminal'),
(' legally', ' confidential'),
(' legal', ' jur'),
(' legal', ' enforcement'),
(' legal', ' lawyers'),
(' legally', ' enforcement'),
(' recruitment', ' recruiting'),
(' recruiting', ' recruit'),
(' criminal', ' sentencing'),
(' legal', ' attorneys'),
(' negotiations', ' negotiating'),
(' legally', ' arbitration'),
(' recruited', ' recruiting'),
(' legally', ' exemptions'),
(' legal', ' judicial'),
(' voting', ' Vote'),
(' negotiated', ' negotiating'),
(' legislative', ' veto'),
(' funding', ' funded')
\end{lstlisting}
\textit{GPT-2 Medium} - Layer 17 Head 7
\begin{lstlisting}[backgroundcolor=\color{white}]
('tar', 'idia'),
(' [...]', '..."'),
(' lecture', ' lectures'),
(' Congress', ' senate'),
(' staff', ' staffers'),
(' Scholarship', ' collegiate'),
(' executive', ' overseeing'),
(' Scholarship', ' academic'),
(' academ', ' academic'),
('."', '..."'),
(' [', '..."'),
('";', '..."'),
(' Memorial', 'priv'),
(' festival', 'conference'),
('crew', ' supervisors'),
(' certification', ' grading'),
(' scholarship', ' academic'),
(' rumored', ' Academic'),
(' Congress', ' delegated'),
(' staff', ' technicians'),
('Plex', ' CONS'),
(' congress', ' senate'),
(' university', ' tenure'),
(' Congress', ' appointed'),
(' Congress', ' duly'),
(' investigative', ' investig'),
(' legislative', ' senate'),
('ademic', ' academic'),
('bench', ' academic'),
(' scholarship', ' tenure'),
(' campus', ' campuses'),
(' staff', ' Facilities'),
(' Editorial', 'mn'),
(' clinic', ' laboratory'),
(' crew', ' crews'),
(' Scholarship', ' academ'),
(' staff', ' staffer'),
('icken', 'oles'),
('?"', '..."'),
(' Executive', ' overseeing'),
(' academic', ' academ'),
(' Congress', 'atra'),
('aroo', 'anny'),
(' academic', ' academia'),
(' Congress', ' Amendments'),
(' academic', ' academics'),
('student', ' academic'),
(' committee', ' convened'),
('",', '..."'),
('ove', 'idia')
\end{lstlisting}
\textit{GPT-2 Medium} - Layer 16 Head 13
\begin{lstlisting}[backgroundcolor=\color{white}]
(' sugg', ' hindsight'),
(' sugg', ' anecdotal'),
(' unsuccessfully', ' hindsight'),
('didn', ' hindsight'),
('orously', 'staking'),
('illions', 'uries'),
('until', 'era'),
(' lobbied', ' hindsight'),
(' incorrectly', ' incorrect'),
(' hesitate', ' hindsight'),
('ECA', ' hindsight'),
(' regret', ' regrets'),
('inventoryQuantity', 'imore'),
('consider', ' anecdotal'),
(' errone', ' incorrect'),
(' someday', ' eventual'),
('illions', 'Murray'),
(' recently', 'recent'),
(' Learned', ' hindsight'),
('before', ' hindsight'),
(' lately', 'ealous'),
('upon', 'rity'),
('ja', ' hindsight'),
(' regretted', ' regrets'),
(' unsuccessfully', 'udging'),
(' lately', 'dated'),
(' sugg', ' anecd'),
(' inform', 'imore'),
(' lately', 'recent'),
(' anecd', ' anecdotal'),
('orously', ' hindsight'),
(' postwar', ' Era'),
(' lately', ' recent'),
(' skept', ' cynicism'),
(' sugg', 'informed'),
(' unsuccessfully', 'ealous'),
('ebin', ' hindsight'),
(' underest', ' overest'),
(' Jinn', ' hindsight'),
(' someday', '2019'),
(' recently', 'turned'),
(' sugg', ' retrospect'),
(' unsuccessfully', 'didn'),
(' unsuccessfully', 'gged'),
(' mistakenly', ' incorrect'),
('assment', ')</'),
('ja', 'didn'),
('illions', ' hindsight'),
(' sugg', ' testimony'),
('jri', ' hindsight')
\end{lstlisting}
\textit{GPT-2 Medium} - Layer 12 Head 9
\begin{lstlisting}[backgroundcolor=\color{white}]
(' PST', ' usual'),
('etimes', ' foreseeable'),
('uld', 'uld'),
(' Der', ' Mankind'),
(' statewide', ' yearly'),
(' guarantees', ' guarantees'),
(' Flynn', ' Logged'),
('borne', ' foreseeable'),
(' contiguous', ' contiguous'),
(' exceptions', ' exceptions'),
(' redist', ' costly'),
(' downstream', ' day'),
(' ours', ' modern'),
(' foreseeable', ' foreseeable'),
(' Posted', ' Posted'),
(' anecdotal', ' anecdotal'),
(' moot', ' costly'),
(' successor', ' successor'),
(' any', ' ANY'),
(' generational', ' modern'),
(' temporarily', ' costly'),
(' overall', ' overall'),
(' effective', ' incentiv'),
(' future', ' tomorrow'),
(' ANY', ' lifetime'),
(' dispatch', ' dispatch'),
(' legally', ' WARRANT'),
(' guarantees', ' incentiv'),
(' listed', ' deductible'),
(' CST', ' foreseeable'),
(' anywhere', ' any'),
(' guaranteed', ' incentiv'),
(' successors', ' successor'),
(' weekends', ' day'),
('iquid', ' expensive'),
(' Trib', ' foreseeable'),
(' phased', ' modern'),
(' constitutionally', ' foreseeable'),
(' any', ' anybody'),
(' anywhere', ' ANY'),
(' veto', ' precedent'),
(' veto', ' recourse'),
(' hopefully', ' hopefully'),
(' potentially', ' potentially'),
(' ANY', ' ANY'),
(' substantive', ' noteworthy'),
('morrow', ' day'),
('ancial', ' expensive'),
('listed', ' breastfeeding'),
(' holiday', ' holidays')
\end{lstlisting}
\textit{GPT-2 Medium} - Layer 11 Head 10
\begin{lstlisting}[backgroundcolor=\color{white}]
(' Journalism', ' acron'),
(' democracies', ' governments'),
('/-', 'verty'),
(' legislatures', ' governments'),
('ocracy', ' hegemony'),
('osi', ' RAND'),
(' Organizations', ' organisations'),
('ellectual', ' institutional'),
(' Journalists', ' acron'),
('eworks', ' sponsors'),
(' Inqu', ' reviewer'),
('ocracy', ' diversity'),
(' careers', ' Contributions'),
('gency', '\\-'),
('ellectual', ' exceptions'),
(' Profession', ' specializing'),
('online', ' Online'),
(' Publications', ' authorised'),
('Online', ' Online'),
(' sidx', ' Lazarus'),
('eworks', ' Networks'),
(' Groups', ' organisations'),
(' Governments', ' governments'),
(' democracies', ' nowadays'),
(' psychiat', ' Mechdragon'),
(' educ', ' Contributions'),
(' Ratings', ' organisations'),
('vernment', 'spons'),
('..."', '),"'),
(' Caucas', ' commodity'),
(' dictators', ' governments'),
('istration', ' sponsor'),
('iquette', ' acron'),
(' Announce', ' answ'),
(' Journalism', ' empowering'),
('Media', ' bureaucr'),
(' Discrimination', ' organizations'),
(' Journalism', 'Online'),
('FAQ', 'sites'),
(' antitrust', ' Governments'),
('..."', '..."'),
('Questions', ' acron'),
('rities', ' organisations'),
(' Editorial', ' institutional'),
(' tabl', ' acron'),
(' antitrust', ' governments'),
(' Journalism', ' Everyday'),
('icter', ' Lieberman'),
(' defect', 'SPONSORED'),
(' Journalists', ' organisations')
\end{lstlisting}
\textit{GPT-2 Medium} - Layer 22 Head 5 \footnotesize{(names and parts of names seem to attend to each other here)}
\begin{lstlisting}[backgroundcolor=\color{white}]
(' Smith', 'ovich'),
(' Jones', 'ovich'),
(' Jones', 'Jones'),
(' Smith', 'Williams'),
(' Rogers', 'opoulos'),
('Jones', 'ovich'),
(' Jones', 'inez'),
('ug', ' Ezek'),
(' Moore', 'ovich'),
('orn', 'roit'),
('van', 'actionDate'),
(' Jones', 'inelli'),
(' Edwards', 'opoulos'),
(' Jones', ' Lyons'),
('Williams', 'opoulos'),
('Moore', 'ovich'),
(' Rodriguez', 'hoff'),
(' North', ' suburbs'),
(' Smith', 'chio'),
('Smith', 'ovich'),
(' Smith', 'opoulos'),
('Mc', 'opoulos'),
('Johnson', 'utt'),
(' Jones', 'opoulos'),
('Ross', 'Downloadha'),
('pet', 'ilage'),
(' Everett', ' Prairie'),
(' Cass', 'isma'),
(' Jones', 'zynski'),
('Jones', 'Jones'),
(' McCl', 'elman'),
(' Smith', 'Jones'),
(' Simmons', 'opoulos'),
(' Smith', 'brown'),
(' Mc', 'opoulos'),
(' Jones', 'utt'),
(' Richards', 'Davis'),
(' Johnson', 'utt'),
(' Ross', 'bred'),
(' McG', 'opoulos'),
(' Stevens', 'stadt'),
('ra', 'abouts'),
(' Johnson', 'hoff'),
(' North', ' Peninsula'),
(' Smith', 'Smith'),
('Jones', 'inez'),
(' Hernandez', 'hoff'),
(' Lucas', 'Nor'),
(' Agu', 'hoff'),
('Jones', 'utt')
\end{lstlisting}
\textit{GPT-2 Medium} - Layer 19 Head 12
\begin{lstlisting}[backgroundcolor=\color{white}]
(' 2015', 'ADVERTISEMENT'),
(' 2014', '2014'),
(' 2015', '2014'),
(' 2015', 'Present'),
(' 2013', '2014'),
(' 2017', 'ADVERTISEMENT'),
(' 2016', 'ADVERTISEMENT'),
('itor', ' Banner'),
('2015', ' Bulletin'),
('2012', ' Bulletin'),
('2014', ' Bulletin'),
(' Airl', 'Stream'),
('2016', ' Bulletin'),
(' 2016', '2014'),
('2017', ' Bulletin'),
(' 2013', ' 2014'),
(' 2012', '2014'),
(' stadiums', 'ventions'),
(' 2015', ' Bulletin'),
('2013', ' Bulletin'),
(' 2017', '2014'),
(' 2011', ' 2011'),
(' 2014', ' 2014'),
(' 2011', ' 2009'),
(' mile', 'eming'),
(' 2013', 'ADVERTISEMENT'),
(' 2014', '2015'),
(' 2014', 'Present'),
(' 2011', '2014'),
(' 2011', '2009'),
(' 2015', ' 2014'),
(' 2013', ' Bulletin'),
(' 2015', '2015'),
(' 2011', ' 2003'),
(' 2011', ' 2010'),
(' 2017', 'Documents'),
('2017', 'iaries'),
(' 2013', '2015'),
('2017', 'Trend'),
(' 2011', '2011'),
(' 2016', 'Present'),
(' 2011', ' 2014'),
(' years', 'years'),
('Plug', 'Stream'),
(' 2014', 'ADVERTISEMENT'),
('2015', 'Present'),
(' 2018', 'thora'),
(' 2017', 'thora'),
(' 2012', ' 2011'),
(' 2012', ' 2014')
\end{lstlisting}

\subsection{Feedforward Keys and Values}
\label{appendix:samples_ffn}
Key-value pairs, $(k_i, v_i)$, where at least 15\% of the top-$k$ vocabulary items overlap, with $k=100$. We follow our forerunner's convention of calling the index of the value in the layer ``dimension'' (Dim). 

Here again we use two asterisks (\textsuperscript{**}) to represent lists where we discarded tokens outside the corpus vocabulary. 
\textit{GPT-2 Medium} -  Layer 0 Dim 116
\begin{lstlisting}[backgroundcolor=\color{white}]
#annels        #Els
#netflix       #osi
telev          #mpeg
#tv            #vous
#avi           #iane
#flix          transmitter
Television     Sinclair
#outube        Streaming
#channel       #channel
Vid            mosqu
#Channel       broadcaster
documentaries  airs
#videos        Broadcasting
Hulu           broadcasts
channels       streams
#levision      channels
DVDs           broadcasters
broadcasts     broadcasting
#azeera        #RAFT
MPEG           #oded
televised      htt
aired          transmissions
broadcasters   playback
Streaming      Instruction
viewership     nic
#TV            Sirius
Kodi           viewership
ITV            radio
#ovies         #achers
channel        channel
\end{lstlisting}
\textit{GPT-2 Medium} -  Layer 3 Dim 2711
\begin{lstlisting}[backgroundcolor=\color{white}]
purposes     purposes
sake         sake
purpose      reasons
reasons      purpose
convenience  ages
reason       reason
Seasons      #ummies
#Plex        #going
Reasons      foreseeable
#ummies      Reasons
#asons       #reason
#lation      #pur
#alsh        Developers
#agos        #akers
#ACY         transl
STATS        Reason
#itas        consideration
ages         #purpose
#purpose     beginners
#=[          awhile
#gencies     Pur
Millennium   #benefit
Brewers      #atel
Festival     #tun
EVENT        pur
#payment     Ages
#=-          preservation
#printf      Metatron
beginners    um
Expo         #KEN
\end{lstlisting}
\textit{GPT-2 Medium} -  Layer 4 Dim 621
\begin{lstlisting}[backgroundcolor=\color{white}]
#ovie          headlined
newspapers     pestic
television     dime
editorial      describ
#journal       Afric
broadcasters   broadcasts
#Journal       #('
publication    #umbnails
Newsweek       #adish
Zeit           #uggest
columnist      splash
Editorial      #ZX
newsletter     objectionable
cartoon        #article
#eport         Bucc
telev          #London
radio          reprint
headlined      #azine
#ribune        Giov
BBC            #ender
reprint        headline
sitcom         #oops
reprinted      #articles
broadcast      snipp
tabloid        Ajax
documentaries  marqu
journalist     #("
TV             #otos
headline       mast
news           #idem
\end{lstlisting}
\textit{GPT-2 Medium} -  Layer 7 Dim 72
\begin{lstlisting}[backgroundcolor=\color{white}]
sessions      session
dinners       sessions
#cation       #cation
session       #iesta
dinner        Booth
#eteria       screenings
Dinner        booked
#Session      #rogram
rehears       vacation
baths         baths
Lunch         #pleasant
#hops         meetings
visits        #Session
Session       greet
#session      #athon
meetings      Sessions
chatting      boarding
lunch         rituals
chats         booking
festivities   Grape
boarding      #miah
#workshop     #session
#rooms        Pars
#tests        simulated
seated        Dispatch
visit         Extras
appointments  toile
#vu           Evening
#rations      showers
#luaj         abroad
\end{lstlisting}
\textit{GPT-2 Medium} -  Layer 10 Dim 8
\begin{lstlisting}[backgroundcolor=\color{white}]
Miy     Tai
#imaru  #jin
Gong    Jin
Jinn    Makoto
Xia     #etsu
Makoto  Shin
Kuro    Hai
Shin    Fuj
#Tai    Dai
Yamato  Miy
Tai     #iku
Ichigo  Yun
#Shin   Ryu
#atsu   Shu
Haku    Hua
Chun    Suzuki
#ku     Yang
Qing    Xia
Tsuk    #Shin
Hua     #iru
Jiang   Yu
Nanto   #yu
manga   Chang
Yosh    Nan
yen     Qian
Osaka   #hao
Qian    Fuk
#uku    Chun
#iku    Yong
Yue     #Tai
\end{lstlisting}
\textit{GPT-2 Medium} -  Layer 11 Dim 2
\begin{lstlisting}[backgroundcolor=\color{white}]
progressing  toward
#Progress    towards
#progress    Pace
#osponsors   progression
#oppable     #inness
advancement  onward
progress     canon
Progress     #progress
#senal       pace
#venge       #peed
queue        advancement
#pun         advancing
progression  progressing
#wagon       ladder
advancing    path
#cknowled    honoring
#Goal        ranks
momentum     standings
#zag         goal
#hop         #grand
pursuits     momentum
#encing      #ometer
#Improve     timetable
STEP         nearing
#chini       quest
standings    spiral
#eway        trajectory
#chie        progress
#ibling      accelerating
Esports      escal
\end{lstlisting}
\textit{GPT-2 Medium} -  Layer 15 Dim 4057
\begin{lstlisting}[backgroundcolor=\color{white}]
EDITION       copies
versions      Version
copies        #edition
version       #Version
Version       version
edition       #download
editions      download
reprint       versions
#edition      #Download
EDIT          copy
Edition       #release
reproduce     #version
originals     release
#edited       #copy
VERS          VERS
#Versions     #pub
#Publisher    Download
reprodu       #released
#uploads      editions
playthrough   edition
Printed       reprint
reproduction  Release
#Reviewed     #Available
copy          #published
#Version      #Published
paperback     EDITION
preview       print
surv          #Quantity
#Download     #available
circulate     RELEASE
\end{lstlisting}
\textit{GPT-2 Medium} -  Layer 16 Dim 41
\begin{lstlisting}[backgroundcolor=\color{white}]
#duino          alarm
#Battery        alarms
Morse           signal
alarms          circuit
GPIO            GPIO
LEDs            timers
batteries       voltage
#toggle         signals
signal          circuitry
circuitry       electrical
#PsyNetMessage  circuits
alarm           LEDs
autop           standby
signalling      signalling
#volt           signaling
volt            lights
signals         Idle
voltage         triggers
LED             batteries
electrom        Morse
timers          LED
malfunction     #LED
amplifier       button
radios          Signal
wiring          timer
#Alert          wiring
signaling       buzz
#Clock          disconnect
arming          Arduino
Arduino         triggered
\end{lstlisting}
\textit{GPT-2 Medium} -  Layer 17 Dim 23
\begin{lstlisting}[backgroundcolor=\color{white}]
responsibility    responsibility
Responsibility    respons
responsibilities  responsibilities
#ipolar           Responsibility
#responsible      oversee
duties            #respons
#respons          duties
superv            supervision
supervision       superv
#abwe             stewards
Adin              chore
respons           oversight
oversee           oversees
entrusted         responsible
overseeing        #responsible
helicop           handling
presided          handles
overseen          overseeing
#dyl              chores
responsible       manage
#ADRA             managing
reins             duty
#accompan         Respons
chores            charge
oversees          reins
supervised        handle
blame             oversaw
oversaw           CONTROL
#archment         RESP
RESP              tasks
\end{lstlisting}
\textit{GPT-2 Medium} -  Layer 19 Dim 29
\begin{lstlisting}[backgroundcolor=\color{white}]
subconscious  thoughts
thoughts      thought
#brain        Thoughts
#Brain        minds
memories      mind
OCD           thinking
flashbacks    #thought
brainstorm    imagination
Anxiety       Thinking
#mind         Thought
fantas        imagin
amygdala      thinker
impuls        #thinking
Thinking      #mind
#Memory       memories
Thoughts      #think
dreams        imagining
#ocamp        impulses
#Psych        fantasies
#mares        think
mentally      urges
#mental       desires
mind          dreams
#thinking     delusions
#Mind         subconscious
#dream        emotions
psyche        imag
prefrontal    #dream
PTSD          conscience
Memories      visions
\end{lstlisting}
\textit{GPT-2 Medium} -  Layer 20 Dim 65
\begin{lstlisting}[backgroundcolor=\color{white}]
exercises    volleyball
#Sport       tennis
#athlon      sports
Exercise     sport
#ournaments  #basketball
volleyball   Tennis
Recre        soccer
Mahjong      golf
#basketball  playground
exercise     Golf
bowling      athletics
skating      #athlon
spar         athletic
skiing       rugby
gymn         amusement
#sports      gymn
drills       sled
#Training    #Sport
tournaments  cricket
sled         Soccer
Volunte      amuse
skate        Activities
golf         recreational
#Pract       Ski
dunk         activities
#hower       basketball
athletics    #games
sport        skating
Solitaire    hockey
#BALL        #sports
\end{lstlisting}
\textit{GPT-2 Medium} -  Layer 21 Dim 86
\begin{lstlisting}[backgroundcolor=\color{white}]
IDs              number
identifiers      #number
surname          #Number
surn             Number
identifier       NUM
initials         numbers
#Registered      Numbers
NAME             #Numbers
#names           address
pseudonym        #address
#codes           #Num
nomine           #NUM
names            addresses
username         Address
#IDs             identifier
ID               #Address
registration     #num
#76561           ID
#soDeliveryDate  numbering
#ADRA            IDs
CLSID            #ID
numbering        identifiers
#ername          identification
#address         numer
addresses        digits
codes            #numbered
#Names           numerical
regist           Ident
name             numeric
Names            Identification
\end{lstlisting}
\textit{GPT-2 Medium} -  Layer 21 Dim 400
\begin{lstlisting}[backgroundcolor=\color{white}]
#July       Oct
July        Feb
#February   Sept
#January    Dec
#Feb        Jan
November    Nov
#October    Aug
January     #Oct
Feb         May
October     #Nov
#September  Apr
September   March
#June       April
#Sept       #Sept
February    June
#November   #Aug
#April      October
April       #Feb
June        July
#December   December
August      Sep
#March      November
Sept        #Jan
December    #May
Aug         August
March       Jul
#August     Jun
#Aug        September
#wcs        January
Apr         February
\end{lstlisting}
\textit{GPT-2 Medium} -  Layer 23 Dim 166
\begin{lstlisting}[backgroundcolor=\color{white}]
#k       #k
#ks      #K
#kish    #ks
#K       #KS
#kat     k
#kus     #kt
#KS      K
#ked     #kr
#kr      #kl
#kB      #kish
#kan     #kos
#kw      #king
#ket     #ked
#king    #kie
#kb      #KB
#kos     #kk
#kHz     #kowski
#kk      #KR
#kick    #KING
#kers    #KT
#kowski  #KK
#KB      #KC
#krit    #kw
#KING    #kb
#kt      #Ka
#ksh     #krit
#kie     #KN
#ky      #kar
#KY      #kh
#ku      #ket
\end{lstlisting}
\textit{GPT-2 Medium} -  Layer 23 Dim 907
\begin{lstlisting}[backgroundcolor=\color{white}]
hands       hand
hand        #Hand
#hands      Hand
#hand       #hand
fingers     hands
#feet       Hands
fingertips  fist
claws       #hands
paw         finger
paws        handed
metab       thumb
palms       fingers
fingert     foot
#Hand       #handed
fists       paw
wrists      handing
levers      #finger
thumbs      #hander
tentacles   fingertips
feet        claw
limb        fingert
slider      #Foot
#handed     Stick
#dimension  arm
jaws        #Accessory
skelet      #fing
lapt        Foot
ankles      index
weap        toe
foot        #auntlet
\end{lstlisting}
\textit{GPT-2 Large} -  Layer 25 Dim 2685\textsuperscript{**}
\begin{lstlisting}[backgroundcolor=\color{white}]
#manager       engineering
#Engineers     Marketing
chemist        #engineering
humanities     Communications
sciences       #communications
anthropology   anthropology
lingu          Engineering
#engineering   lingu
psychologist   psychology
Coordinator    neurolog
Analyst        Economics
#iologist      designer
accountant     sociology
strategist     communications
#ographer      marketing
curator        pharmac
Engineers      sciences
archae         economics
Designer       Accounting
Editing        #econom
biologist      chemist
#ologist       merch
psychologists  pharm
theolog        economist
Marketing      architect
#Manager       engineer
Architects     Architect
sociology      #technical
engineer       architects
physicist      logistics
\end{lstlisting}
\textit{GPT-2 Large} -  Layer 21 Dim 3419\textsuperscript{**}
\begin{lstlisting}[backgroundcolor=\color{white}]
#overty         impoverished
#wana           poverty
poverty         poorest
#Saharan        poorer
poorest         Yemen
Poverty         families
malnutrition    Poverty
Senegal         marginalized
impoverished    refugees
#poor           subsistence
Gujar           displaced
homelessness    hardship
Homeless        refugee
#heid           households
Ramadan         migrant
#Palest         disadvantaged
poorer          Sudan
Rahman          oppressed
#amily          socioeconomic
illiter         peasant
Mahmoud         homeless
Haitian         poor
#advertisement  Ethiopian
#hya            Kaf
#African        Rw
wealthier       #poor
Africans        Af
caste           rural
homeless        #fam
Hait            needy
\end{lstlisting}
\textit{GPT-2 Large} -  Layer 25 Dim 2442\textsuperscript{**}
\begin{lstlisting}[backgroundcolor=\color{white}]
Tracker      tracking
gau          Tracker
charts       tracker
tracker      Tracking
#Measure     quant
measurement  #Stats
measuring    gau
#Tracker     GPS
gauge        Track
tracking     estimating
Tracking     tally
#Monitor     #ometers
#chart       tracked
Meter        calculate
#HUD         calculating
#ometers     measurement
surve        gauge
#Stats       estimation
#Statistics  monitoring
calculate    #stats
Measure      #tracking
quant        track
#asuring     measuring
Calculator   Monitoring
#ometer      #Detailed
calculator   #ometer
Monitoring   estim
#Maps        stats
pione        charts
timet        timet
\end{lstlisting}
\textit{GPT-2 Base} -  Layer 9 Dim 1776
\begin{lstlisting}[backgroundcolor=\color{white}]
radios              cable
antennas            modem
radio               wireless
modem               WiFi
voltage             wired
transformer         broadband
Ethernet            Ethernet
telev               radios
#Radio              power
electricity         radio
loudspe             Cable
kW                  Wireless
#radio              telephone
broadband           network
volt                signal
microphones         Networks
telecommunications  networks
cable               electricity
Telephone           wifi
amplifier           #levision
wifi                coax
broadcasting        transmit
transistor          transmitter
Radio               TV
wireless            Network
LTE                 television
watts               transmission
microwave           router
telephone           cables
amps                amplifier
\end{lstlisting}
\textit{GPT-2 Base} -  Layer 9 Dim 2771
\begin{lstlisting}[backgroundcolor=\color{white}]
arous        increase
freeing      increasing
incent       accelerating
stimulate    allev
induce       exped
discourage   enhanced
inducing     aggrav
mitigating   enhance
stimulating  inhib
emanc        improving
alleviate    infl
empowering   #oint
preventing   alien
#ufact       alter
#HCR         enabling
influencing  incre
handc        indu
disadvant    #Impro
#roying      intens
arresting    improve
allev        easing
weaken       elevate
depri        encouraging
dissu        accelerate
impede       enlarg
convol       energ
encouraging  accent
#xiety       acceler
#akening     depri
lowering     elong
\end{lstlisting}
\textit{GPT-2 Base} -  Layer 1 Dim 2931
\begin{lstlisting}[backgroundcolor=\color{white}]
evening         week
#shows          evening
night           night
#sets           morning
#lav            afternoon
afternoon       month
#/+             #'s
Night           #naissance
Loll            #genre
Kinnikuman      semester
Weekend         #ched
morning         #ague
#enna           weekend
Saturday        latest
Sunday          #cher
week            #EST
Blossom         #icter
#Night          happens
#atto           day
#vertising      happened
#spr            #essim
#Sunday         Masquerade
#morning        #ished
#Thursday       sounded
Week            #ching
Panc            pesky
Evening         #chy
#allery         trope
#ADVERTISEMENT  #feature
#Street         #fy
\end{lstlisting}
\textit{GPT-2 Base} -  Layer 0 Dim 1194
\begin{lstlisting}[backgroundcolor=\color{white}]
Pay           receipts
#Pay          depos
refund        Deposit
police        deduct
#pay          #milo
#paying       #igree
#Tax          #eln
debit         levied
PayPal        deposit
ATM           #enforcement
cops          endot
tax           #soType
ID            paperwork
#payment      deposits
payment       loopholes
checkout      waivers
#police       receipt
agents        waive
DMV           loophole
application   arresting
card          commissioner
applications  Forms
office        transporter
arrested      Dupl
#paid         confisc
pay           Clapper
#tax          #ventures
RCMP          #Tax
PAY           whistleblowers
APPLIC        #ADRA
\end{lstlisting}
\textit{GPT-2 Base} -  Layer 9 Dim 2771
\begin{lstlisting}[backgroundcolor=\color{white}]
flaws            flaws
lurking          weaknesses
failings         dangers
vulnerabilities  scams
inaccur          shortcomings
scams            pitfalls
shortcomings     injust
flawed           faults
glitches         flawed
pitfalls         abuses
inconsistencies  imperfect
rigged           lurking
biases           wrongdoing
deficiencies     corruption
weaknesses       inaccur
discrepancies    inadequ
hypocrisy        fraud
rigging          inequ
deceptive        weakness
misinformation   scam
#urities         hazards
lur              problematic
imperfect        hoax
regress          danger
#abase           failings
#errors          problems
#lived           injustice
abuses           plagiar
misinterpret     plag
suspic           deceptive
\end{lstlisting}

 \subsection{Knowledge Lookup}
 \label{appendix:examples_lookup}
 Given a few seed embeddings of vocabulary items we find related FF values by taking a product of the average embeddings with FF values.

 Seed vectors: \\
 \verb|["python", "java", "javascript"]|
 \\
  Layer 14 Dim 1215 (ranked 3rd) \begin{lstlisting}[backgroundcolor=\color{white}]
filesystem
debugging
Windows
HTTP
configure
Python
debug
config
Linux
Java
configuration
cache
Unix
lib
runtime
kernel
plugins
virtual
FreeBSD
hash
plugin
header
file
server
PHP
GNU
headers
Apache
initialization
Mozilla
\end{lstlisting}
Seed vectors: \verb|["cm", "kg", "inches"]|\\
 Layer 20 Dim 2917 (ranked 1st) \begin{lstlisting}[backgroundcolor=\color{white}]
percent
years
hours
minutes
million
seconds
inches
months
miles
weeks
pounds
#%
kilometers
ounces
kilograms
grams
kilometres
metres
centimeters
thousand
days
km
yards
Years
meters
#million
acres
kg
#years
inch\end{lstlisting} 
Seed vectors: \verb|["horse", "dog", "lion"]|
\\
Layer 21 Dim 3262 (ranked 2nd) \begin{lstlisting}[backgroundcolor=\color{white}]
animal
animals
Animal
dogs
horse
wildlife
Animals
birds
horses
dog
mammal
bird
mammals
predator
beasts
Wildlife
species
#Animal
#animal
Dogs
fish
rabbits
deer
elephants
wolves
pets
veterinary
canine
beast
predators
reptiles
rodent
primates
hunting
livestock
creature
rabbit
rept
elephant
creatures
human
hunters
hunter
shark
Rept
cattle
wolf
Humane
tiger
lizard\end{lstlisting} 
  
\onecolumn
\section{Sentiment Analysis Fine-Tuning Vector Examples}
\label{appendix:finetune_examples}

{\color{red} \textbf{This section contains abusive language}}
{
\setlength{\tabcolsep}{1cm}
\subsection*{Classification Head Parameters}
Below we show the finetuning vector of the classifier weight. ``POSITIVE'' designates the vector corresponding to the label ``POSITIVE'', and similarly for ``NEGATIVE''. 
\begin{lstlisting}[backgroundcolor=\color{white}, mathescape=true]
POSITIVE     $\textbf{NEGATIVE}$
-----------  ------------
#yssey       bullshit
#knit        lame
#etts        crap
passions     incompetent
#etooth      inco
#iscover     bland
pioneers     incompetence
#emaker      idiots
Pione        crappy
#raft        shitty
#uala        idiot
prosper      pointless
#izons       retarded
#encers      worse
#joy         garbage
cherish      CGI
loves        FUCK
#accompan    Nope
strengthens  useless
#nect        shit
comr         mediocre
honoured     poorly
insepar      stupid
embraces     inept
battled      lousy
#Together    fuck
intrig       sloppy
#jong        Worse
friendships  Worst
#anta        meaningless
\end{lstlisting}
In the following sub-sections, we sample 4 difference vectors per each parameter group (FF keys, FF values; attention query, key, value, and output subheads), and each one of the fine-tuned layers (layers 9-11). We present the ones that seemed to contain relevant patterns upon manual inspection. We also report the number of ``good'' vectors among the four sampled vectors for each layer and parameter group.

\subsection*{FF Keys}
\textbf{Layer 9}

4 out of 4

\noindent\begin{longtable}{ll}
\begin{lstlisting}[backgroundcolor=\color{white}, mathescape=true]
$\textbf{diff}$         -diff
-----------  ---------------
amazing      seiz
movies       coerc
wonderful    Citiz
love         #cffff
movie        #GBT
cinematic    targ
enjoyable    looph
wonderfully  Procedures
beautifully  #iannopoulos
enjoy        #Leaks
films        #ilon
comedy       grievance
fantastic    #merce
awesome      Payments
#Enjoy       #RNA
cinem        Registrar
film         Regulatory
loving       immobil
enjoyment    #bestos
masterpiece  #SpaceEngineers
\end{lstlisting}
& 
\begin{lstlisting}[backgroundcolor=\color{white}, mathescape=true]
diff             $\textbf{-diff}$
---------------  ------------
reperto          wrong
congratulations  unreasonable
Citation         horribly
thanks           inept
Recording        worst
rejo             egregious
Profile          #wrong
Tradition        unfair
canopy           worse
#ilion           atro
extracts         stupid
descendant       egreg
#cele            bad
enthusiasts      terribly
:-)              ineffective
#photo           nonsensical
awaits           awful
believer         #worst
#IDA             incompetence
welcomes         #icably
\end{lstlisting}
\\
\begin{lstlisting}[backgroundcolor=\color{white}, mathescape=true]
$\textbf{diff}$     -diff
-------  ----------
movie    seiz
fucking  Strongh
really   #etooth
movies   #20439
damn     #Secure
funny    Regulation
shit     Quarterly
kinda    concess
REALLY   Recep
Movie    #aligned
stupid   targ
#movie   mosqu
goddamn  #verning
crap     FreeBSD
shitty   PsyNet
film     Facilities
crappy   #Lago
damned   #Register
#Movie   #"}],"
cheesy   Regist
\end{lstlisting} 
& 
\begin{lstlisting}[backgroundcolor=\color{white}, mathescape=true]
$\textbf{diff}$          -diff
------------  ------------
incompetence  #knit
bullshit      #Together
crap          Together
useless       versatile
pointless     #Discover
incompetent   richness
idiots        #iscover
incompet      forefront
garbage       inspiring
meaningless   pioneering
stupid        #accompan
crappy        unparalleled
shitty        #Explore
nonexistent   powerfully
worthless     #"},{"
Worse         #love
lame          admired
worse         #uala
inco          innovative
ineffective   enjoyed
\end{lstlisting}

\end{longtable}

\textbf{Layer 10}

4 out of 4

\noindent\begin{longtable}{ll}
\begin{lstlisting}[backgroundcolor=\color{white}, mathescape=true]
diff             $\textbf{-diff}$
---------------  -------------
quotas           wonderfully
#RNA             wonderful
cessation        beautifully
subsidy          amazing
#SpaceEngineers  fantastic
placebo          incredible
exemptions       amazingly
treadmill        great
Labs             unforgettable
receipt          beautiful
moratorium       brilliantly
designation      hilarious
ineligible       love
reimbursement    marvelous
roundup          vividly
Articles         terrific
PubMed           memorable
waivers          #Enjoy
Citiz            loving
landfill         fascinating
\end{lstlisting}
&
\begin{lstlisting}[backgroundcolor=\color{white}, mathescape=true]
diff                $\textbf{-diff}$
------------------  -------------
isEnabled           wonderfully
guiActiveUnfocu...  beautifully
#igate              cinem
waivers             cinematic
expires             wonderful
expire              amazing
reimb               Absolutely
expired             storytelling
#rollment           fantastic
#Desktop            Definitely
prepaid             unforgettable
#verning            comedy
#andum              movie
reimbursement       comedic
Advisory            hilarious
permitted           #movie
#pta                #Amazing
issuance            scenes
Priebus             Amazing
#iannopoulos        enjoyable
\end{lstlisting} \\
\begin{lstlisting}[backgroundcolor=\color{white}, mathescape=true]
$\textbf{diff}$          -diff
------------  ----------
horror        #deals
whim          #iband
subconscious  [&
unrealistic   #heid
imagination   #APD
viewers       withdrew
enjoyment     #Shares
nostalgia     mathemat
absolute      [+]
sentimental   #Tracker
unreal        #zb
Kubrick       testified
awe           #ymes
inspiration   mosqu
subtle        #Commerce
cinematic     administr
perfection    feder
comedic       repaired
fantasy       #pac
mindless      #Community
\end{lstlisting}
&
\begin{lstlisting}[backgroundcolor=\color{white}, mathescape=true]
diff           $\textbf{-diff}$
-------------  -------------
#Leaks         loving
quotas         love
#RNA           loved
subsidy        lovers
#?'"           wonderful
Penalty        lover
#iannopoulos   nostalgic
#>]            alot
discredited    beautiful
#conduct       amazing
#pta           great
waivers        passionate
Authorization  admire
#admin         passion
HHS            lovely
arbitrarily    loves
#arantine      unforgettable
#ERC           proud
memorandum     inspiration
#Federal       #love
\end{lstlisting}
\end{longtable}

\textbf{Layer 11}

4 out of 4

\noindent\begin{longtable}{ll}
\begin{lstlisting}[backgroundcolor=\color{white}, mathescape=true]
$\textbf{diff}$         -diff
-----------  -----------
inco         cherish
pointless    #knit
Nope         #terday
bullshit     #accompan
crap         prosper
useless      versatile
nonsense     friendships
futile       #uala
anyways      Lithuan
anyway       cherished
meaningless  redes
clueless     inspires
lame         Proud
wasting      friendship
bogus        exceptional
vomit        #beaut
nonsensical  #ngth
retarded     pioneering
idiots       pioneers
shit         nurt
\end{lstlisting}
& 
\begin{lstlisting}[backgroundcolor=\color{white}, mathescape=true]
diff             $\textbf{-diff}$
---------------  -----------
#SpaceEngineers  love
nuisance         definitely
#erous           always
#aband           wonderful
Brist            loved
racket           wonderfully
Penalty          cherish
bystand          loves
#iannopoulos     truly
Citiz            enjoy
Codec            really
courier          #olkien
#>]              beautifully
#termination     #love
incapac          great
#interstitial    LOVE
fugitive         never
breaching        adore
targ             loving
thug             amazing
\end{lstlisting}
\\
\begin{lstlisting}[backgroundcolor=\color{white}, mathescape=true]
diff         $\textbf{-diff}$
-----------  ------------
#accompan    bad
Pione        crap
celebrate    inefficient
#Discover    stupid
#knit        worse
pioneering   mistake
recogn       incompetence
reunited     mistakes
comr         incompetent
thriving     miser
#iscover     garbage
commemorate  retarded
Remem        #bad
ecstatic     poor
forefront    ineffective
enthusi      retard
renewed      Poor
colle        bullshit
Inspired     inept
#uala        errors
\end{lstlisting}
&
\begin{lstlisting}[backgroundcolor=\color{white}, mathescape=true]
diff          $\textbf{-diff}$
------------  ------------
#knit         bullshit
passions      crap
#accompan     idiots
#ossom        goddamn
#Explore      stupid
welcomes      shitty
pioneering    shit
forefront     garbage
embraces      fuck
pioneers      incompetence
intertw       crappy
#izons        bogus
#iscover      useless
unparalleled  idiot
evolving      #shit
Together      pointless
vibrant       stupidity
prosper       fucking
strengthens   nonsense
#Together     FUCK
\end{lstlisting}

\end{longtable}
\subsection*{FF Values}
\textbf{Layer 9}

0 out of 4

\textbf{Layer 10}

0 out of 4

\textbf{Layer 11}

0 out of 4

\subsection*{$\WQ$ Subheads}
\textbf{Layer 9}

3 out of 4
\noindent \begin{longtable}{ll}
\begin{lstlisting}[backgroundcolor=\color{white}, mathescape=true]
diff          $\textbf{-diff}$
------------  ------------
#ARGET        kinda
#idal         alot
#--+          amazing
Prev          interesting
#enger        wonderful
#iannopoulos  definitely
#report       unbelievable
#RELATED      really
issuance      amazingly
#earcher      pretty
Previous      nice
Legislation   absolutely
#astical      VERY
#iper         wonderfully
#>[           incredible
#</           hilarious
Vendor        funny
#">           fantastic
#phrine       quite
#wcsstore     defin
\end{lstlisting}
& 
\begin{lstlisting}[backgroundcolor=\color{white}, mathescape=true]
$\textbf{diff}$          -diff
------------  -----------
bullshit      strengthens
bogus         Also
faux          #helps
spurious      adjusts
nonsense      #ignt
nonsensical   evolves
inept         helps
crap          grew
junk          grows
shitty        #cliffe
fake          recognizes
incompetence  #assadors
crappy        regulates
phony         flourished
sloppy        improves
dummy         welcomes
mediocre      embraces
lame          gathers
outrage       greets
inco          prepares
\end{lstlisting} 
\\
\begin{lstlisting}[backgroundcolor=\color{white}, mathescape=true]
$\textbf{diff}$        -diff
----------  ------------
alot        Provision
kinda       coerc
amazing     Marketable
definitely  contingency
pretty      #Dispatch
tho         seiz
hilarious   #verning
VERY        #iannopoulos
really      #Reporting
lol         #unicip
wonderful   Fiscal
thats       issuance
dont        provision
pics        #Mobil
doesnt      #etooth
underrated  policymakers
funny       credential
REALLY      Penalty
#love       #activation
alright     #Officials
\end{lstlisting}
\end{longtable}

\textbf{Layer 10}

4 out of 4
\noindent\begin{longtable}{ll}
\begin{lstlisting}[backgroundcolor=\color{white}, mathescape=true]
$\textbf{diff}$      -diff
--------  ---------
crap      #Register
shit      Browse
bullshit  #etooth
stupid    #ounces
shitty    #verning
horrible  #raft
awful     #egu
fucking   #Lago
comedic   Payments
crappy    #orsi
cheesy    Coinbase
comedy    #ourse
fuck      #iann
mediocre  #"}],"
terrible  #onductor
movie     #obil
bad       #rollment
gimmick   #ivot
filler    #Secure
inept     #ETF
\end{lstlisting}
& 
\begin{lstlisting}[backgroundcolor=\color{white}, mathescape=true]
$\textbf{diff}$           $\textbf{-diff}$
-------------  ------------
love           Worse
unforgettable  Nope
beautiful      #Instead
loved          Instead
#love          #Unless
loving         incompetence
amazing        incapable
#joy           Unless
inspiring      #failed
passion        incompet
adventure      incompetent
loves          ineffective
excitement     #Fuck
joy            #Wr
LOVE           inept
together       spurious
memories       #Failure
wonderful      worthless
enjoyment      obfusc
themes         inadequate
\end{lstlisting}
\\ 
\begin{lstlisting}[backgroundcolor=\color{white}, mathescape=true]
diff                $\textbf{-diff}$
------------------  ------------
#knit               crap
#"},{"              bullshit
#"}],"              stupid
#estones            inept
#Learn              shit
#ounces             idiots
#egu                shitty
#Growing            crappy
#ributes            incompetence
#externalAction...  fuck
#encers             pointless
Browse              nonsense
jointly             nonsensical
Growing             stupidity
#ossom              gimmick
honoured            inco
#accompan           lame
#agos               incompetent
#raft               mediocre
#iership            bland
\end{lstlisting}
& 
\begin{lstlisting}[backgroundcolor=\color{white}, mathescape=true]
$\textbf{diff}$       -diff
---------  -----------
crap       #egu
bullshit   #etooth
shit       #verning
:(         #ounces
lol        #accompan
stupid     coh
filler     #assadors
shitty     #pherd
fucking    #acio
pointless  #uchs
idiots     strengthens
anyways    #reprene
nonsense   Scotia
anyway     #rocal
crappy     reciprocal
stupidity  Newly
fuck       fost
#shit      #ospons
anymore    #onductor
Nope       governs
\end{lstlisting}

\end{longtable}

\textbf{Layer 11}

3 out of 4
\noindent\begin{longtable}{ll}
\begin{lstlisting}[backgroundcolor=\color{white}, mathescape=true]
diff           $\textbf{-diff}$
-------------  ------------
#utterstock    amazing
#ARGET         movie
#cffff         alot
#etooth        scenes
#Federal       comedy
POLITICO       movies
#Register      cinematic
#Registration  greatness
#rollment      wonderful
#ETF           storytelling
#ulia          film
Payments       tho
#IRC           masterpiece
Regulatory     films
Alternatively  Kubrick
#RN            realism
#pta           comedic
Regulation     cinem
#GBT           #movie
#":""},{"      genre
\end{lstlisting}
& 
\begin{lstlisting}[backgroundcolor=\color{white}, mathescape=true]
diff           $\textbf{-diff}$
-------------  ------------------
#also          meaningless
#knit          incompetence
helps          inco
strengthens    pointless
:)             incompetent
broaden        Worse
#ossom         inept
incorporates   nonsensical
#Learn         coward
incorporate    unint
#"},{"         obfusc
enjoy          excuses
enjoyed        panicked
complementary  useless
#etts          bullshit
enhances       stupid
integrates     incompet
#ospons        incomprehensibl...
differs        stupidity
#arger         lifeless
\end{lstlisting}
\\
\begin{lstlisting}[backgroundcolor=\color{white}, mathescape=true]
$\textbf{diff}$           -diff
-------------  ---------------
amazing        #iannopoulos
beautifully    expired
love           ABE
wonderful      Yiannopoulos
wonderfully    liability
unforgettable  #SpaceEngineers
beautiful      #isance
loving         Politico
#love          waivers
#beaut         #utterstock
enjoyable      excise
#Beaut         #Stack
inspiring      phantom
fantastic      PubMed
defin          #ilk
incredible     impunity
memorable      ineligible
greatness      Coulter
amazingly      issuance
timeless       IDs
\end{lstlisting}
\end{longtable}

\subsection*{$\WK$ Subheads}
\textbf{Layer 9}

3 out of 4

\noindent\begin{longtable}{ll}
\begin{lstlisting}[backgroundcolor=\color{white}, mathescape=true]
diff     $\textbf{-diff}$
-------  ----------
enclave  horrible
#.       pretty
#;       alot
#omial   MUCH
apiece   VERY
#assian  nothing
#.</     #much
#ulent   terrible
#,[      crappy
#eria    strange
#ourse   everything
exerc    very
#\/      shitty
#Wire    nice
#arium   many
#icle    wonderful
#.[      genuinely
#/$\$$      beautiful
#API     much
#ium     really
\end{lstlisting}
&
\begin{lstlisting}[backgroundcolor=\color{white}, mathescape=true]
$\textbf{diff}$           -diff
-------------  -----------
Then           any
Instead        #ady
Unfortunately  #imate
Why            #cussion
Sometimes      #ze
Secondly       appreci
#Then          #raq
But            currently
Luckily        #kers
Anyway         #apixel
And            active
Suddenly       significant
Thankfully     #ade
Eventually     #imal
Somehow        specific
Fortunately    #ability
Meanwhile      anyone
What           #ker
Obviously      #unction
Because        reap
\end{lstlisting}
\\ 
\begin{lstlisting}[backgroundcolor=\color{white}, mathescape=true]
$\textbf{diff}$         -diff
-----------  ---------
bullshit     #avorite
anyway       #ilyn
crap         #xtap
anyways      #insula
unless       #cedented
nonsense     #aternal
#falls       #lyak
fuck         #rieve
#.           #uana
fallacy      #accompan
#tics        #ashtra
#punk        #icer
damned       #andum
#fuck        Mehran
stupidity    #andise
shit         #racuse
commercials  #assadors
because      #Chel
despite      rall
movies       #abella
\end{lstlisting}
&
\end{longtable}

\textbf{Layer 10}

2 out of 4

\noindent\begin{longtable}{ll}
\begin{lstlisting}[backgroundcolor=\color{white}, mathescape=true]
diff          $\textbf{-diff}$
------------  ------------
#,            Nope
work          Instead
#icle         Thankfully
#.            Surely
outdoors      #Instead
inspiring     Fortunately
exped         Worse
ahead         Luckily
together      #Thankfully
touches       Unless
out           Apparently
personalized  Perhaps
#joy          #Unless
#unction      #Fortunately
warm          Sorry
exceptional   Secondly
experience    #Luckily
lasting       #Rather
integ         Hence
#astic        Neither
\end{lstlisting}
&

\begin{lstlisting}[backgroundcolor=\color{white}, mathescape=true]
$\textbf{diff}$      -diff
--------  ---------
#sup      #etting
Amazing   #liness
#airs     #ktop
awesome   #ulkan
Bless     #enthal
Loving    #enance
my        #yre
#OTHER    #eeds
#BW       omission
#perfect  #reys
#-)       #lihood
amazing   #esian
#adult    #holes
perfect   syndrome
welcome   grievance
Rated     offenders
#Amazing  #wig
#anch     #hole
FANT      #creen
#anche    #pmwiki
\end{lstlisting}
\end{longtable}

\textbf{Layer 11}

2 out of 4

\noindent\begin{longtable}{ll}
\begin{lstlisting}[backgroundcolor=\color{white}, mathescape=true]
$\textbf{diff}$        -diff
----------  ------------
shots       #Kind
shit        suscept
bullshit    Fathers
stuff       #Footnote
tits        concess
crap        #accompan
boobs       Strait
creepy      #orig
noises      #ESE
spectacle   #ufact
boring      Founder
things      #iere
everything  #HC
noise       #Prev
#anim       #alias
ugly        participated
garbage     #Have
stupidity   #coe
visuals     #Father
selfies     strugg
\end{lstlisting}
&

\begin{lstlisting}[backgroundcolor=\color{white}, mathescape=true]
$\textbf{diff}$            -diff
--------------  -----------
#ly             #say
storytelling    actionGroup
sounding        prefers
spectacle       #ittees
#ness           #reon
#hearted        presumably
cinematic       waivers
#est            #aucuses
portrayal       #Phase
quality         #racuse
paced           #arge
combination     #hers
juxtap          #sup
representation  #later
mixture         expired
#!!!!!          stricter
filmmaking      #onds
enough          #RELATED
thing           #rollment
rendition       #orders
\end{lstlisting}
\end{longtable}

\subsection*{$\WV$ Subheads}
\textbf{Layer 9}

4 out of 4

\noindent\begin{longtable}{ll}
\begin{lstlisting}[backgroundcolor=\color{white}, mathescape=true]
diff         $\textbf{-diff}$
-----------  ----------
#":""},{"    honestly
#etooth      definitely
#ogenesis    hilarious
#verning     alot
broker       amazing
#ounces      funn
threatens    cinem
#astical     Cinem
foothold     comedic
intruder     Absolutely
#vernment    comedy
#activation  absolutely
#Oracle      amazingly
fugitive     satire
visitor      underrated
#assian      really
barrier      fantastic
#":[         enjoyable
#vier        REALLY
#oak         wonderful
\end{lstlisting}
&

\begin{lstlisting}[backgroundcolor=\color{white}, mathescape=true]
$\textbf{diff}$      -diff
--------  -------------
crap      jointly
shit      #verning
bullshit  #pora
fucking   #rocal
idiots    #raft
fuck      #etooth
goddamn   #estead
stupid    #ilitation
FUCK      #ourse
#fuck     migr
shitty    #ourses
damn      #iership
#shit     Pione
lol       #iscover
fuckin    pioneering
nonsense  #egu
crappy    #ivities
kinda     neighbourhood
Fuck      pioneer
idiot     nurt
\end{lstlisting}
\\
\begin{lstlisting}[backgroundcolor=\color{white}, mathescape=true]
$\textbf{diff}$          -diff
------------  --------------
crap          Pione
bullshit      pioneers
shit          complementary
vomit         pioneering
nonsense      #knit
stupid        #raits
idiots        Browse
fucking       #iscover
#shit         strengthened
idiot         #rocal
fuck          prosper
gimmick       Communities
stupidity     neighbourhoods
goddamn       #Learn
shitty        strengthens
incompetence  #iscovery
lame          #ributes
FUCK          strengthen
inco          #izons
blah          Mutual
\end{lstlisting}
&
\begin{lstlisting}[backgroundcolor=\color{white}, mathescape=true]
$\textbf{diff}$       -diff
---------  --------------
anime      #rade
kinda      #jamin
stuff      #ounces
shit       #pherd
lol        Unable
tho        #pta
realism    Roche
damn       Payments
:)         Gupta
fucking    #odan
alot       #uez
movie      #adr
funny      #ideon
anyways    #Secure
enjoyable  #raught
crap       Bei
comedy     sovere
genre      unsuccessfully
anyway     #moil
fun        #Register
\end{lstlisting}

\end{longtable}

\textbf{Layer 10}

4 out of 4

\noindent\begin{longtable}{ll}
\begin{lstlisting}[backgroundcolor=\color{white}, mathescape=true]
diff           $\textbf{-diff}$
-------------  ------------
#knit          crap
welcomes       bullshit
Together       idiots
Growing        stupid
#Explore       shitty
pioneering     incompetence
complementary  pointless
milestone      goddamn
pioneer        retarded
#Together      lame
strengthens    Worse
#ossom         crappy
pioneers       incompet
#Learn         shit
jointly        stupidity
#Growing       fucking
embraces       Nope
#"},{"         FUCK
sharing        incompetent
#Discover      pathetic
\end{lstlisting}
&

\begin{lstlisting}[backgroundcolor=\color{white}, mathescape=true]
diff         $\textbf{-diff}$
-----------  ---------
#"}],"       crap
#verning     stupid
#etooth      shit
#"},{"       fucking
Browse       fuck
#Register    shitty
#Lago        bullshit
#raft        crappy
#egu         idiots
jointly      horrible
#iership     stupidity
strengthens  kinda
Scotia       goddamn
#ounces      awful
#uania       mediocre
#iann        pathetic
workspace    #fuck
seiz         damn
Payments     FUCK
#Learn       damned
\end{lstlisting}
\\
\begin{lstlisting}[backgroundcolor=\color{white}, mathescape=true]
$\textbf{diff}$          $\textbf{-diff}$
------------  -------------
bullshit      inspiring
incompetence  unforgettable
Worse         #knit
idiots        #love
crap          passions
dummy         cherish
incompetent   richness
Nope          timeless
stupid        loves
retarded      passionate
lame          beautifully
nonexistent   overcoming
wasting       unique
#Fuck         highs
bogus         nurture
worse         unparalleled
nonsense      vibrant
ineligible    #beaut
pointless     intertw
inco          insepar
\end{lstlisting}
&
\begin{lstlisting}[backgroundcolor=\color{white}, mathescape=true]
$\textbf{diff}$          -diff
------------  -------------
bullshit      Pione
crap          pioneers
stupid        pioneering
nonsense      complementary
incompetence  #knit
idiots        #Learn
shit          #accompan
stupidity     pioneer
pointless     invaluable
inco          #ossom
retarded      #Together
idiot         Browse
vomit         versatile
lame          welcomes
meaningless   #"},{"
goddamn       admired
nonsensical   jointly
garbage       Sharing
#shit         Together
useless       #Discover
\end{lstlisting}
\end{longtable}

\textbf{Layer 11}

4 out of 4

\noindent\begin{longtable}{ll}
\begin{lstlisting}[backgroundcolor=\color{white}, mathescape=true]
diff          $\textbf{-diff}$
------------  ------------
Provision     alot
issuance      amazing
Securities    kinda
#ogenesis     fucking
Holdings      awesome
Regulatory    funny
indefinitely  damn
Advisory      REALLY
designation   hilarious
unilaterally  tho
Province      unbelievable
Regulation    fuckin
#Lago         wonderful
issued        doesnt
Recep         definitely
Advis         thats
#verning      yeah
broker        fantastic
#Mobil        badass
Policy        dont
\end{lstlisting}
&

\begin{lstlisting}[backgroundcolor=\color{white}, mathescape=true]
$\textbf{diff}$          -diff
------------  ---------
crap          #rocal
fucking       #verning
bullshit      #etooth
fuck          #uania
goddamn       caches
shit          Browse
#fuck         #"},{"
stupidity     #imentary
pathetic      exerc
spoiler       #Lago
stupid        #"}],"
inept         #cium
blah          #enges
FUCK          #ysis
awful         quarterly
shitty        #iscover
trope         Scotia
Godd          #resso
inco          #appings
incompetence  jointly
\end{lstlisting}
\\
\begin{lstlisting}[backgroundcolor=\color{white}, mathescape=true]
$\textbf{diff}$            $\textbf{-diff}$
--------------  ------------
pioneers        bullshit
pioneering      crap
Browse          shit
Pione           idiots
complementary   stupid
#knit           vomit
prosper         incompetence
#raits          nonsense
#Trend          gimmick
#ributes        stupidity
#Learn          idiot
strengthen      shitty
strengthened    fucking
#ossom          lame
pioneer         crappy
#iscover        goddamn
#Growing        pointless
prosperity      inco
neighbourhoods  #shit
#owship         Nope
\end{lstlisting}
&
\begin{lstlisting}[backgroundcolor=\color{white}, mathescape=true]
$\textbf{diff}$          $\textbf{-diff}$
------------  -------------
Worse         #knit
bullshit      pioneers
Nope          pioneering
crap          inspiring
incompetence  #iscover
idiots        complementary
incompetent   pioneer
stupid        #ossom
incompet      passionate
pointless     passions
inco          journeys
Stupid        unique
meaningless   embraces
nonsense      admired
lame          forefront
idiot         richness
worse         invaluable
#Fuck         prosper
whining       vibrant
nonsensical   enriched
\end{lstlisting}

\end{longtable}

\subsection*{$\WO$ Subheads}
}
\textbf{Layer 9}

0 out of 4

\textbf{Layer 10}

0 out of 4

\textbf{Layer 11}

0 out of 4

\end{document}